\documentclass[10pt,twocolumn,letterpaper]{article}

\usepackage{iccv}
\usepackage{times}
\usepackage{epsfig}
\usepackage{graphicx}
\usepackage{amsmath}
\usepackage{amssymb}
\usepackage{tabularx}
\usepackage{multirow}
\usepackage{caption}
\usepackage{hhline}
\usepackage[export]{adjustbox}
\usepackage{subfig}
\usepackage{tabu}
\usepackage{comment}
\usepackage{ctable}
\usepackage{boldline}
\usepackage{pifont}
\usepackage{tablefootnote}
\usepackage{footnote}

\usepackage[pagebackref=true,breaklinks=true,letterpaper=true,colorlinks,bookmarks=false]{hyperref}

\iccvfinalcopy 

\def\iccvPaperID{2252} 

\ificcvfinal\fi

\begin{document}


\title{StereOBJ-1M: Large-scale Stereo Image Dataset for 6D Object Pose Estimation}

\author{Xingyu Liu \qquad Shun Iwase \qquad Kris M. Kitani \\ Carnegie Mellon University}

\maketitle
\ificcvfinal\thispagestyle{empty}\fi

\definecolor{viz_red}{RGB}{204, 0, 0}
\definecolor{viz_yellow}{RGB}{241, 194, 50}
\definecolor{viz_green}{RGB}{52, 183, 47}
\definecolor{viz_blue}{RGB}{17, 85, 204}

\newcommand\chck{{\color{black}\ding{51}}}
\newcommand\chx{{\color{black}\ding{55}}}


\begin{abstract}
We present a large-scale stereo RGB image object pose estimation dataset named the \textbf{StereOBJ-1M} dataset. 
The dataset is designed to address challenging cases such as object transparency, translucency, and specular reflection, in addition to the common challenges of occlusion, symmetry, and variations in illumination and environments. 
In order to collect data of sufficient scale for modern deep learning models, we propose a novel method for efficiently annotating pose data in a multi-view fashion that allows data capturing in complex and flexible environments.  
Fully annotated with 6D object poses, our dataset contains over 393K frames and over 1.5M annotations of 18 objects recorded in 182 scenes constructed in 11 different environments.
The 18 objects include 8 symmetric objects, 7 transparent objects, and 8 reflective objects.
We benchmark two state-of-the-art pose estimation frameworks on StereOBJ-1M as baselines for future work.
We also propose a novel object-level pose optimization method for computing 6D pose from keypoint predictions in multiple images.
Project website: \url{https://sites.google.com/view/stereobj-1m}.
\end{abstract}

\vspace{-2ex}
\section{Introduction}

Effectively leveraging 3D cues from visual data to infer the pose of an object is crucial for applications such as augmented reality (AR) and robotic manipulation.
Compared to objects with opaque and Lambertian surfaces, estimating the pose of transparent and reflective objects is especially challenging. 
To leverage depth information from sensors, previous approaches have explored deep models that take RGB-D maps as input \cite{densefusion,nocs,pvn3d,maskedfusion,robust:6d:rgbd}.
Unfortunately, as the experiments in \cite{keypose,transparent:6d:pose,cleargrasp,transparent:tof} have shown, existing commercial depth-sensing methods, such as time-of-flight (ToF) or projected light sensors, failed to capture the depths of transparent or reflective surfaces.
As a result, monocular RGB-D maps cannot serve as a reliable input for object pose estimation models in these challenging scenarios. 
Based on this observation, we focus on using stereo RGB images as our input modality, allowing for object pose estimation on a wider range of objects, including transparent or highly reflective objects.

\begin{figure}[t] 
\centering
\newcommand\teaserwidth{0.238}
\subfloat{
    \includegraphics[width=0.75\linewidth, valign=t]{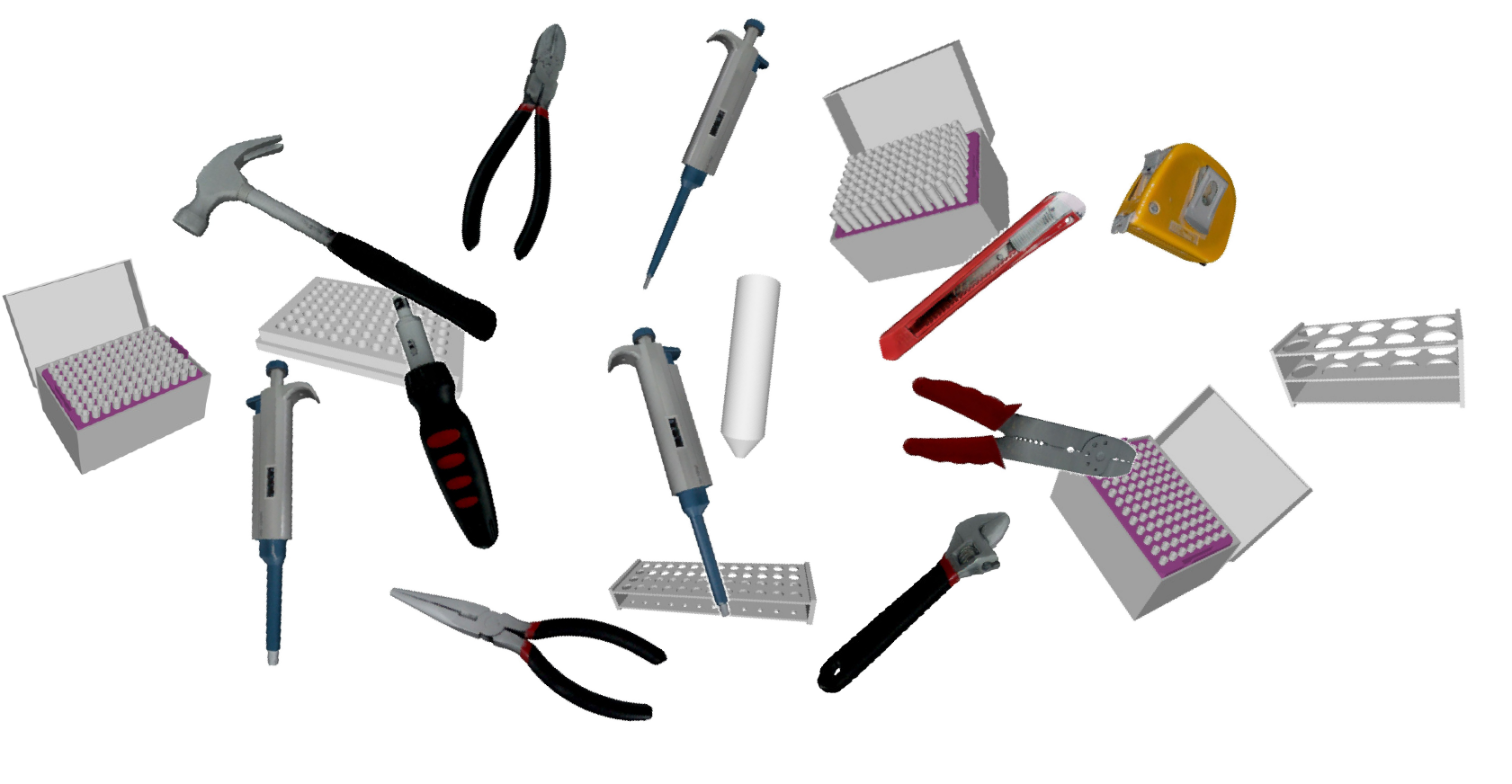}
}
\vspace{-2ex}
\\
\subfloat{
    \includegraphics[height=\teaserwidth\linewidth, valign=t]{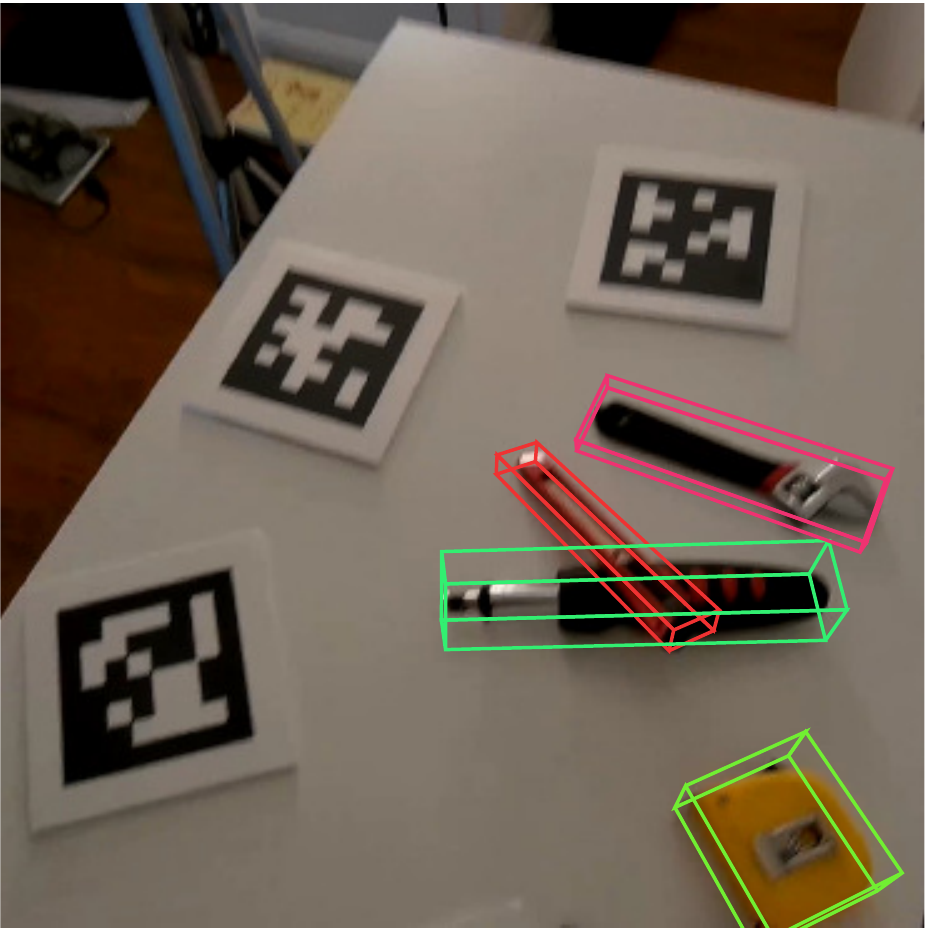}
}
\hspace{-1.25ex}
\subfloat{
    \includegraphics[height=\teaserwidth\linewidth, valign=t]{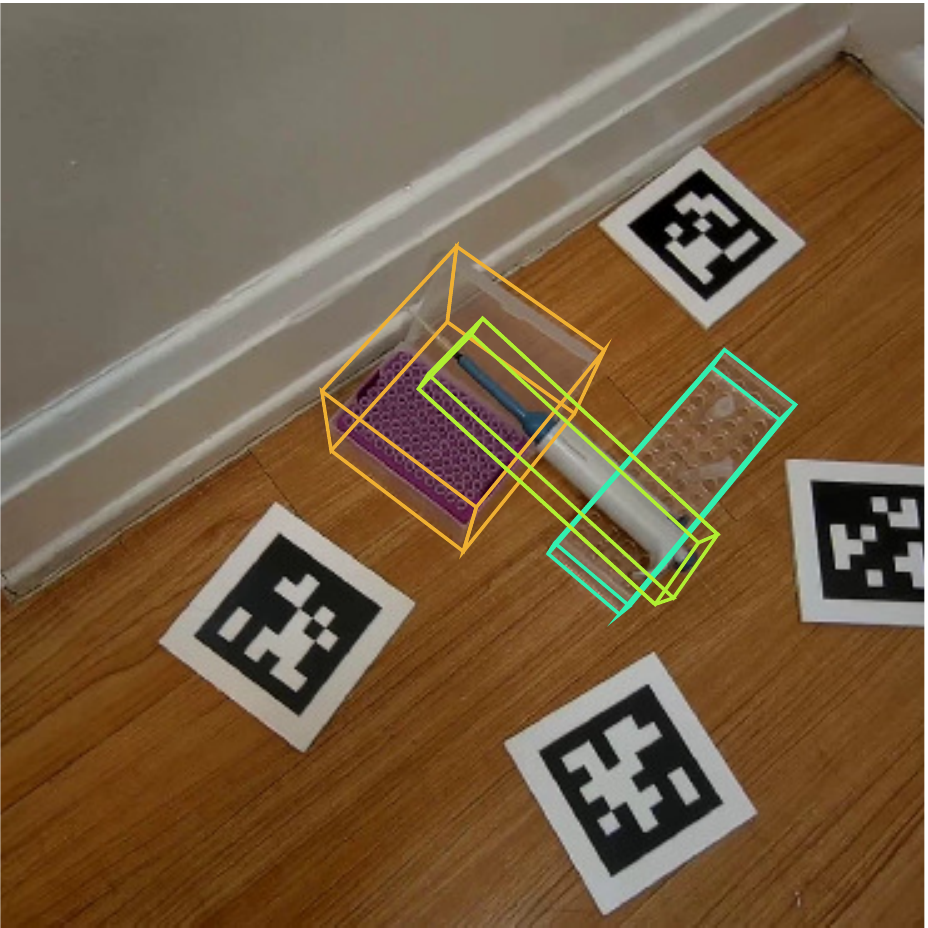}
}
\hspace{-1.25ex}
\subfloat{
    \includegraphics[height=\teaserwidth\linewidth, valign=t]{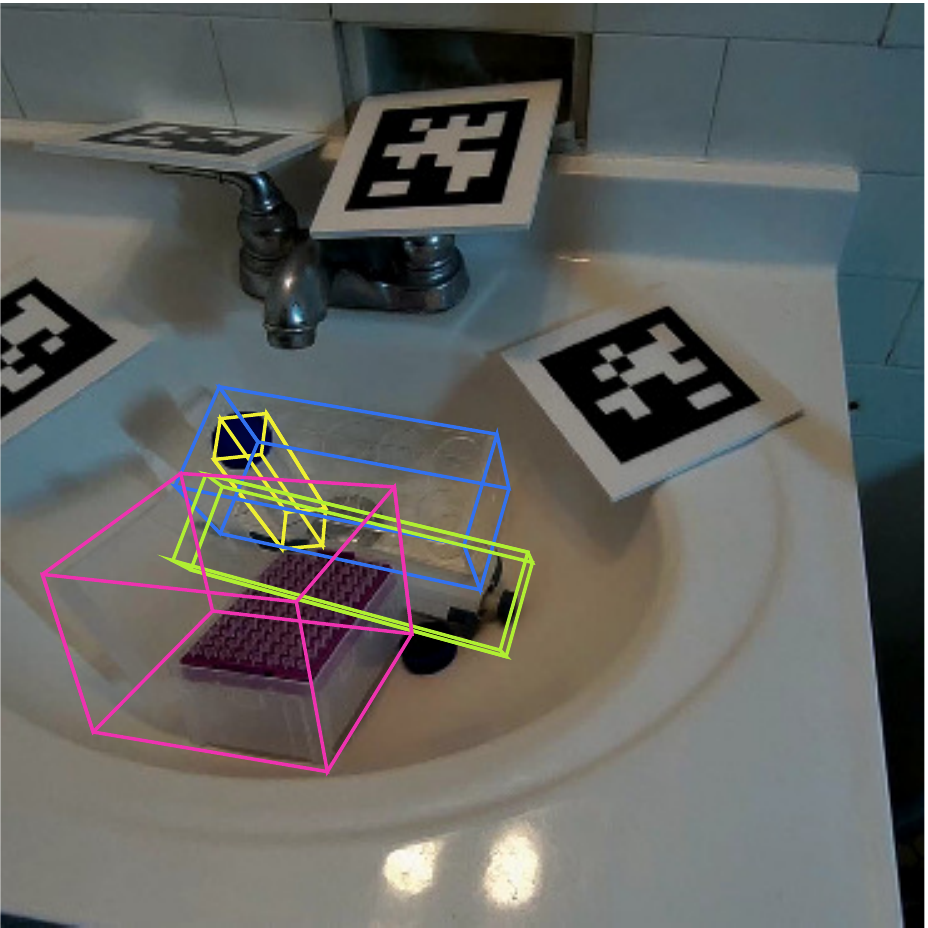}
}
\hspace{-1.25ex}
\subfloat{
    \includegraphics[height=\teaserwidth\linewidth, valign=t]{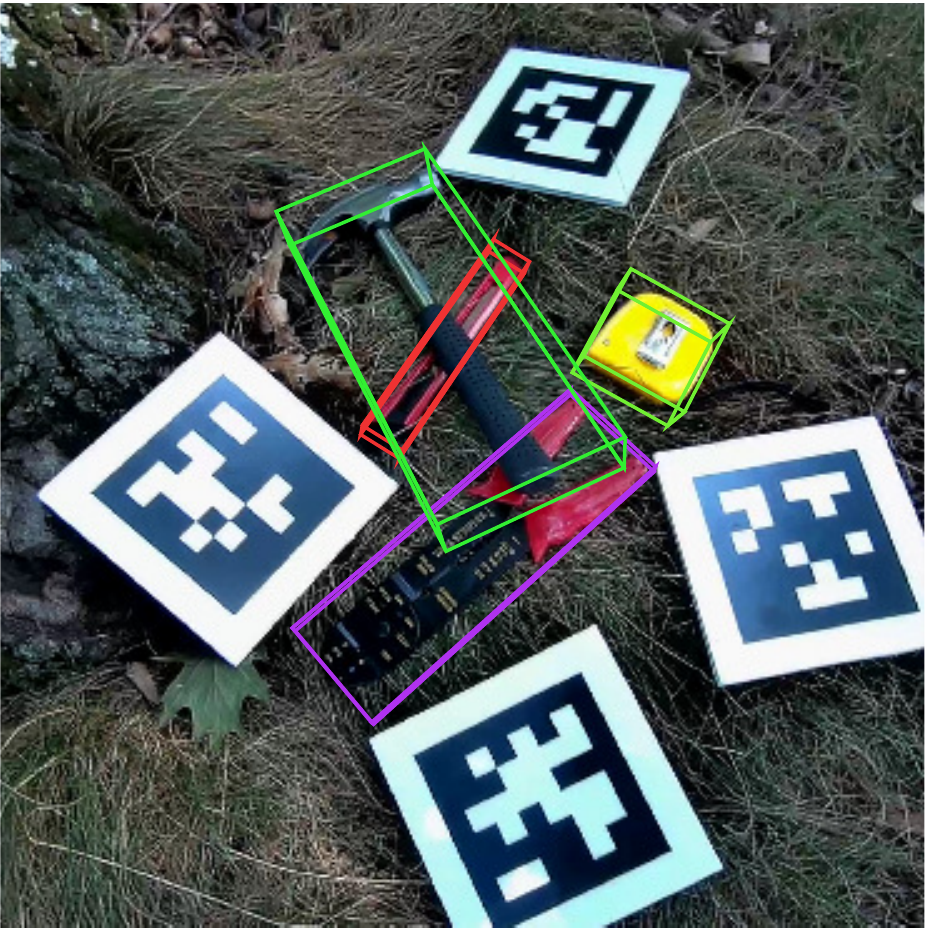}
}
\\
\vspace{-3ex}
\subfloat{
    \includegraphics[height=\teaserwidth\linewidth, valign=t]{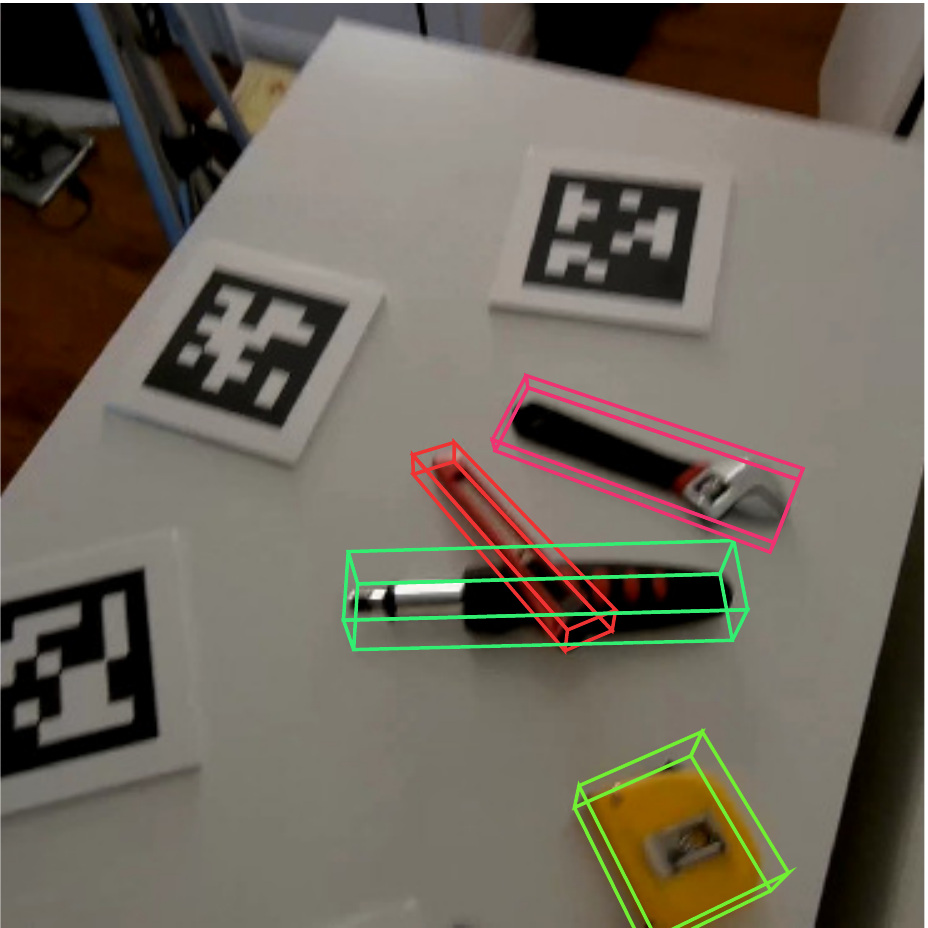}
}
\hspace{-1.25ex}
\subfloat{
    \includegraphics[height=\teaserwidth\linewidth, valign=t]{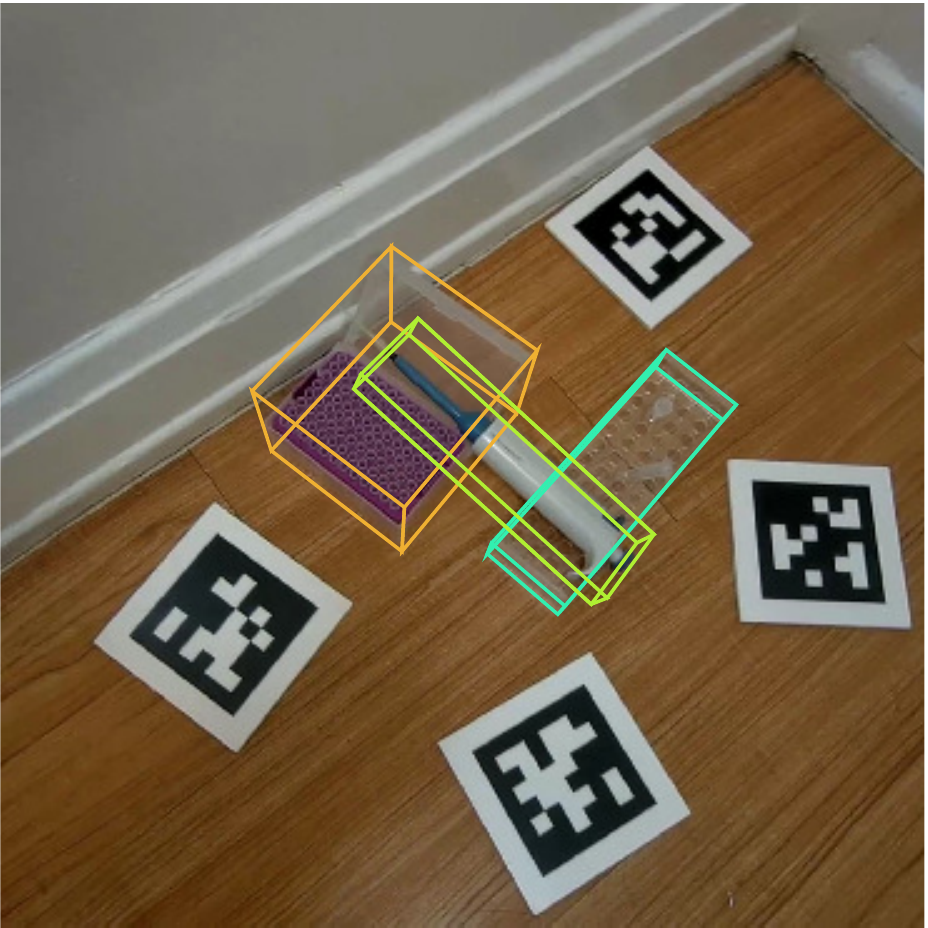}
}
\hspace{-1.25ex}
\subfloat{
    \includegraphics[height=\teaserwidth\linewidth, valign=t]{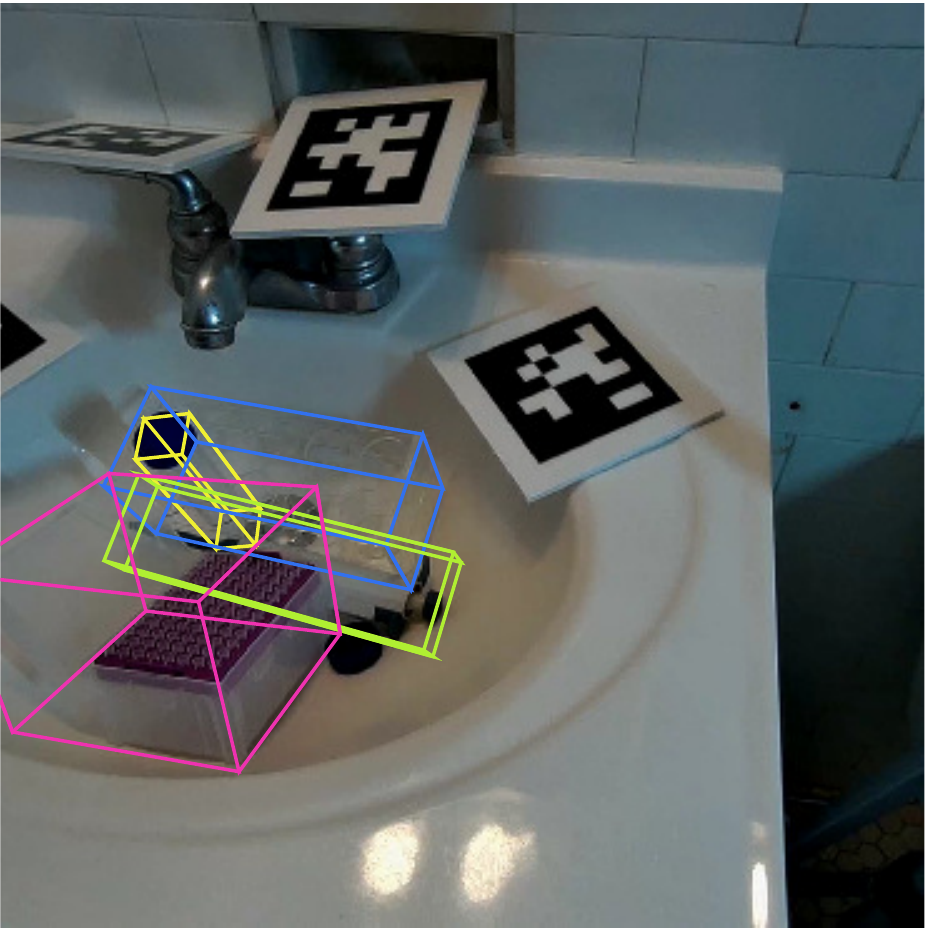}
}
\hspace{-1.25ex}
\subfloat{
    \includegraphics[height=\teaserwidth\linewidth, valign=t]{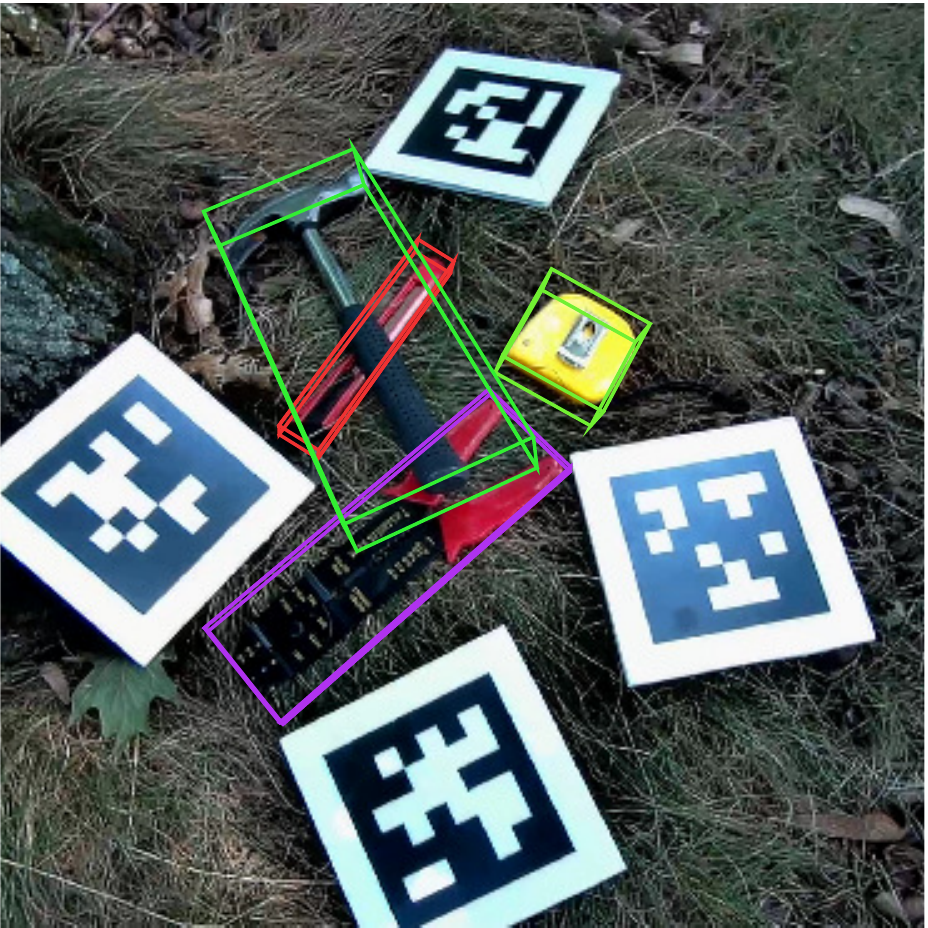}
}
\vspace{-1ex}
\caption{ \textbf{StereOBJ-1M dataset.} Upper: CAD models of the objects in the dataset. Lower: four data stereo image pair samples with bounding box annotations.
}
\vspace{-1ex}
\label{fig:teaser}
\end{figure}

A major challenge in modern object pose estimation is that of acquiring a large-scale training dataset. 
To increase data size for training large-scale neural networks, previous works have explored leveraging synthetically rendered \cite{fat,pvnet,t-less} or augmented images \cite{nocs} with 3D mesh models.
However, photorealistic rendering is still challenging with only basic graphics rendering tools and limited expertise. 
Synthetic image datasets that are currently available typically introduce a very large domain gap.
This is especially true for transparent and reflective objects where variations in illumination and background scenes are crucial but difficult to model.

\begin{savenotes}
\begin{table*}[ht]
\small
\centering
\resizebox{\textwidth}{!}{
\begin{tabular}{l|c|c|c|c|c|c|c|c|c|c|c}
\hline
\multirow{2}{*}{dataset} & \multirow{2}{*}{data type} & \multirow{2}{*}{stereo} & \multirow{2}{*}{depth} & \multirow{2}{*}{occlusion} & \multicolumn{1}{c|}{\multirow{2}{*}{\begin{tabular}[c]{@{}c@{}} transparent \\ objects \end{tabular}}} & \multicolumn{1}{c|}{\multirow{2}{*}{\begin{tabular}[c]{@{}c@{}} reflective \\ objects \end{tabular}}} & \multicolumn{1}{c|}{\multirow{2}{*}{\begin{tabular}[c]{@{}c@{}} \# of \\ frames  \end{tabular}}} &  \multicolumn{1}{c|}{\multirow{2}{*}{\begin{tabular}[c]{@{}c@{}} \# of outdoor \\ environments  \end{tabular}}} & \multicolumn{1}{c|}{\multirow{2}{*}{\begin{tabular}[c]{@{}c@{}} \# of \\ scenes \end{tabular}}} & \multicolumn{1}{c|}{\multirow{2}{*}{\begin{tabular}[c]{@{}c@{}} \# of \\ objects \end{tabular}}} &
\multicolumn{1}{c}{\multirow{2}{*}{\begin{tabular}[c]{@{}c@{}} \# of \\ annotations \end{tabular}}} \\
 & & & & & & &  &  &  & & \\
\hline
FAT \cite{fat} & synthetic & \chck & \chck & \chck & \chx & \chx &  60,000 & 0 & 3,075 & 21 & 205,359 \\ 
\hline
CAMERA \cite{nocs} & mixed reality & \chx & \chck & \chck & \chx & \chx & 300,000 & 0 & 30 & 42 & 4,350,656 \\ 
\hline
YCB \cite{ycb} & real & \chx & \chck & \chck & \chx & \chx & 133,936 &  0 & 92 & 21 & 613,917 \\ 
\hline
LINEMOD \cite{linemod} & real & \chx & \chck & \chck & \chx & \chx & 18,000 & 0 & 15 & 15 & 15,784 \\ 
\hline
GraspNet-1Billion \cite{graspNet-1billion} & real & \chx & \chck & \chck & \chx & \chx  & 97,280 & 0 & 190 & 88 & 970,000 \\
\hline
T-LESS \cite{t-less} & real & \chx & \chck & \chck & \chx & \chx & 47,762 & 0 & - & 30 & 47,762 \\
\hline
kPAM \cite{kpam} & real & \chx & \chck & \chck & \chx & \chck & 100,000 & 0 & - & 91 & - \\
\hline
LabelFusion \cite{labelfusion} & real & \chx & \chck & \chck & \chx & \chck & 352,000 & 0 & 138 & 12 &1,000,000 \\
\hline
REAL275 \cite{nocs} & real & \chx & \chck & \chx & \chx & \chx & 7,072 & 0 & 13 &  42 & 35,356 \\ 
\hline
TOD \cite{keypose} & real & \chck & \chck & \chx & \chck & \chx & 64,000 & 0 & 10 & 20 & 64,000 \\ 
\hline
\textbf{StereOBJ-1M (Ours)} & real & \chck & \chx & \chck & \chck & \chck & \textbf{393,612}
 & \textbf{3} & 182 & 18 & \textbf{1,508,327}  \\
\hline
\end{tabular}
} 
\vspace{-1ex}
\caption{Dataset Comparisons. Our StereOBJ-1M dataset is the only large-scale 6DoF object pose dataset that provides stereo RGB as input modality, includes transparent and reflective objects and is captured in both indoor and outdoor environments. In terms of capacity, our dataset is also the largest real-image dataset in size and the dataset with the most scene diversity.}
\label{tab:dataset:compare}
\vspace{-2ex}
\end{table*}
\end{savenotes}

To address the challenge of costly pose data acquisition, and to enable further training and evaluation of modern object pose estimation models, we introduce \textbf{a novel method for capturing and labeling a large-scale dataset} with high efficiency and quality. 
Our method is based on multi-view geometry to accurately localize fiducial markers, cameras, and object keypoints in the scene. 
We use a hand-held stereo camera to record video data. 
With the help of two other static cameras mounted to tripods, the positions of a set of fiducial markers can be calculated on the fly, from which the pose of every recorded frame can be automatically computed. 
By annotating 2D object keypoints in just a few frames selected in a long recorded video, the 3D locations of the keypoints can be computed by triangulation.
The 6D poses of objects can then be calculated by aligning 3D CAD models to the keypoints before being propagated to all other frames.

Using the procedure outlined above, we generate \textbf{StereOBJ-1M dataset}, the first pose dataset with stereo RGB as input modality with over 100K frames. 
It is also the largest 6D object pose dataset in history: it consists of 393,612 high-resolution stereo frames and over 1.5 million 6D pose annotations of 18 objects recorded in 182 indoor and outdoor scenes.
The capacity of StereOBJ-1M is sufficient for training large-scale neural networks without additional synthetic images. 
The average labeling error of StereOBJ-1M is 2.3mm which is the best annotation precision among all public object pose datasets.

We implement two state-of-the-art methods \cite{pvnet,keypose} as the baseline comparisons for 6D pose estimation using stereo on the StereOBJ-1M dataset.
To handle 2D-3D correspondences predictions in two or more images, we propose a novel object-level 6D pose optimization approach named \textbf{Object Triangulation}.
Contrary to classic triangulation that optimizes the 3D location of a \emph{point}, we directly optimize the 6D pose of an \emph{object} from 2D keypoint locations in multiple images.
Experiment results show that Object Triangulation consistently improves pose estimation over monocular input while classic triangulation can yield worse results.
With Object Triangulation, the stereo variants of both baseline methods significantly outperform their monocular counterparts on StereOBJ-1M, by at least 25\% in ADD(-S) AUC and 14\% in ADD(-S) accuracy, which highlights the importance of stereo modality in object pose estimation.
We expect that StereOBJ-1M will serve as a common benchmark dataset for stereo RGB-based object pose estimation.

\section{Related Work}

\textbf{Pose Annotation Methods.}
The first category of pose data annotation methods relies on capturing RGB-D images, reconstructing 3D point clouds, and labeling pose by constructing a 3D mesh \cite{kpam}, or fitting 3D object mesh
models to 3D point clouds
\cite{labelfusion,ycb,linemod}.
However, this type of method cannot reliably deal with transparent objects where depth sensing is usually not possible.
The second category of pose data annotation methods adopts keypoint as representation and leverages multi-view geometry for triangulation \cite{panoptic:studio,keypose}.
Our novel data annotation method is keypoint and multi-view based.
Different from previous methods, we record the scenes using a stereo RGB camera whose poses are computed on the fly based on fiducial markers whose locations are also computed on the fly.

\begin{figure*}[ht] 
  \centering
    \includegraphics[width=\linewidth]{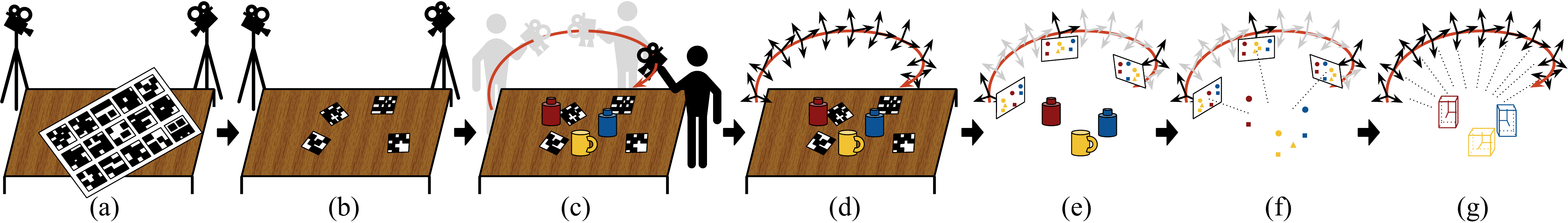} 
  \vspace{-4ex}
  \caption{Our data capturing and labeling pipeline for one captured video: (a) use PnP to calculate fixed cameras' global poses from fudicial marker board ; (b) triangulation of fudicial marker locations; (c) scanning the scene with stereo camera; (d) use PnP to calculate moving camera's global poses from fudicial markers; (e) annotate keypoints on sampled images; (f) triangulation of object 3D keypoints; (g) 6D pose fitting from 3D keypoints and propagation to all images. 
  }
  \label{fig:data:pipeline}
  \vspace{-1ex}
\end{figure*}

\textbf{Stereo Methods.}
Studying correspondence, depth, and other downstream tasks from two or multiple RGB images has been a long-standing topic in computer vision and robotics. 
Previous works have explored stereo-based methods for 3D object detection \cite{stereo:r-cnn}, disparity estimation \cite{segstereo}, point-based 3D reconstruction \cite{pointmvsnet} and keypoint detection \cite{keypose}.
Recently, multi-view based methods have been proposed for object pose estimation \cite{unified:multi:view:pose,cosypose}.
Our 6D object pose dataset provides binocular stereo RGB images as the input modality, allowing stereo-based deep methods to be trained on object pose data.
In addition, the  annotation process of our dataset utilizes multi-view geometry.

\textbf{Keypoint in Pose Representation.}
Keypoints is a popular intermediate representation for object or human pose.
Previous work has explored deep learning methods for localing keypoints of an object \cite{keypointnet,6dof:pose:from:kp,keypose,pvn3d,kundu2018object} or a human \cite{integralnet,monocular:3d:human:pose,openpose} from a RGB image.
Several public object pose estimation datasets are also constructed by using keypoints to simplify pose annotation \cite{labelfusion,keypose}.
Our data annotation pipeline also uses keypoints as a bridge to the 6D pose, where the 3D positions of the keypoints are calculated through multi-view triangulation.

\textbf{Related Dataset.}
Most existing pose datasets provide RGB-D as input modality \cite{ycb, ycb-m, linemod, graspNet-1billion, t-less, kpam, labelfusion, nocs}.
Since directly labeling 3D object pose in real RGB images is costly and inaccurate, most existing datasets rely on capturing RGB-D images and fitting 3D mesh models to 3D point clouds as their labeling method \cite{labelfusion,ycb,nocs,t-less,linemod}.
TOD \cite{keypose} is the first object pose dataset with binocular stereo RGB as the input modality, and it uses a data labeling method based on multi-view geometry.
However, TOD records in a studio environment and does not include occluded objects.
Our dataset provides binocular stereo RGB as input modality and records objects with occlusion in 11 different real environments. 
A more comprehensive comparison of datasets is illustrated in Table \ref{tab:dataset:compare}.

\section{Data Capturing \& Labeling Pipeline}

One of the major challenges in the pose estimation of 3D objects is the acquisition and annotation of large-scale and high-quality real object pose data.
Limitations of previous efforts are in the following three aspects.

\textbf{Sensor Modality.} Most existing datasets such as \cite{ycb, ycb-m, linemod, t-less, kpam, labelfusion, nocs} only provide monocular RGBD from commercial depth sensors as 3D cue.
These datasets and the associated labeling methods did not and could not handle transparent or reflective objects on which depth sensing is not reliable.
Moreover, different technologies of depth sensing, e.g. infrared and LiDAR, may return different depths for the same objects and scenes.
Thus a model trained on one RGBD-based pose dataset may not be able to generalize to another with a different depth-sensing technology.

\textbf{Data Annotation.} 
Existing data annotation methods usually require annotators to manually align object CAD model to 3D sensor signals, e.g. reconstructed 3D point clouds from depth maps, which are expensive and inaccurate.
Limited by the cost of data annotation, the sizes of public real-world datasets such as \cite{ycb-m,linemod,t-less,nocs} are in the order of 10K or fewer images, which are insufficient for training large-scale deep neural net models.
An alternative solution is to leverage synthetically rendered or augmented images.
However, the problem of the domain gap is still yet to be solved and is especially challenging for transparent and reflective objects.

\textbf{Scene Environment.} Datasets such as \cite{keypose,graspNet-1billion,t-less,linemod} were captured in a small number ($<3$) of special indoor environments or studios and lack the diversity of real scenes. It is hard for models trained on such data to generalize to unseen environments, especially for transparent and reflective objects where background scenes and illumination are crucial.

\subsection{Data Capturing and Labeling}

To address the above problems, we propose a novel method for efficiently capturing and labeling 3D object pose data.
We opt to use stereo RGB modality to provide 3D cues for the data.
For labeling, our philosophy is to abandon depth sensing and utilize multi-view geometry for high-precision 3D localization of object keypoints for pose fitting.
An overview of our pipeline is illustrated in Figure \ref{fig:data:pipeline}.
It consists of the following seven steps which respectively correspond to Figure \ref{fig:data:pipeline}(a)-(g).

\textbf{1. Pose Calculation for Static Cameras.}
We set up two static cameras that record the scene simultaneously.
The cameras are held by two tripods.
To obtain the camera poses in the world coordinate,
we place a large customized plastic board printed with an array of fiducial markers into the scene such that most of the fiducial markers are visible in both static cameras.
The accurate 3D positions of the fiducial markers on the board are measured by a vernier caliper and act as the world coordinate in the rest of the pipeline.
The poses of the two static cameras in world coordinates $[\mathbf{R}^\text{S}_1, \mathbf{T}^\text{S}_1] \in \mathbb{R}^{3\times4} $ and $[\mathbf{R}^\text{S}_2, \mathbf{T}^\text{S}_2]$ are calculated with Perspective-n-Point (PnP) algorithm \cite{pnp:ransac}.

\textbf{2. Triangulation of Fiducial Markers. }
We remove the plastic fiducial marker array board and place several other small fiducial markers into the scene such that they are visible in both static cameras.
The dimensions of the small fiducial marker boards are also accurately measured by a vernier caliper.
From $[\mathbf{R}^\text{S}_1, \mathbf{T}^\text{S}_1]$ and $[\mathbf{R}^\text{S}_2, \mathbf{T}^\text{S}_2]$, we use triangulation to locate the 3D locations of the corners of the small fiducial markers in world coordinates $\{ \textbf{x}^{\text{F}}_i \in \mathbb{R}^{4\times3} \}$. 
During this step, the two static cameras continue to record video and their poses remain unchanged.

\textbf{3. Scene Construction and Scanning. }
To construct the scene, we first place a few randomly selected objects from our dataset and mingle them with the small fiducial markers. 
Other occluding objects can also be included if necessary.
Note that during this step, the positions of the small fiducial markers must remain unchanged while the static cameras can be removed.
Then a human data collector holds a stereo RGB camera, slowly moves it around the scene, scans the objects from different viewpoints, and record a stereo video. 
The scanning paths are selected aiming to cover as many viewpoints as possible.

\begin{figure*}[ht]
\newcommand\objheight{0.107}
\centering
\small
\subfloat[]{
    \includegraphics[height=\objheight\linewidth, valign=t]{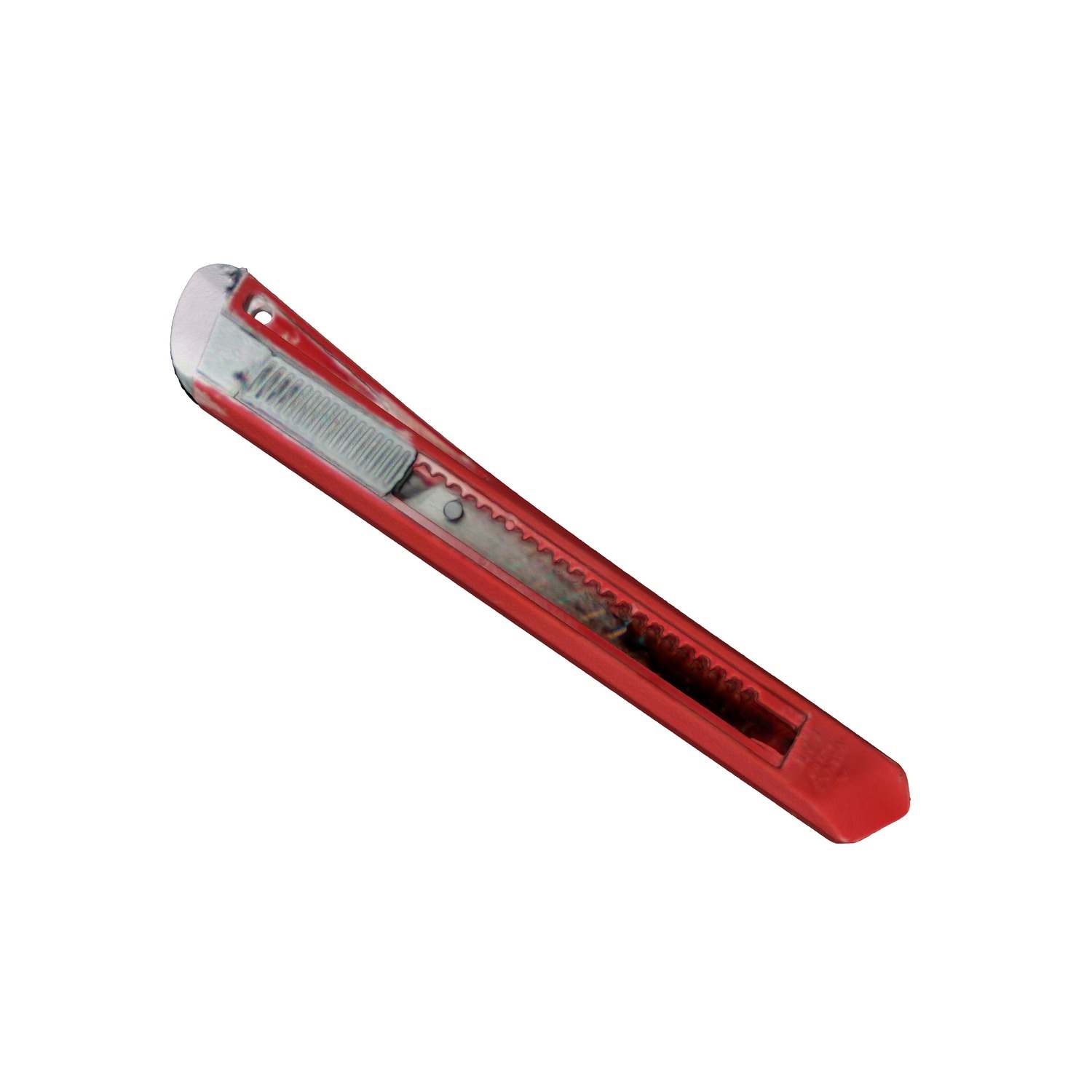}
}
\hspace{-0.01\linewidth}
\subfloat[]{
    \includegraphics[height=\objheight\linewidth, valign=t]{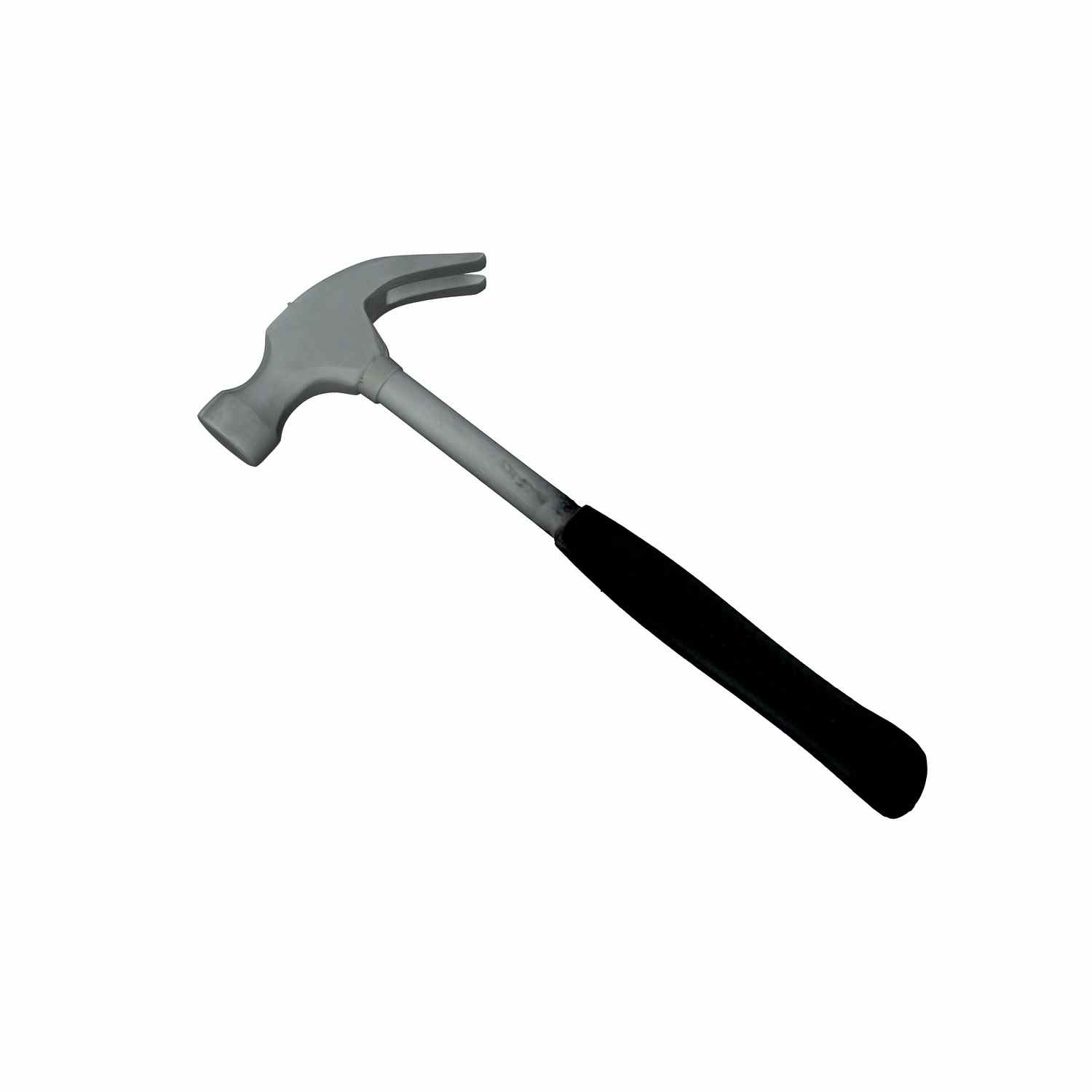}
}
\hspace{-0.01\linewidth}
\subfloat[]{
    \includegraphics[height=\objheight\linewidth, valign=t]{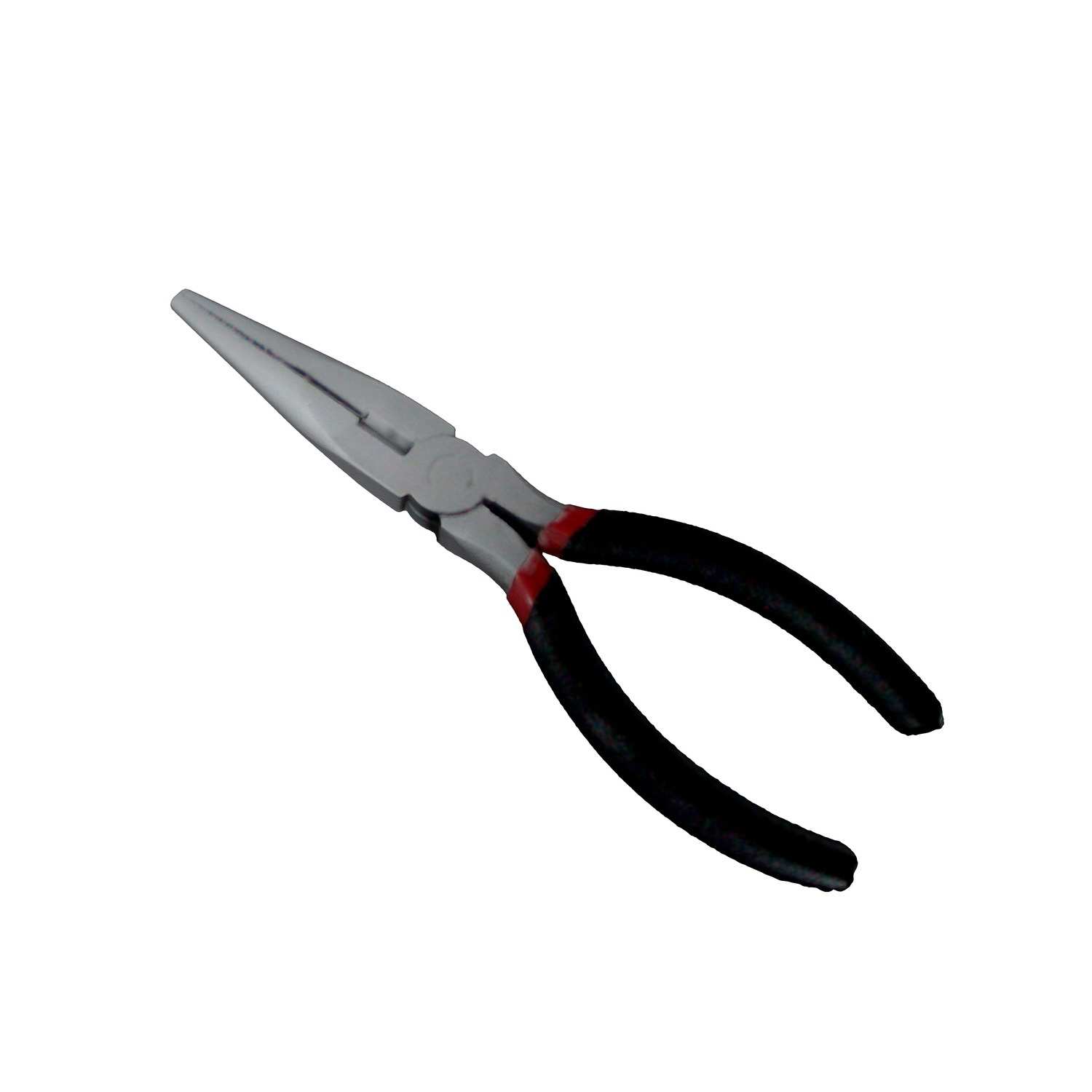}
}
\hspace{-0.01\linewidth}
\subfloat[]{
    \includegraphics[height=\objheight\linewidth, valign=t]{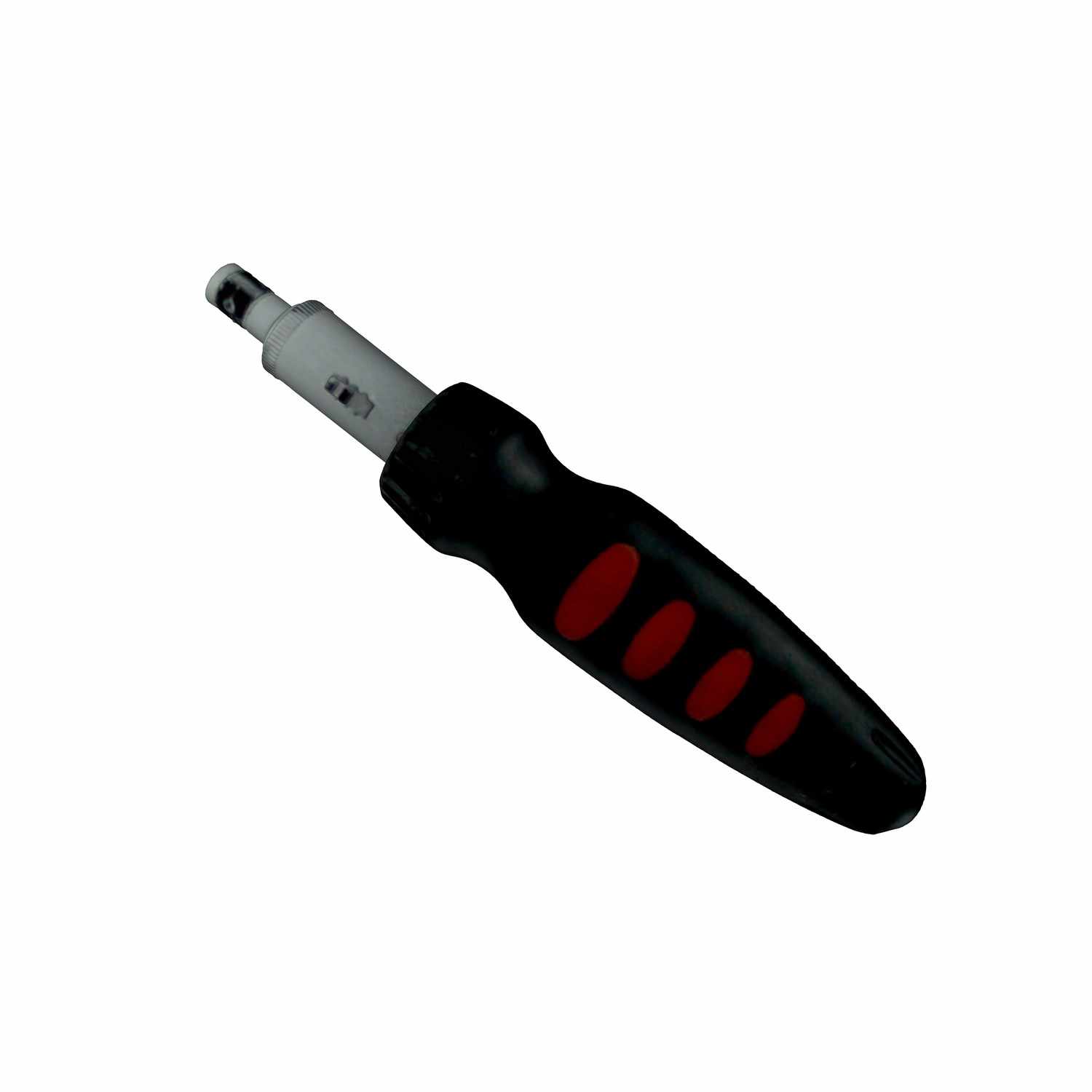}
}
\hspace{-0.01\linewidth}
\subfloat[]{
    \includegraphics[height=\objheight\linewidth, valign=t]{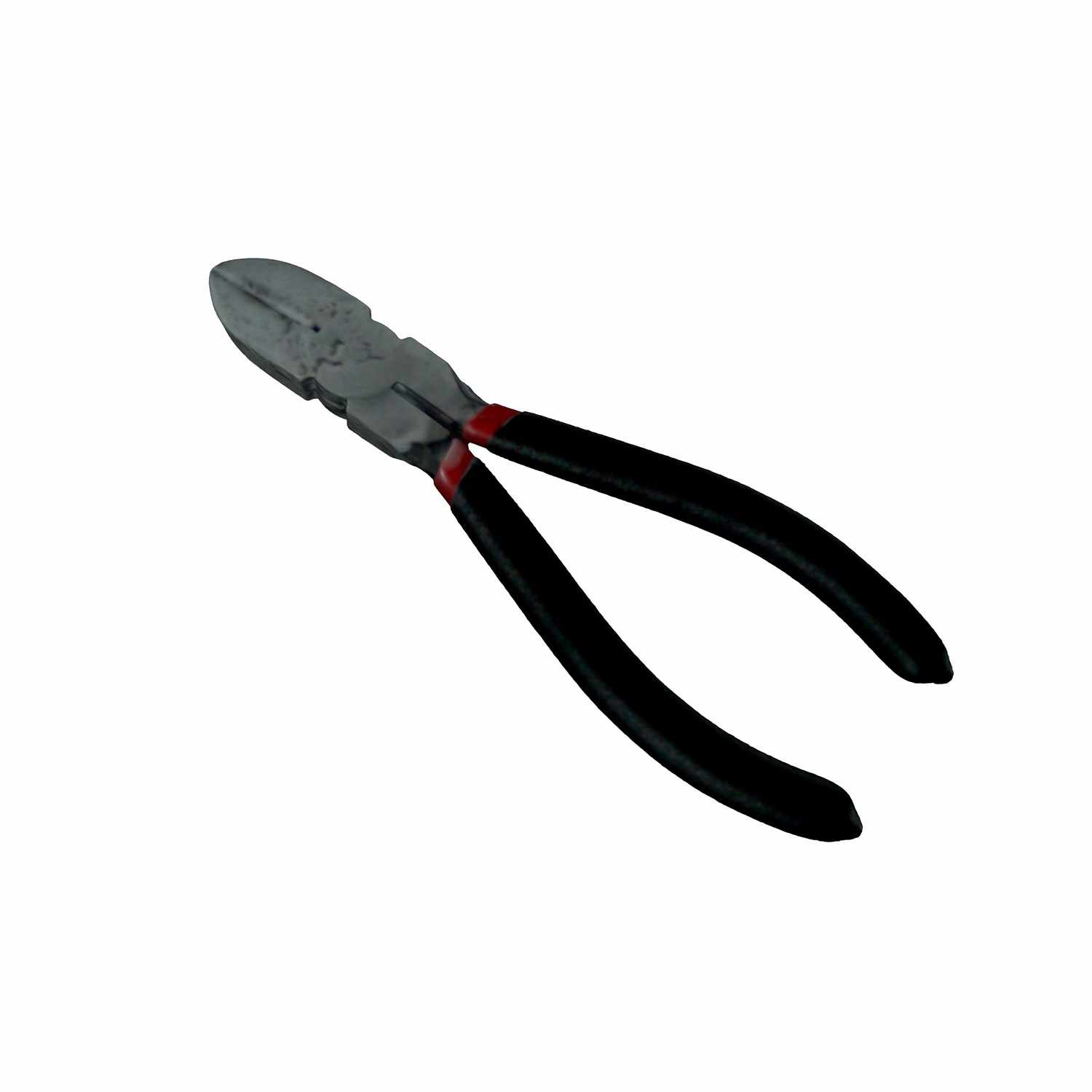}
}
\hspace{-0.01\linewidth}
\subfloat[]{
    \includegraphics[height=\objheight\linewidth, valign=t]{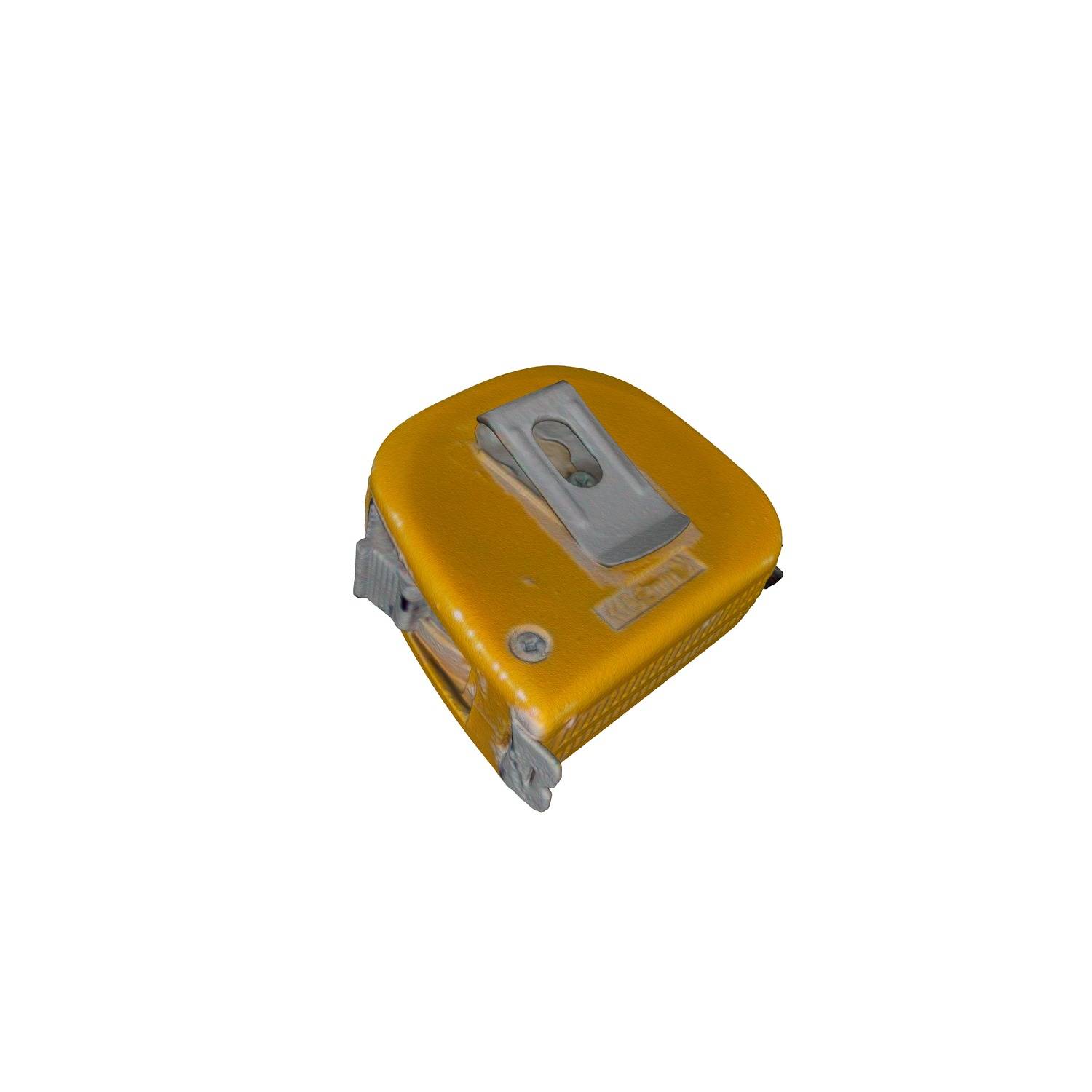}
}
\hspace{-0.01\linewidth}
\subfloat[]{
    \includegraphics[height=\objheight\linewidth, valign=t]{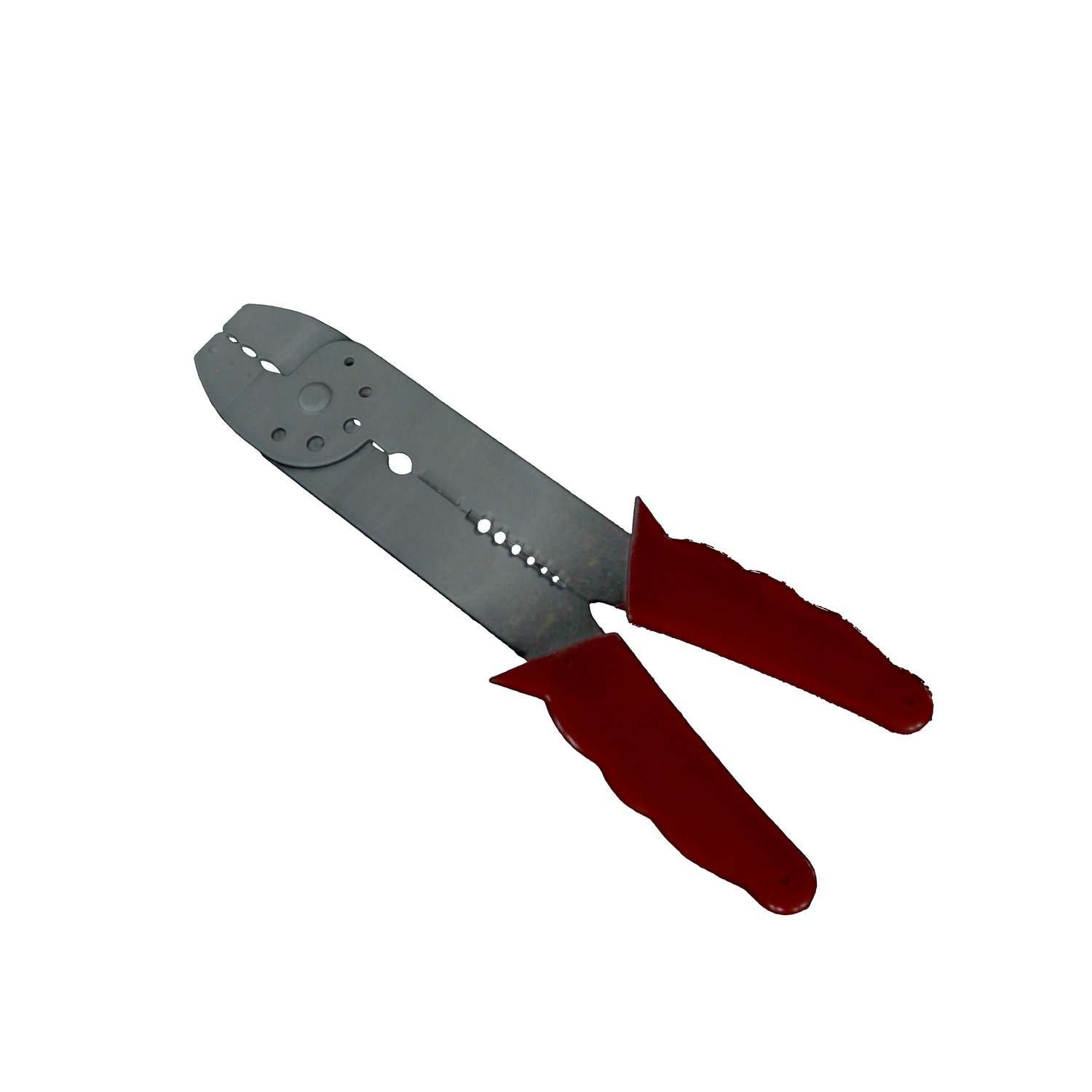}
}
\hspace{-0.01\linewidth}
\subfloat[]{
    \includegraphics[height=\objheight\linewidth, valign=t]{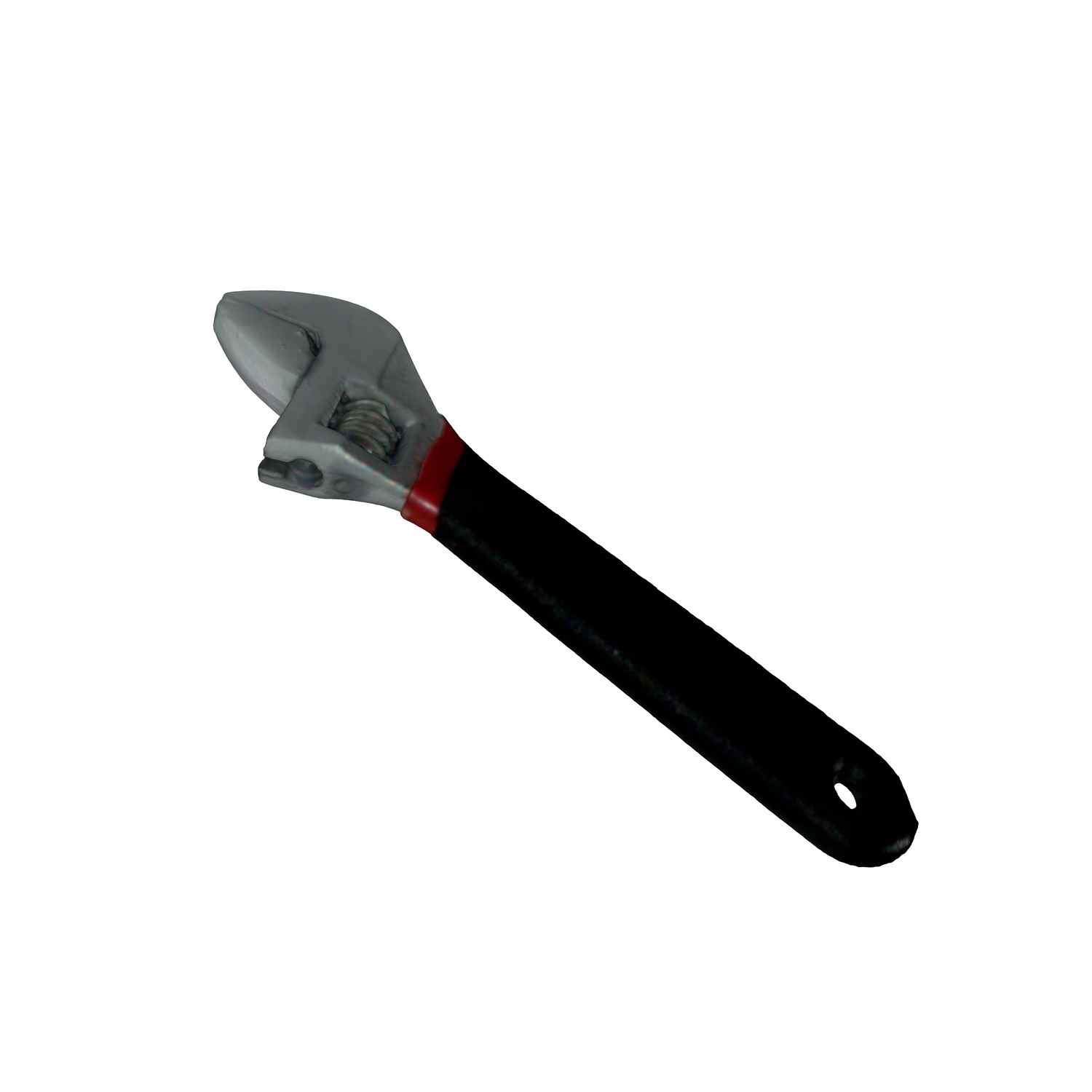}
}
\hspace{-0.01\linewidth}
\subfloat[]{
    \includegraphics[height=\objheight\linewidth, valign=t]{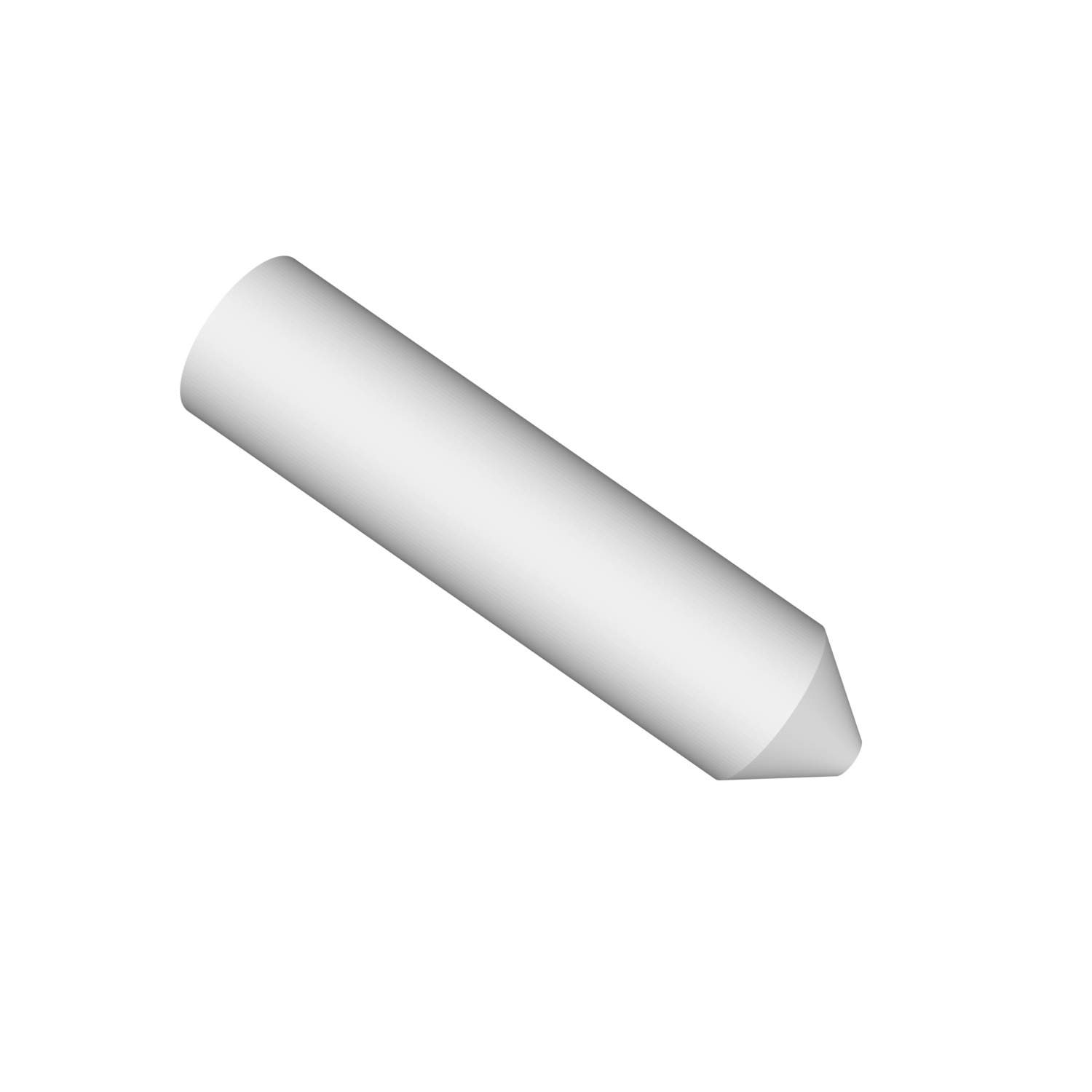}
}
\\
\vspace{-3ex}
\subfloat[]{
    \includegraphics[height=\objheight\linewidth, valign=t]{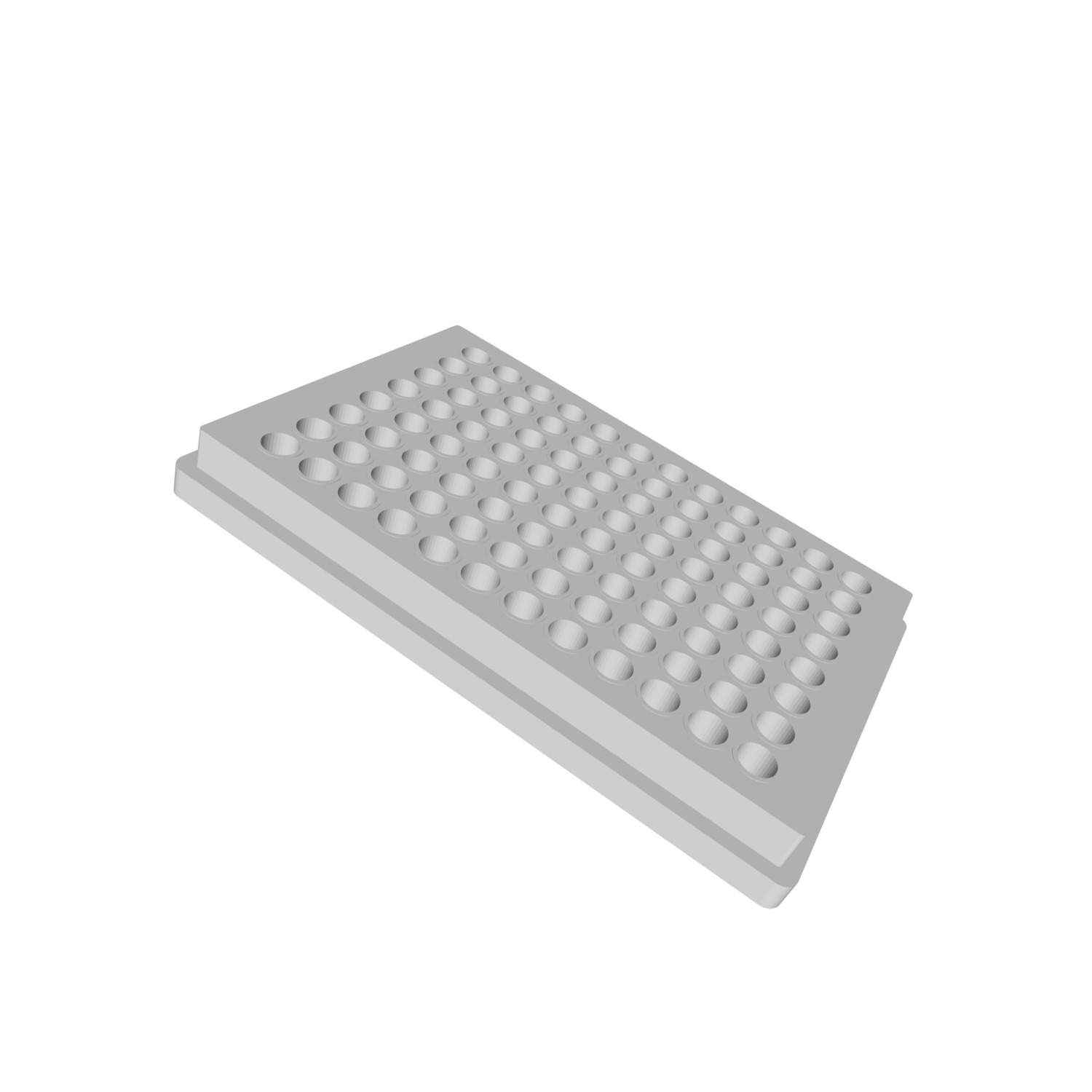}
}
\hspace{-0.01\linewidth}
\subfloat[]{
    \includegraphics[height=\objheight\linewidth, valign=t]{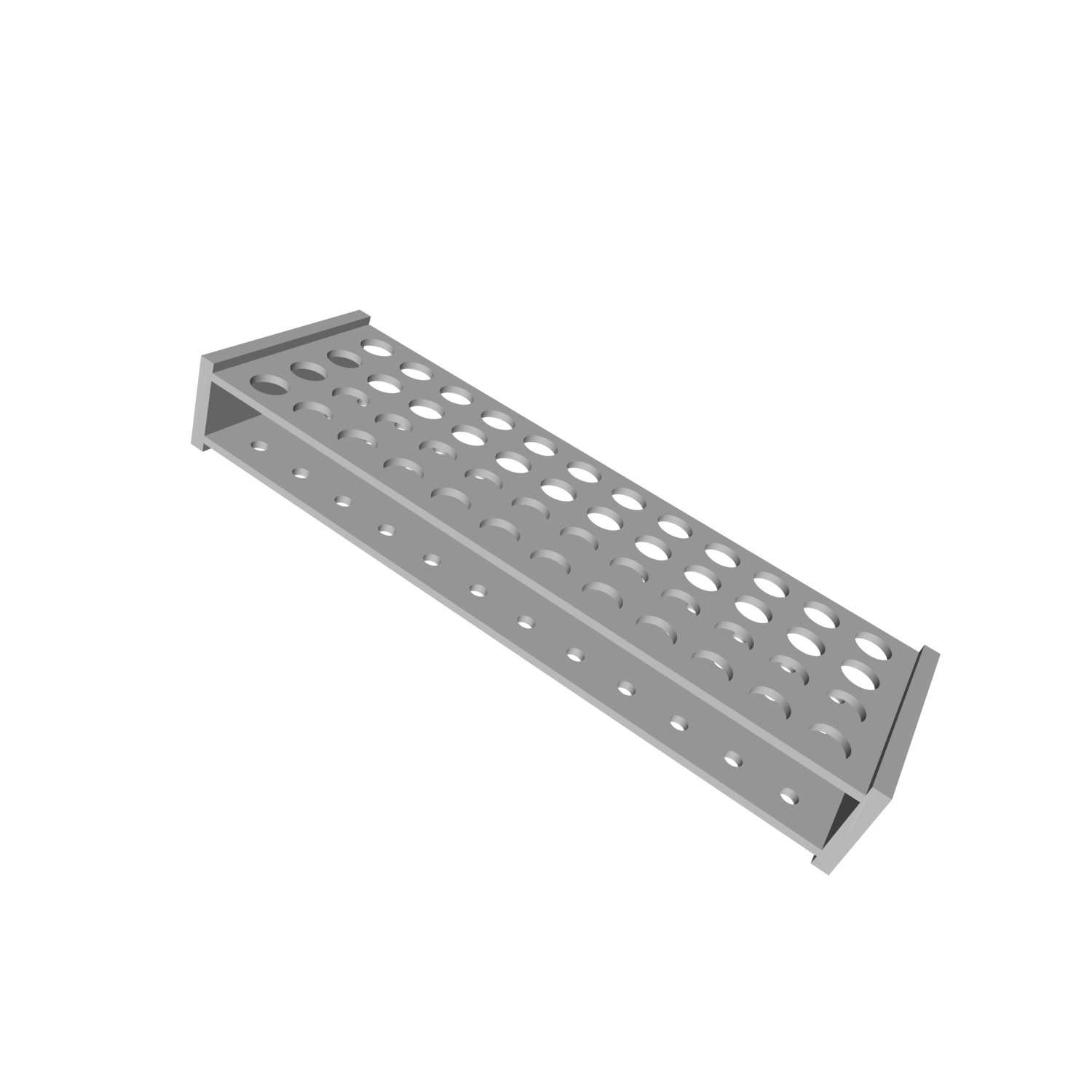}
}
\hspace{-0.01\linewidth}
\subfloat[]{
    \includegraphics[height=\objheight\linewidth, valign=t]{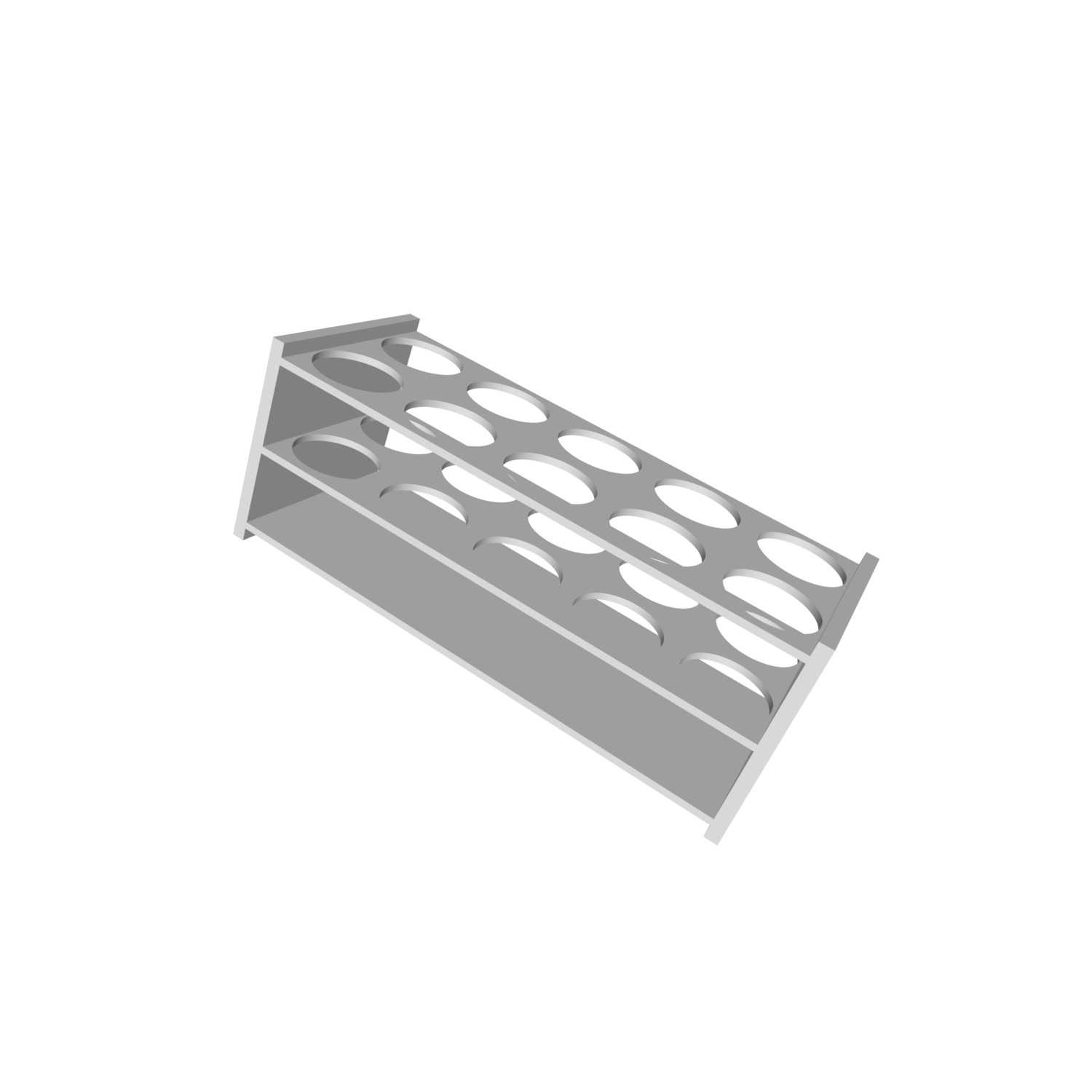}
}
\hspace{-0.01\linewidth}
\subfloat[]{
    \includegraphics[height=\objheight\linewidth, valign=t]{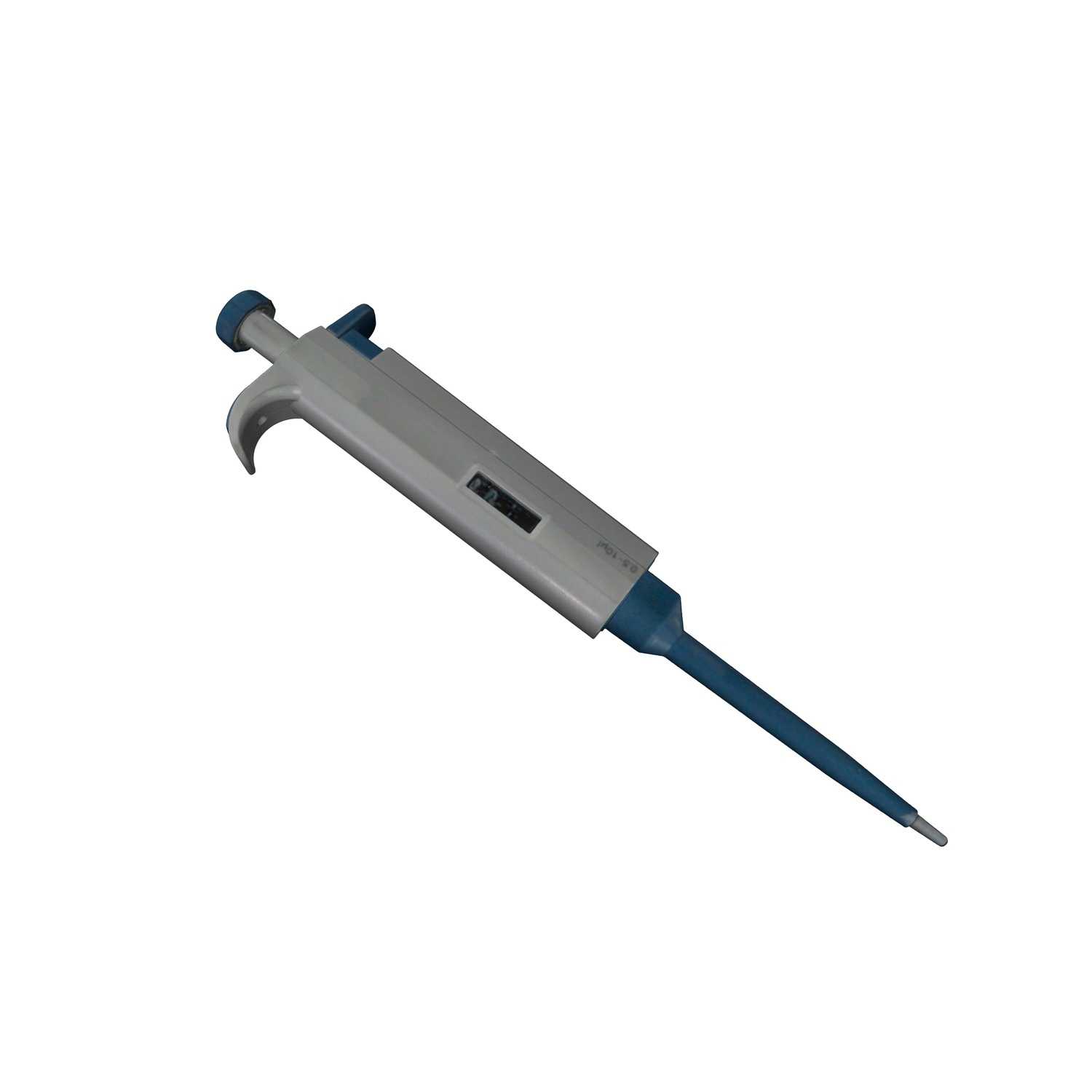}
}
\hspace{-0.01\linewidth}
\subfloat[]{
    \includegraphics[height=\objheight\linewidth, valign=t]{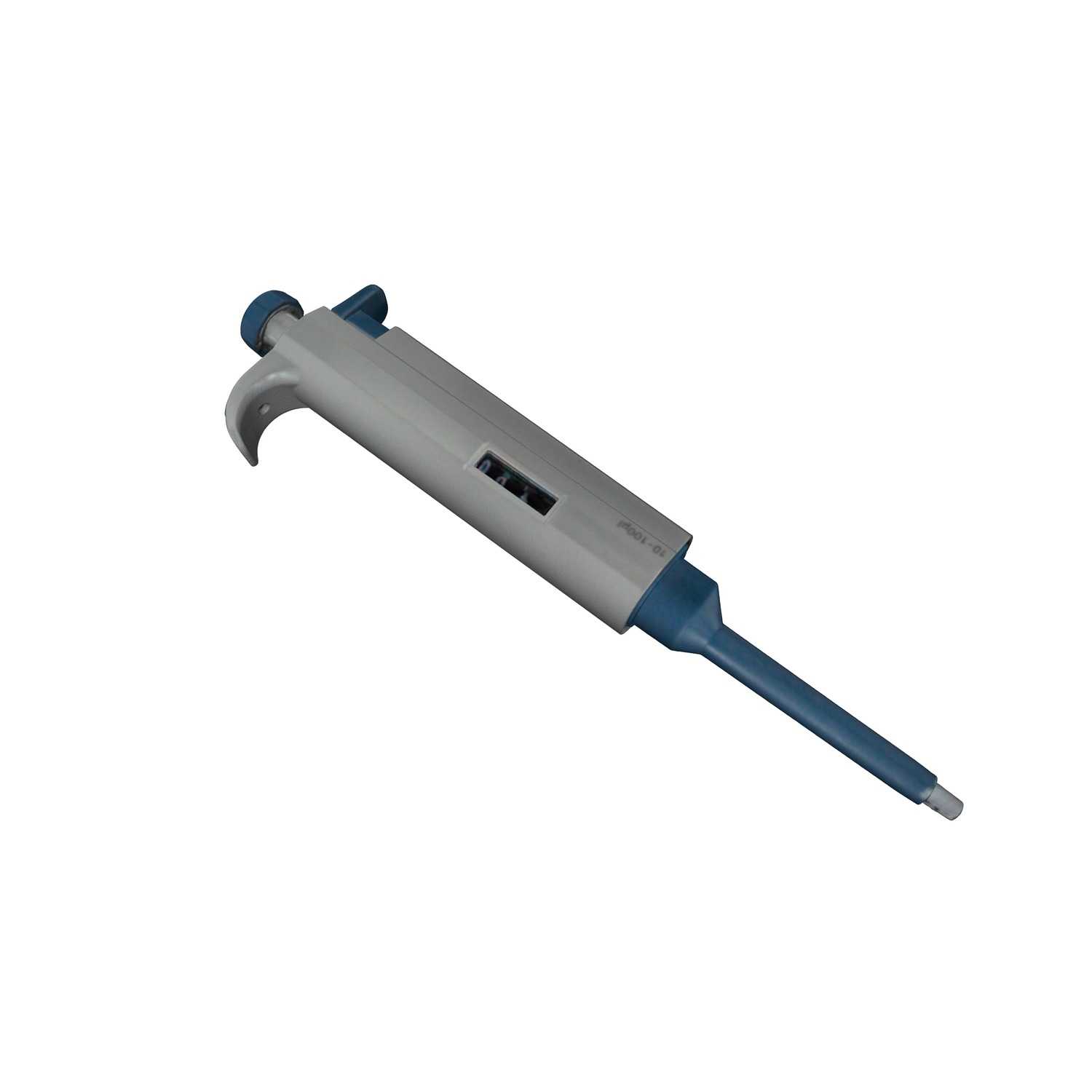}
}
\hspace{-0.01\linewidth}
\subfloat[]{
    \includegraphics[height=\objheight\linewidth, valign=t]{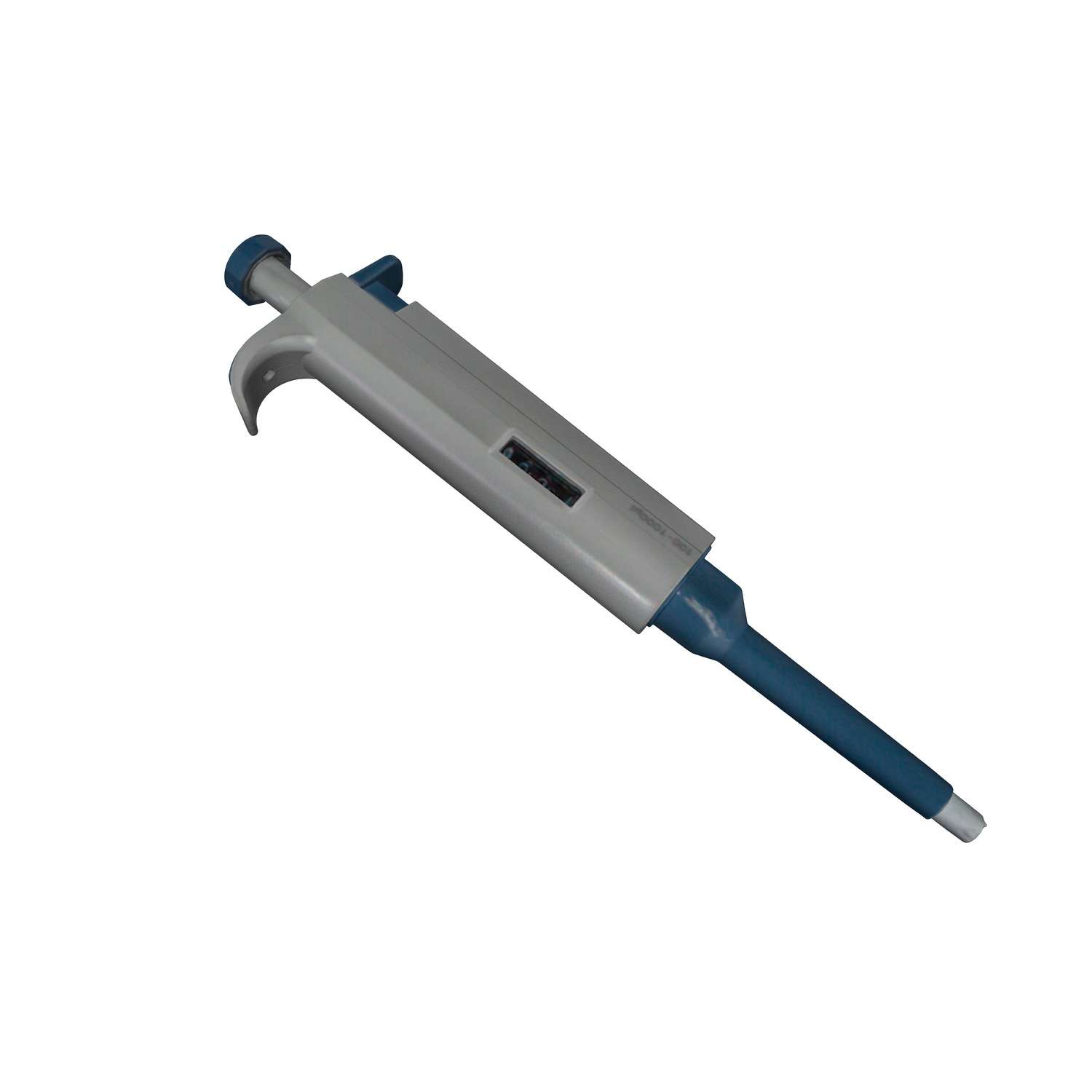}
}
\hspace{-0.01\linewidth}
\subfloat[]{
    \includegraphics[height=\objheight\linewidth, valign=t]{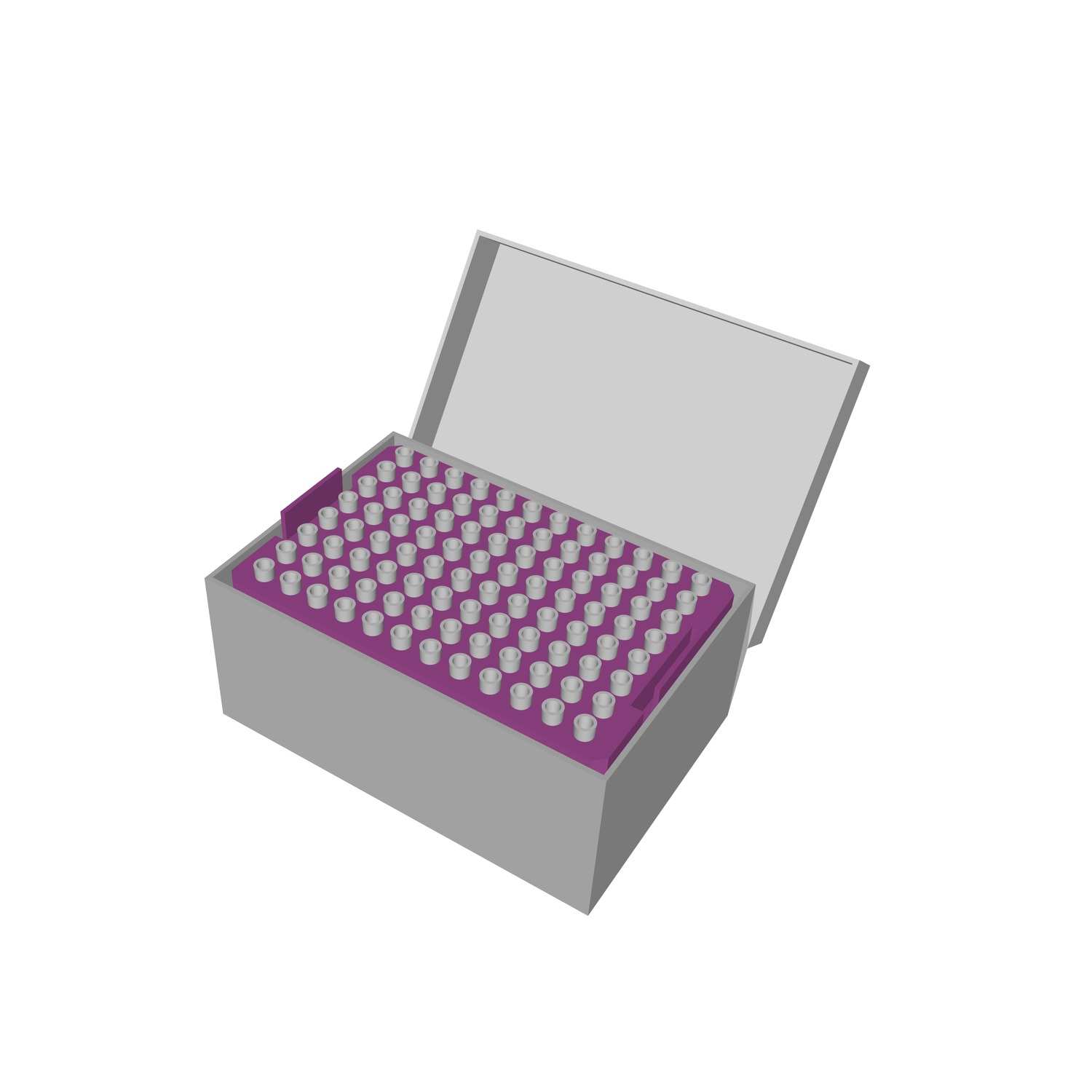}
}
\hspace{-0.01\linewidth}
\subfloat[]{
    \includegraphics[height=\objheight\linewidth, valign=t]{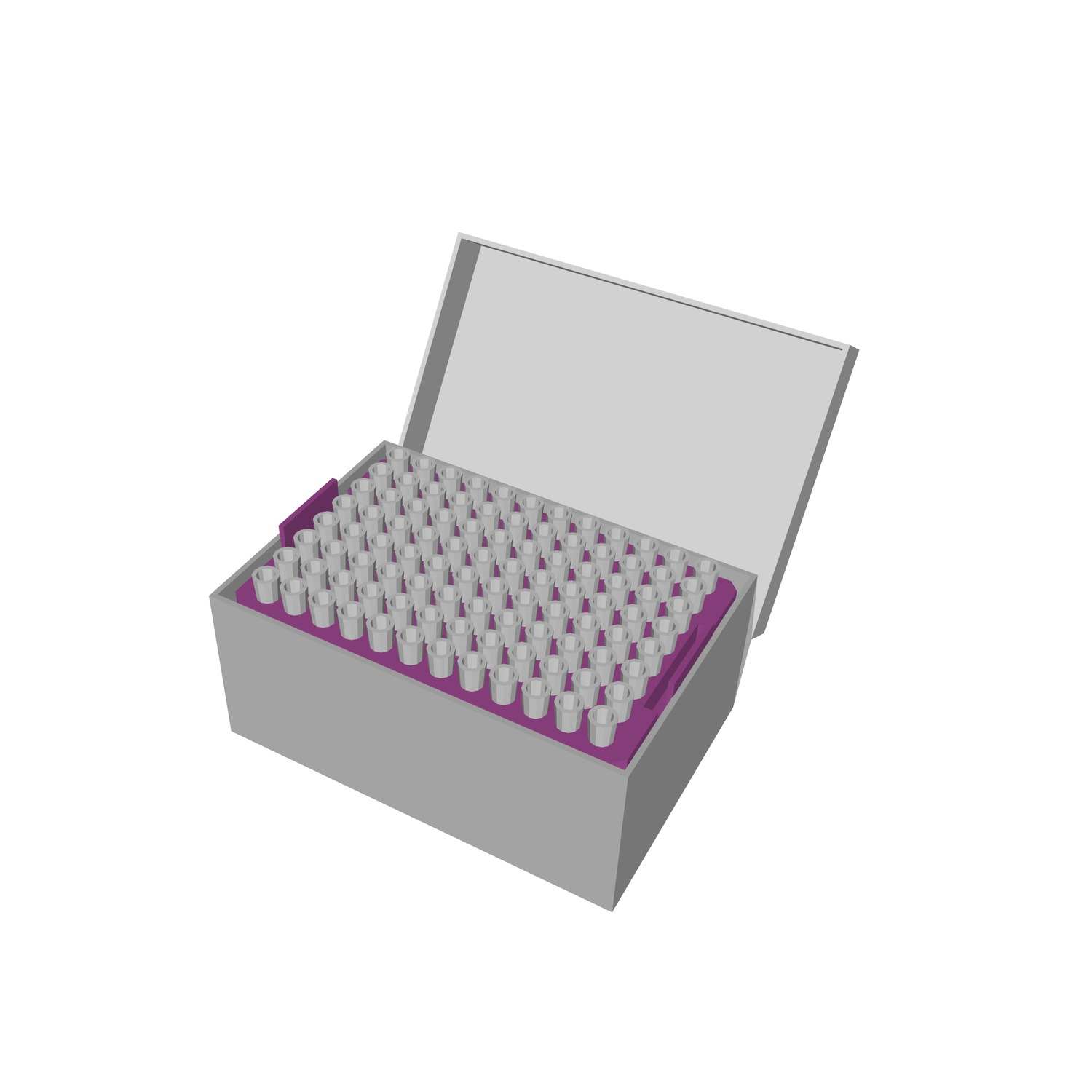}
}
\hspace{-0.01\linewidth}
\subfloat[]{
    \includegraphics[height=\objheight\linewidth, valign=t]{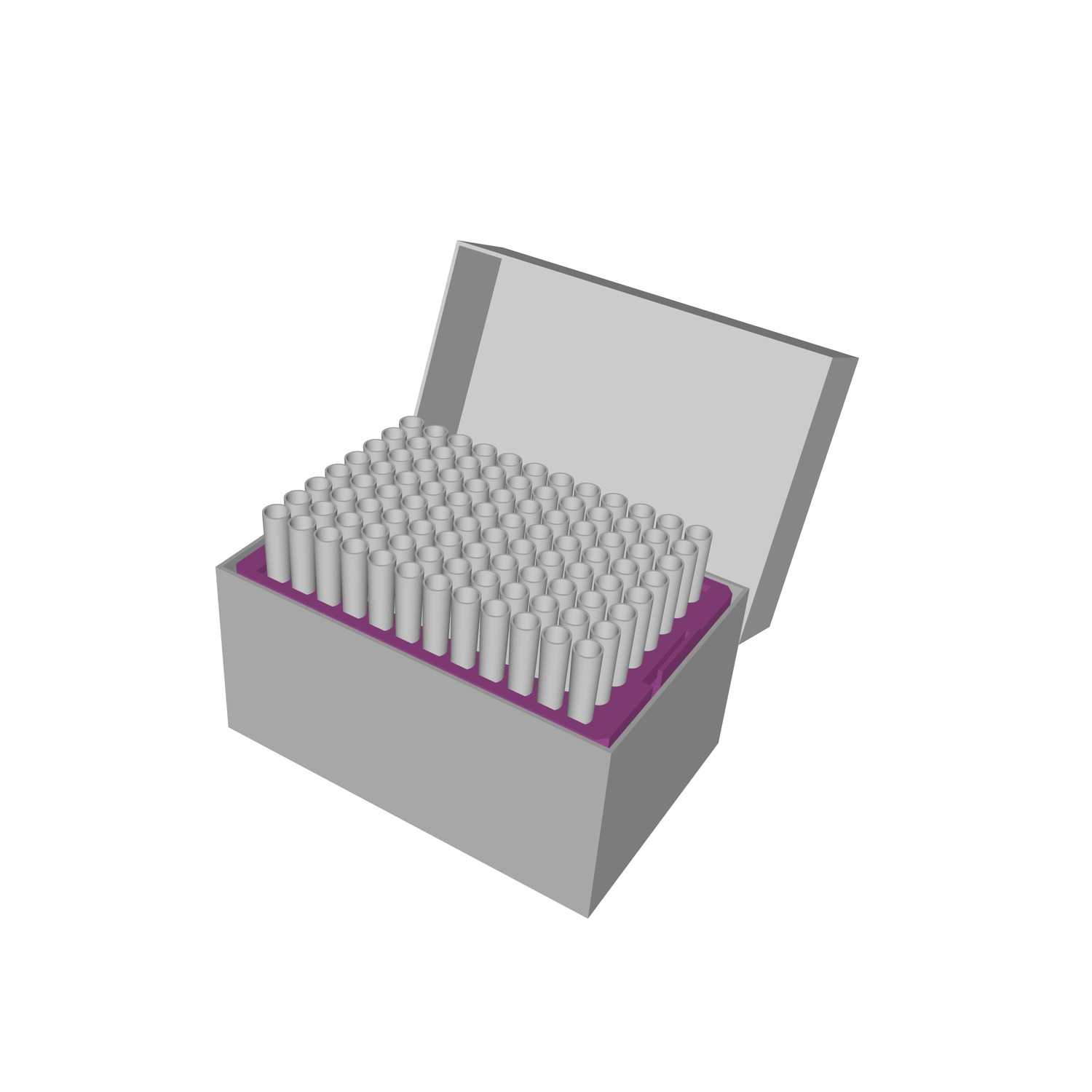}
}
\vspace{-1ex}
\caption{\textbf{3D CAD models of the 18 Objects in our StereOBJ-1M dataset. }
(a) \texttt{blade\_razor}; 
(b) \texttt{hammer}; 
(c) \texttt{needle\_nose\_pliers}
(d) \texttt{screwdriver}; 
(e) \texttt{side\_cutters}; 
(f) \texttt{tape\_measure};
(g) \texttt{wire\_stripper};
(h) \texttt{wrench}; 
(i) \texttt{centrifuge\_tube};
(j) \texttt{microplate};
(k) \texttt{tube\_rack\_2};
(l) \texttt{tube\_rack\_50};
(m) \texttt{pipette\_0.5\_10};
(n) \texttt{pipette\_10\_100}; 
(o) \texttt{pipette\_100\_1000}; 
(p) \texttt{sterile\_rack\_10};
(q) \texttt{sterile\_rack\_200};
(r) \texttt{sterile\_rack\_1000}.
During scanning, reflective parts of the objects are covered with white scanning spray.
Among the objects, (c)(d)(e)(g)(j)(k)(l) have discrete 2-fold rotational symmetry; (i) has continuous rotational symmetry.
}
\label{fig:obj}
\end{figure*}

\textbf{4. Pose Calculation for moving camera. }
Given the recorded stereo RGB video of length $L$, we use PnP algorithm again to calculate the poses of moving stereo camera in world coordinates $[\mathbf{R}^\text{M}_j, \mathbf{T}^\text{M}_j] \in \mathbb{R}^{3\times4}$ for every frame $j \in \{1,2,\ldots,L\}$ of the video, using the small fiducial marker locations $\{\textbf{x}^{\text{F}}_i\}$. 
To reduce the error of PnP, in practice only the frames with at least two small fiducial markers or eight corners detected are selected as valid frames.

\textbf{5. Keypoint Annotation. }
From all valid frames, we select a few to annotate the 2D locations of projected object keypoints on the images.
The frames are selected using farthest point sampling (FPS) such that their camera translations $\mathbf{T}^\text{M}_j$ are as far away from each other as possible.
The keypoints of an object are defined by experts and are easy to be spotted and accurately located, e.g. corners.
Note that it is possible that only a subset of the total keypoints are annotated in one particular frame.

\textbf{6. Keypoint Triangulation. }
For each keypoint of an object, we retrieve all frames in which the keypoint is annotated. 
Using the moving camera pose $[\mathbf{R}^\text{M}_j, \mathbf{T}^\text{M}_j]$ and the 2D annotations, the 3D location of the keypoints in the world coordinate can be calculated by multi-view triangulation.

\textbf{7. Pose Fitting. }
To obtain the 6D poses of the objects in the world coordinate, we solve an Orthogonal Procrustes problem \cite{procrustes} to fit the object CAD models to the annotated 3D keypoints.
Finally, the object poses are propagated to all valid frames via an inverse transform of the camera pose $[ \mathbf{R}^\text{M}_j, \mathbf{T}^\text{M}_j]$.

\begin{savenotes}

\begin{table}[t]
\small
\setlength{\tabcolsep}{5pt}
\centering

\begin{tabular}{l|c|c|c}
\hline
dataset & RGBD datasets & TOD \cite{keypose} & \textbf{StereOBJ-1M} \\
\hline
3D labeling & depth map & \multicolumn{2}{c}{multi-view}  \\
\hline
labeling error & $\ge$ 1.7cm\footnote{\label{footnote:kinect}One of the most recent and advanced commercial depth sensors, Microsoft Azure Kinect, has random depth sensing error of 17mm in standard deviation: \url{https://docs.microsoft.com/en-us/azure/kinect-dk/hardware-specification}} & 0.34cm & \textbf{0.23cm} \\
\hline
\end{tabular}
\vspace{-1ex}
\caption{Labeling error measured in 3D RMSE.}
\label{tab:label:error}
\vspace{-2ex}
\end{table}
\end{savenotes}

\subsection{Labeling Error Analysis}\label{sec:labeling:error:analysis}

An intriguing question that needs to be answered is: how accurate is our labeling method?
We assume the error of the dimensions of the large fiducial marker array board and the small fiducial markers are negligible since they are both accurately measured by vernier caliper.
Then the labeling error can come from two steps: automatic detection of small fiducial marker boards and the annotation of keypoint 2D locations, which contributes to the error in two nonlinear optimizations respectively: camera pose estimation and 3D point estimation from multiple views.

We use Monte Carlo simulation to quantify the pose annotation error with a similar procedure as in \cite{keypose}.
Specifically, we dither the keypoint 2D projections according to the keypoint re-projection RMSE statistics and estimate the 3D keypoint error as an approximate of the labeling error.
We report keypoint label error of 2.3mm RMSE as illustrated in Table \ref{tab:label:error}.
The reasons for label error improvement over \cite{keypose} are two folds. 
First, our stereo camera has a higher resolution than \cite{keypose} and allows more accurate labeling in 2D. 
Second, our object scanning paths are determined by human data collectors on the fly instead of being hard-coded and performed by robot \cite{keypose}, thus are more flexible and can adapt to specific scenes to cover more viewpoints and provide wider baselines for triangulation.

\subsection{Comparison to Previous Labeling Methods}

We point out that the idea of using multi-view and keypoints for pose labeling can also be found in human pose estimation scenarios such as the Panoptic Studio dataset~\cite{panoptic:studio}.
Unlike \cite{panoptic:studio} which relies on 480 fixed cameras mounted in a specially constructed studio for triangulation, our data acquisition method is affordable and portable --- it only requires three cameras and two tripods, and can therefore be deployed in diverse indoor and outdoor environments.
On the contrary, to construct datasets such as \cite{graspNet-1billion, keypose}, a studio equipped with multiple sensors or robot assists has to be specially constructed.
In addition to the logistic cost, such settings are not flexible enough for environments in the wild and therefore suffer from the lack of diversity of data.

TOD \cite{keypose} is the first object pose estimation dataset that provides stereo RGB modality. 
Our data capturing pipeline is different from \cite{keypose} in terms of moving camera pose calculation. 
Datasets such as  \cite{keypose, linemod, t-less} rely on customized board printed with fiducial markers, and objects are placed near the center of the board.
Thus only the simplest planar terrain can be used with the objects and lacks diversity.
Instead, in our data pipeline, we distribute small fiducial markers into the scene and calculate their locations on the fly with the help of two static cameras.
This allows the objects to be placed in more flexible and complex background terrains.

Our data pipeline has a much higher data efficiency than TOD \cite{keypose}.
With the proposed data pipeline, in each constructed scene, we can capture and annotate more than 2,000 valid frames with a single scan.
As a comparison, \cite{keypose} only captures 80 frames per scene with the help of a robot arm.
An explanation is that in \cite{keypose}, the predefined automatic scanning path of the robot is limited by its operational space.
In our data pipeline, the scanning is performed by humans and can adapt to different scenes,
which results in (1) more valid frames per video; (2) larger coverage of viewpoints; and (3) wider baseline during triangulation and therefore higher precision.

\section{StereOBJ-1M Dataset}

With the proposed method, we construct StereOBJ-1M, a large-scale dataset, and benchmark for 3D object pose estimation from stereo RGB images.
In this section, we provide technical details to our StereOBJ-1M dataset in terms of object 3D models and data sample illustration.

\subsection{Objects in Dataset}

\begin{figure*}[ht] 
\centering
\includegraphics[width=\linewidth]{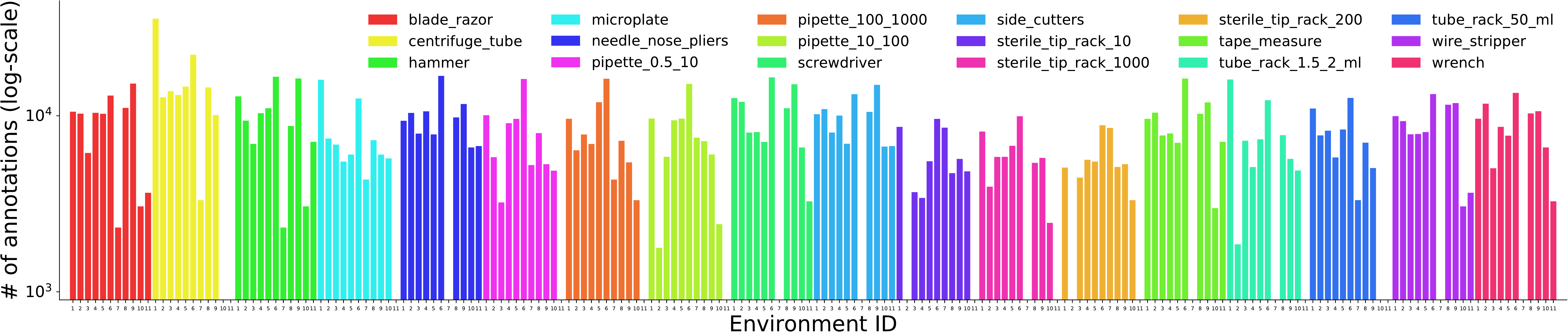} 
\vspace{-4ex}
\caption{ The total number of annotations of each object in each environment.
Environment IDs range from 1 to 11.
}
\label{fig:environment:distr}
\vspace{-1ex}
\end{figure*}

\begin{figure}[t] 
\centering
\includegraphics[width=0.95\linewidth]{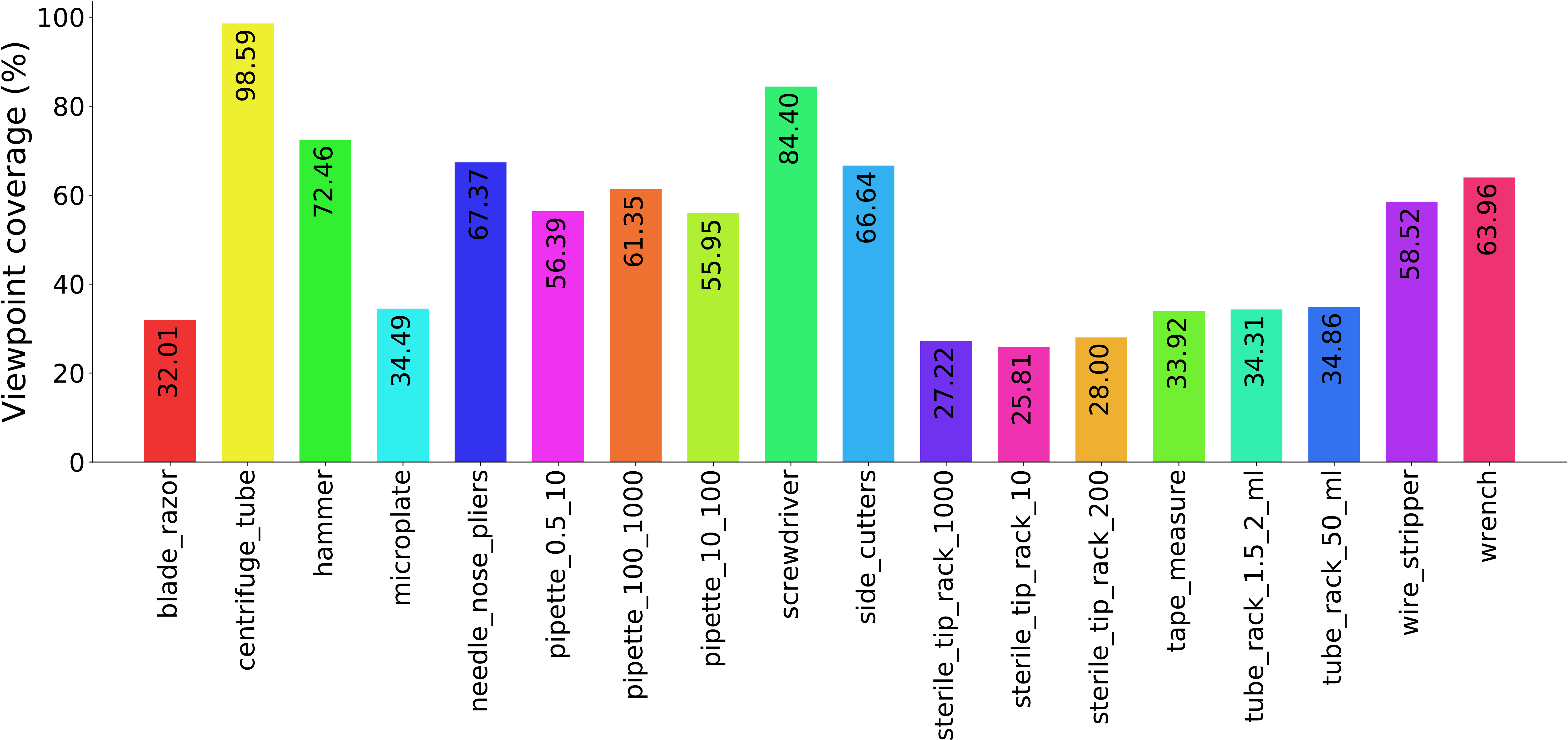} 
\vspace{-2ex}
\caption{ Overall viewpoint coverage percentage of all objects in StereOBJ-1M. 
}
\label{fig:viewpoint:coverage}
\vspace{-1ex}
\end{figure}

There are 18 objects included in our dataset.
Among them, 10 objects are plastic tools used in biochemical labs and 8 objects are metal mechanics tools, which together include both transparent and reflective instances.
We provide 3D CAD models of the 18 objects as illustrated in Figure \ref{fig:obj}.
The CAD models are obtained using a high-precision EinScan Pro 2X Plus scanner \cite{einscan} which has a scan accuracy of 0.04mm.
During scanning, reflective metallic parts of the objects are covered with white scanning spray.
Among the 18 objects, 
there are 8 objects with discrete 2-fold rotational symmetry and one with continuous rotational symmetry.
Among the 18 objects,
\texttt{microplate}, 
\texttt{tube\_rack\_2ml} and
\texttt{tube\_rack\_50ml} are transparent;
\texttt{centrifuge\_tube},
\texttt{sterile\_rack\_10ml},  
\texttt{sterile\_rack\_200ml} and
\texttt{sterile\_rack\_1000ml} are translucent.

The set of the objects used in our dataset has a special feature: it includes visually similar but different object instances.
For example, as illustrated in Figure \ref{fig:obj}, the three pipettes are almost identical in their geometric features.
In our dataset, we include image sequences where two or more similar but different object instances are present in the same scene.
Thus it poses a new research question for the computer vision community: \textit{how to detect and estimate the poses of visually very similar but different objects?}
We expect that this question can be studied with our dataset.

\subsection{Data Collection and Annotations}

We collected the data in  8 real-life indoor environments, including desktop, washbasin, wooden floor etc.
In addition to the indoor environments, we adopt 3 \emph{outdoor} environments to enrich the diversity in background scenes.
In each environment, we shuffle the objects and occlusion clutters several times to construct multiple scenes.
In total, we constructed 182 scenes.
A stereo video was recorded in every constructed scene.
The lengths of the video range from 2 to 7 minutes.
When sampled at 15 frames/sec, the recorded videos yield 393,612 stereo frames in total.
On average, there are more than 2,100 stereo frames in every scene.
Our dataset consists of 182 videos and contains over 1.5 million object pose annotations.
The number of annotations of each object in each environment is illustrated in Figure \ref{fig:environment:distr}.

\begin{figure*}[ht]
\newcommand\sampleheight{0.158}
\centering
\small
\subfloat{
    \includegraphics[height=\sampleheight\linewidth, valign=t]{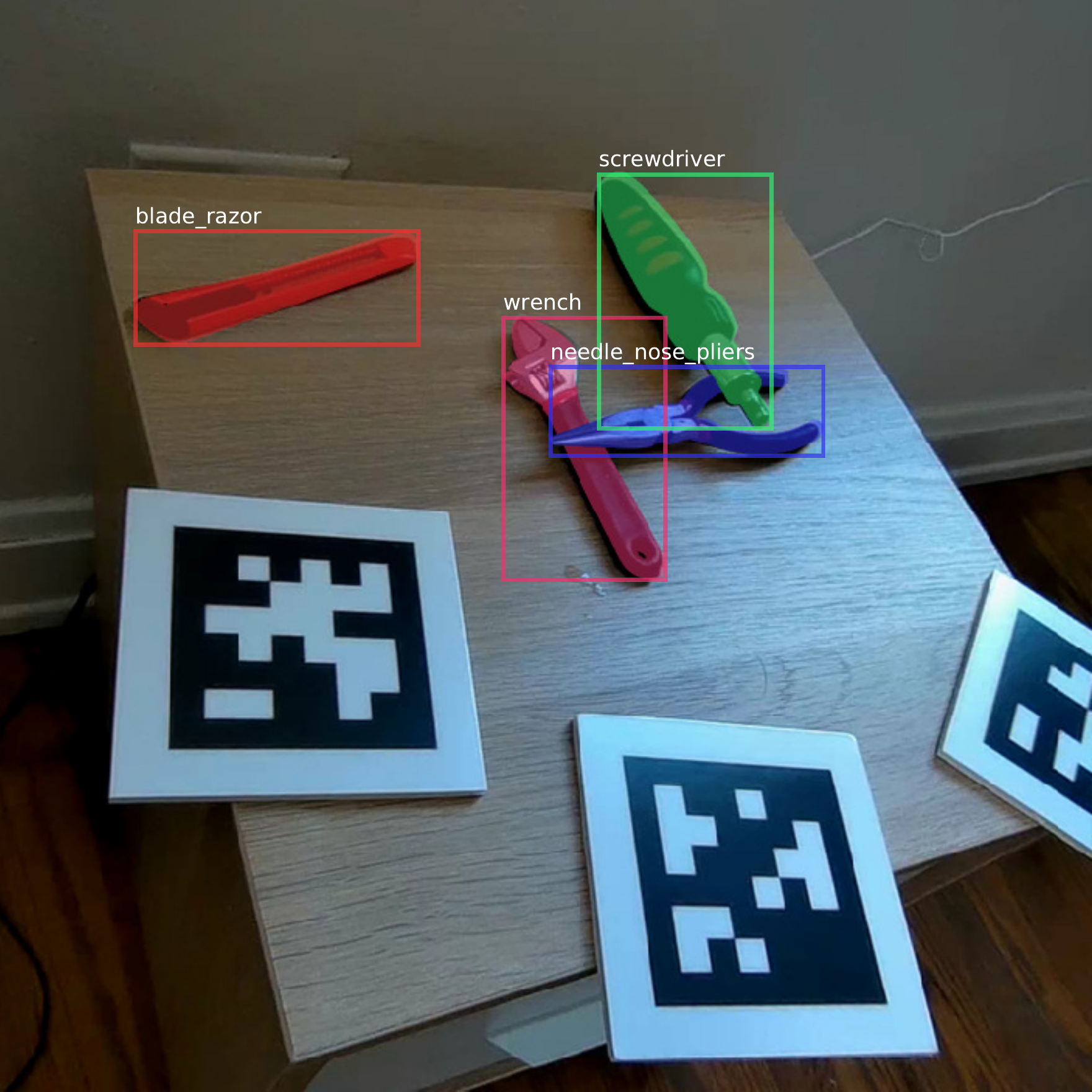}
}
\subfloat{
    \includegraphics[height=\sampleheight\linewidth, valign=t]{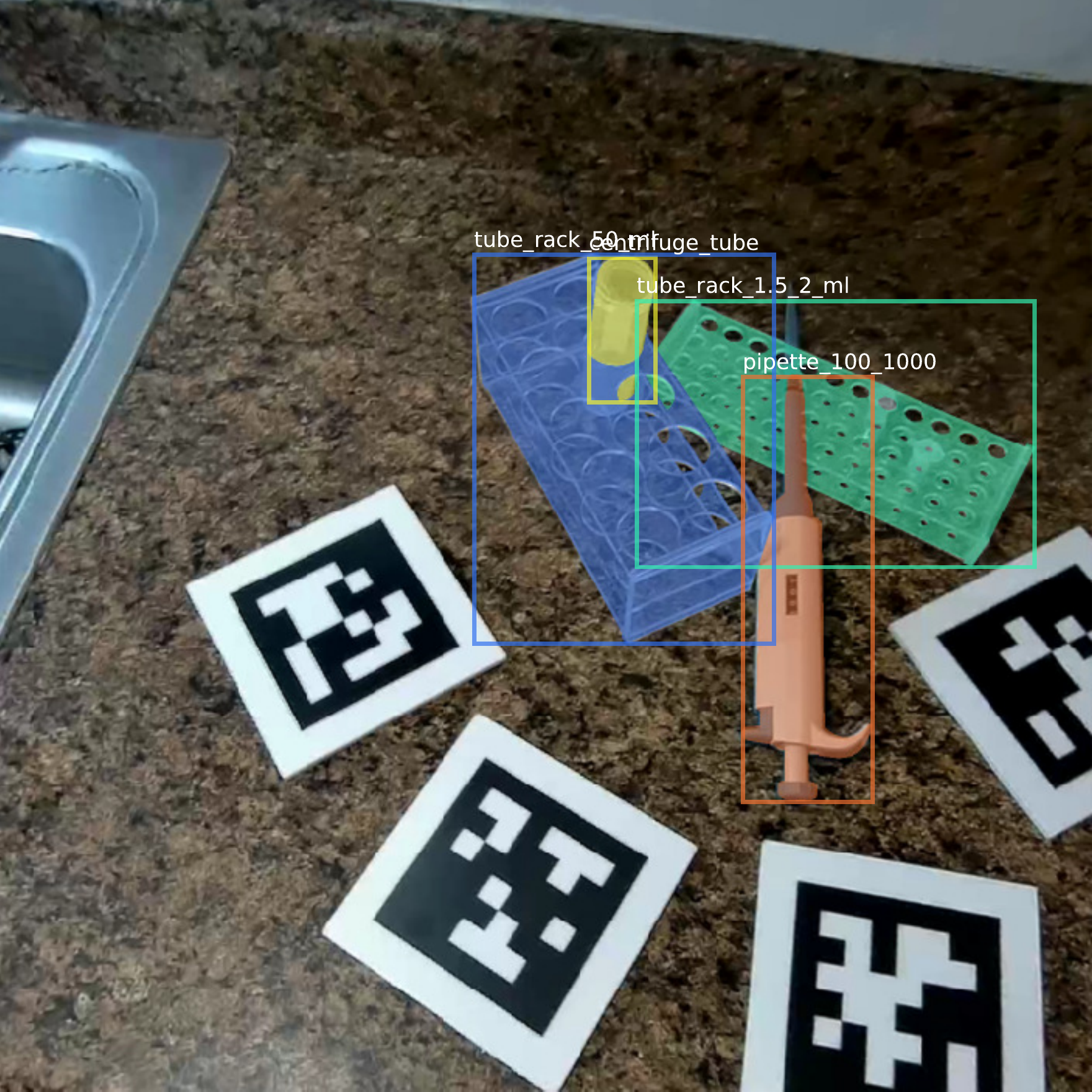}
}
\subfloat{
    \includegraphics[height=\sampleheight\linewidth, valign=t]{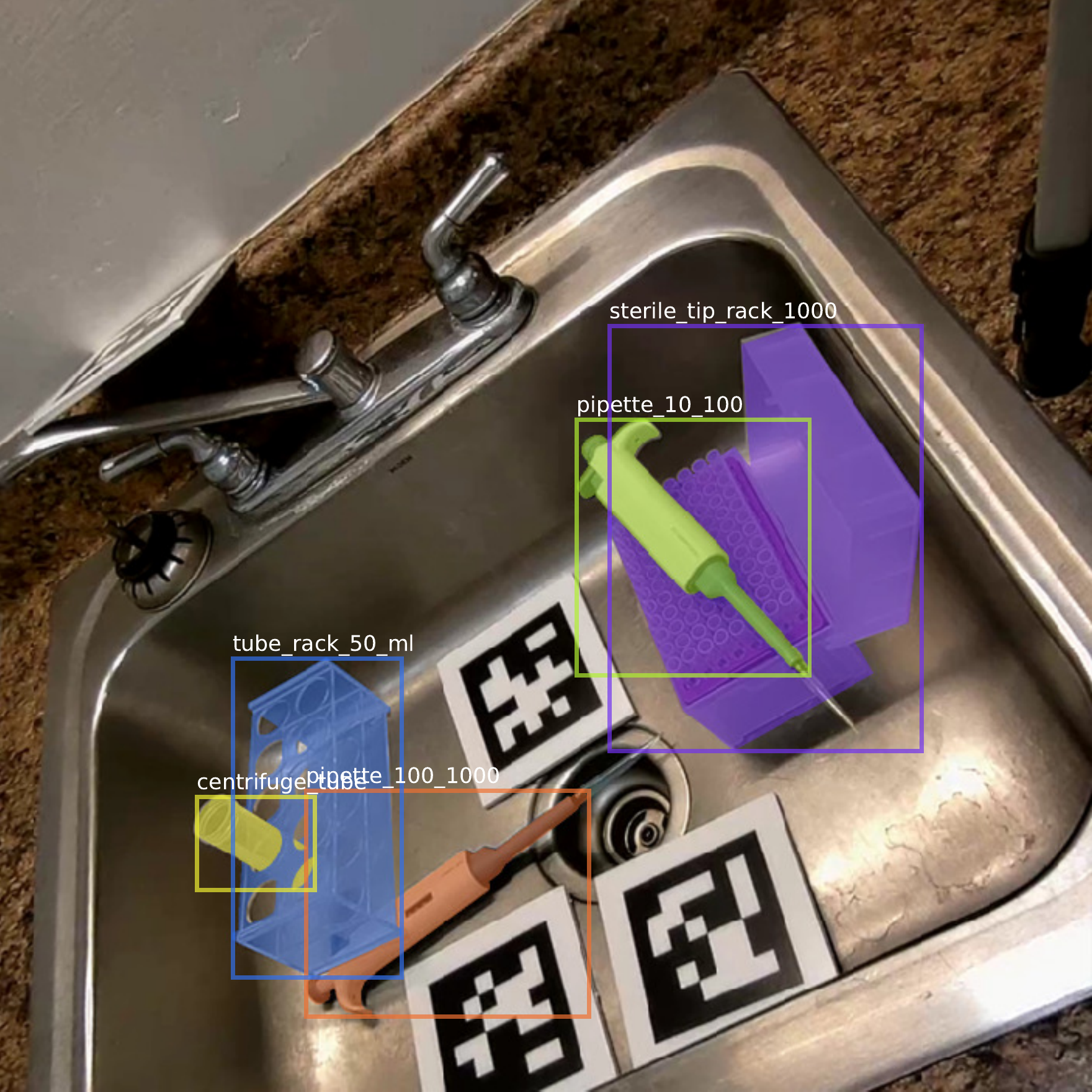}
}
\subfloat{
    \includegraphics[height=\sampleheight\linewidth, valign=t]{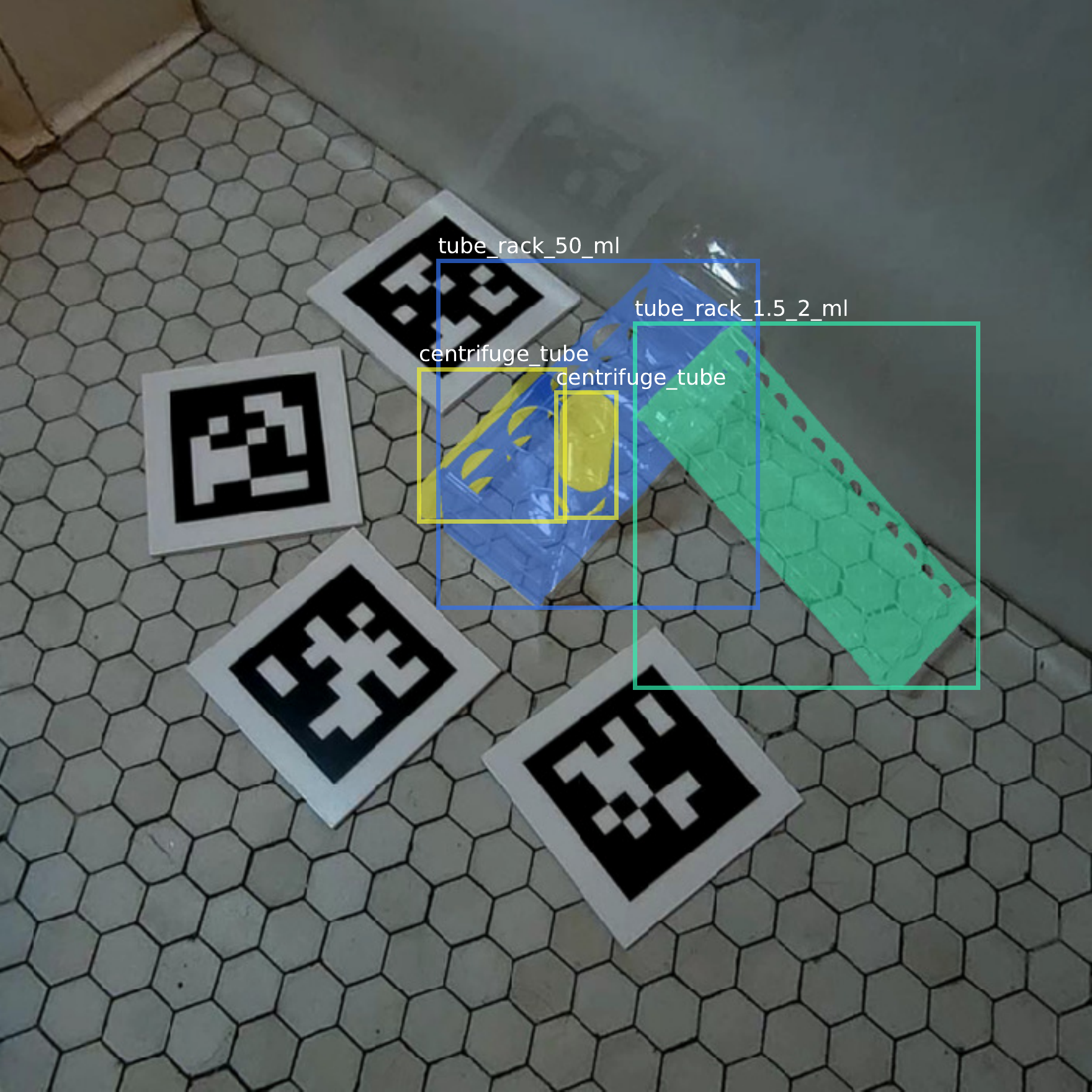}
}
\subfloat{
    \includegraphics[height=\sampleheight\linewidth, valign=t]{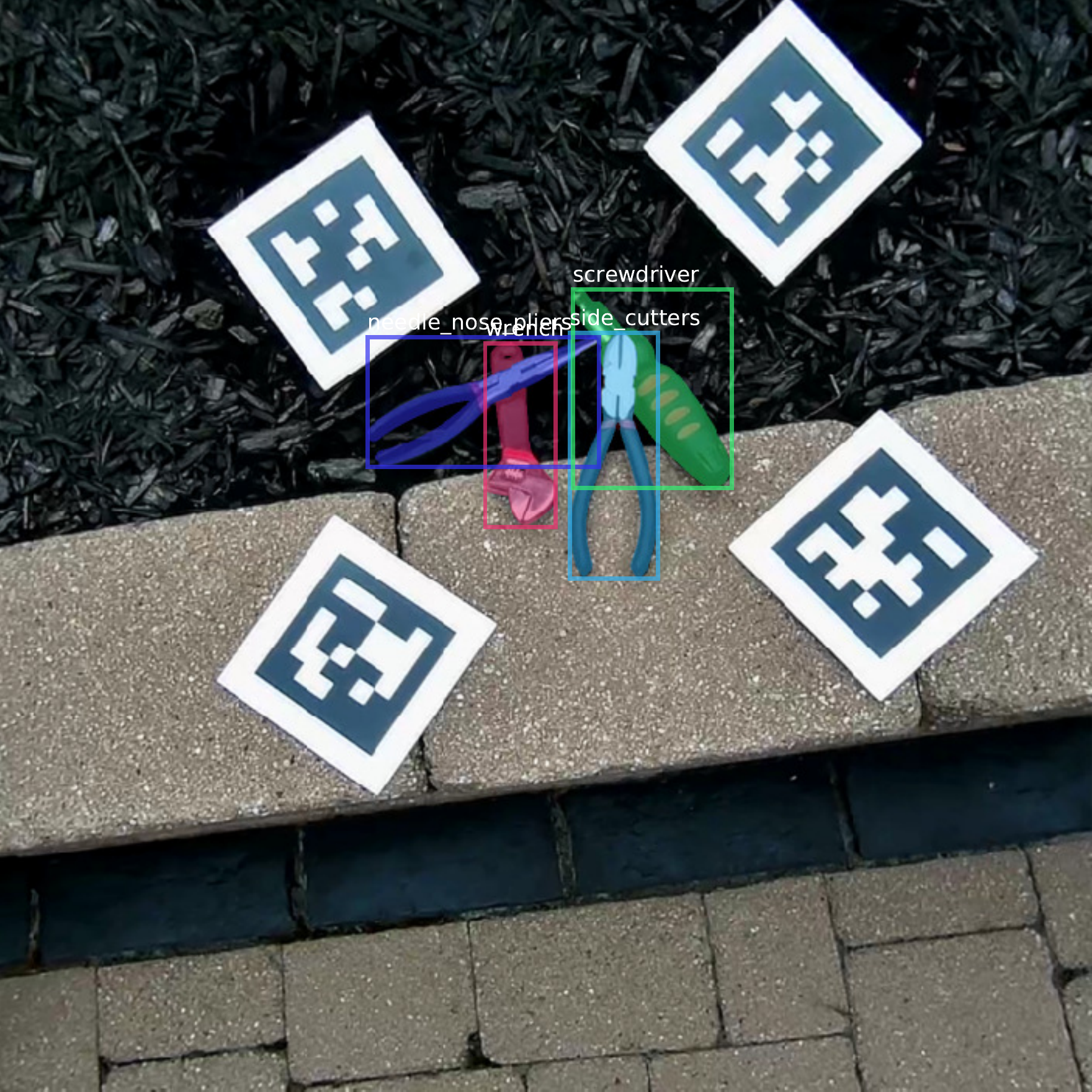}
}
\subfloat{
    \includegraphics[height=\sampleheight\linewidth, valign=t]{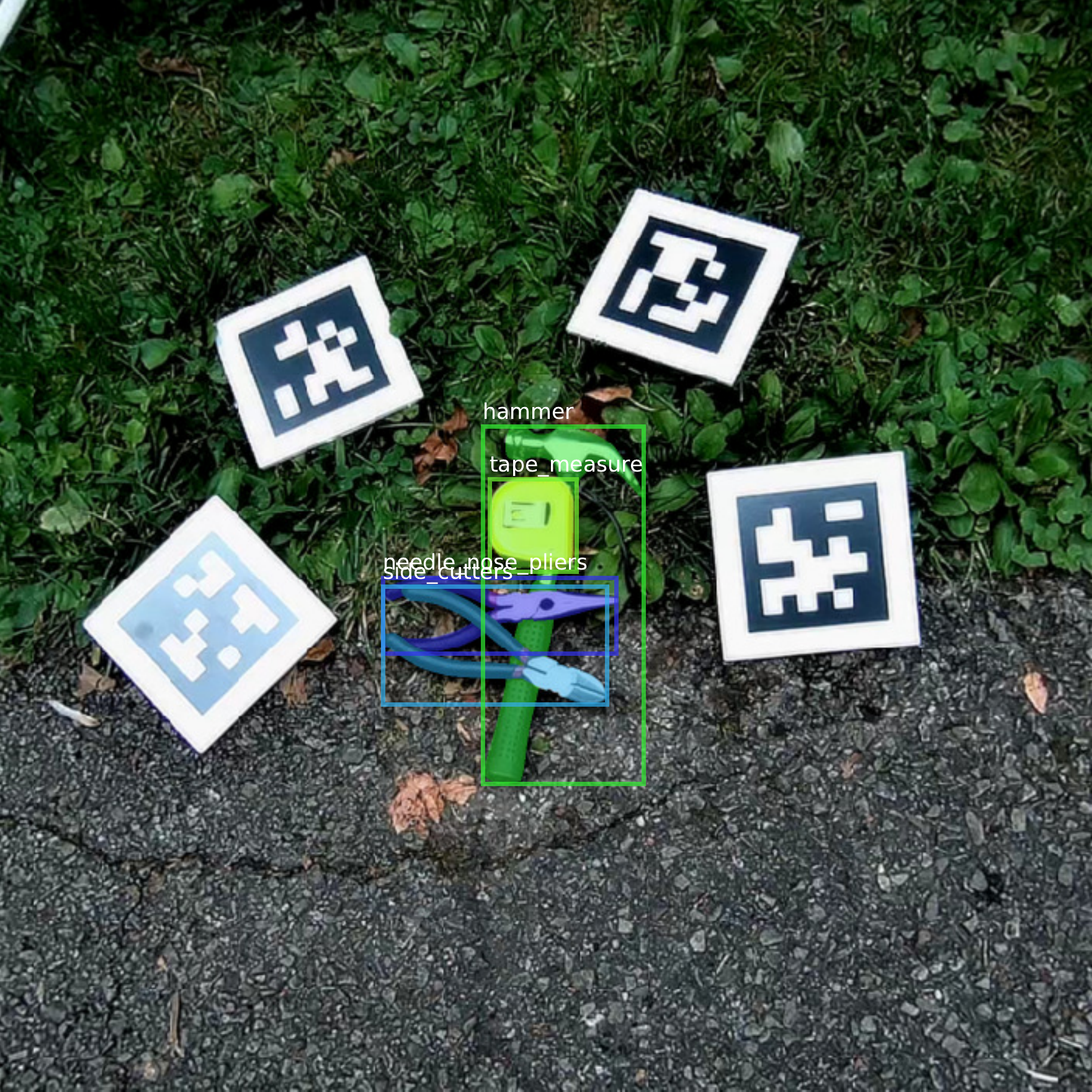}
}
\\
\vspace{-1.5ex}
\subfloat{
    \includegraphics[height=\sampleheight\linewidth, valign=t]{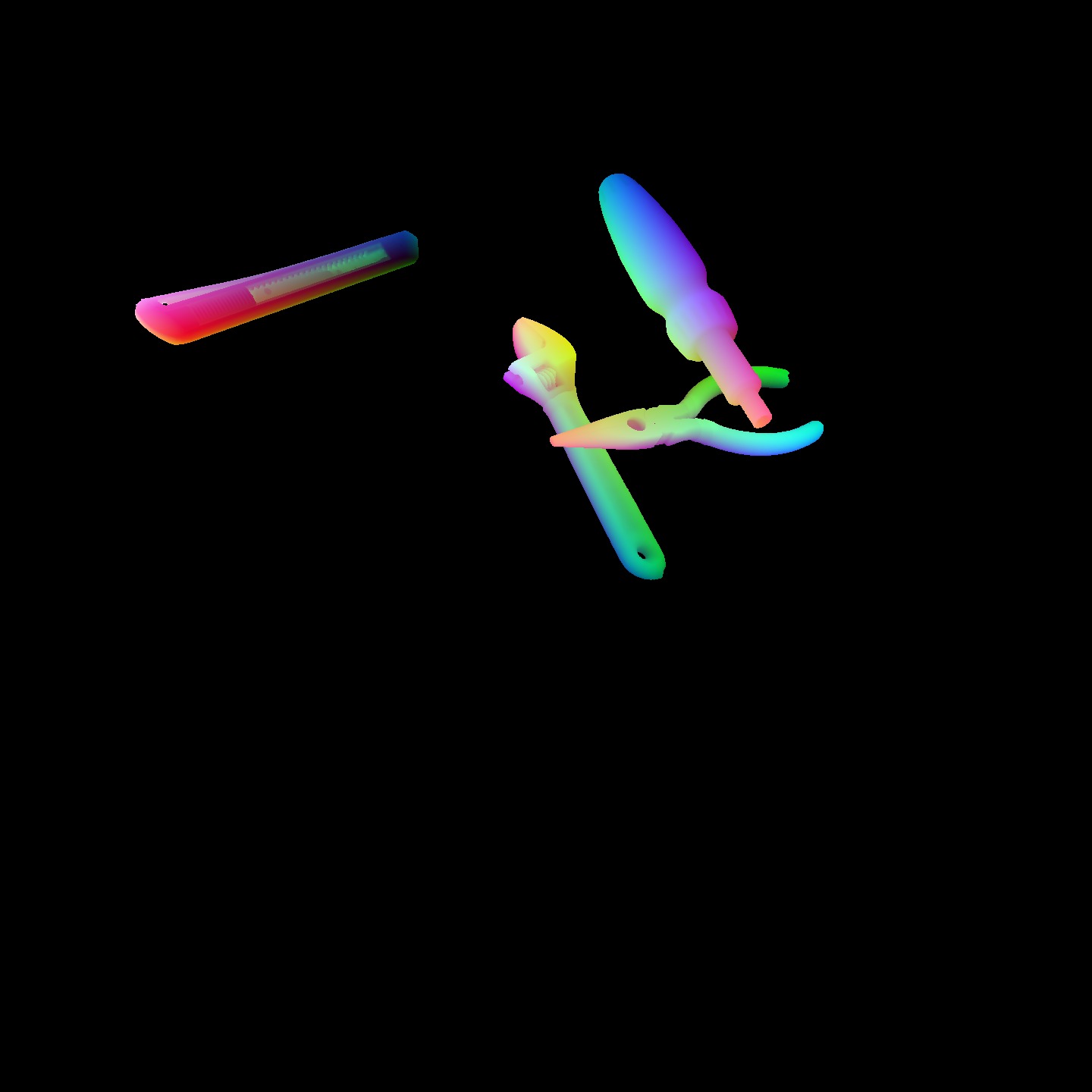}
}
\subfloat{
    \includegraphics[height=\sampleheight\linewidth, valign=t]{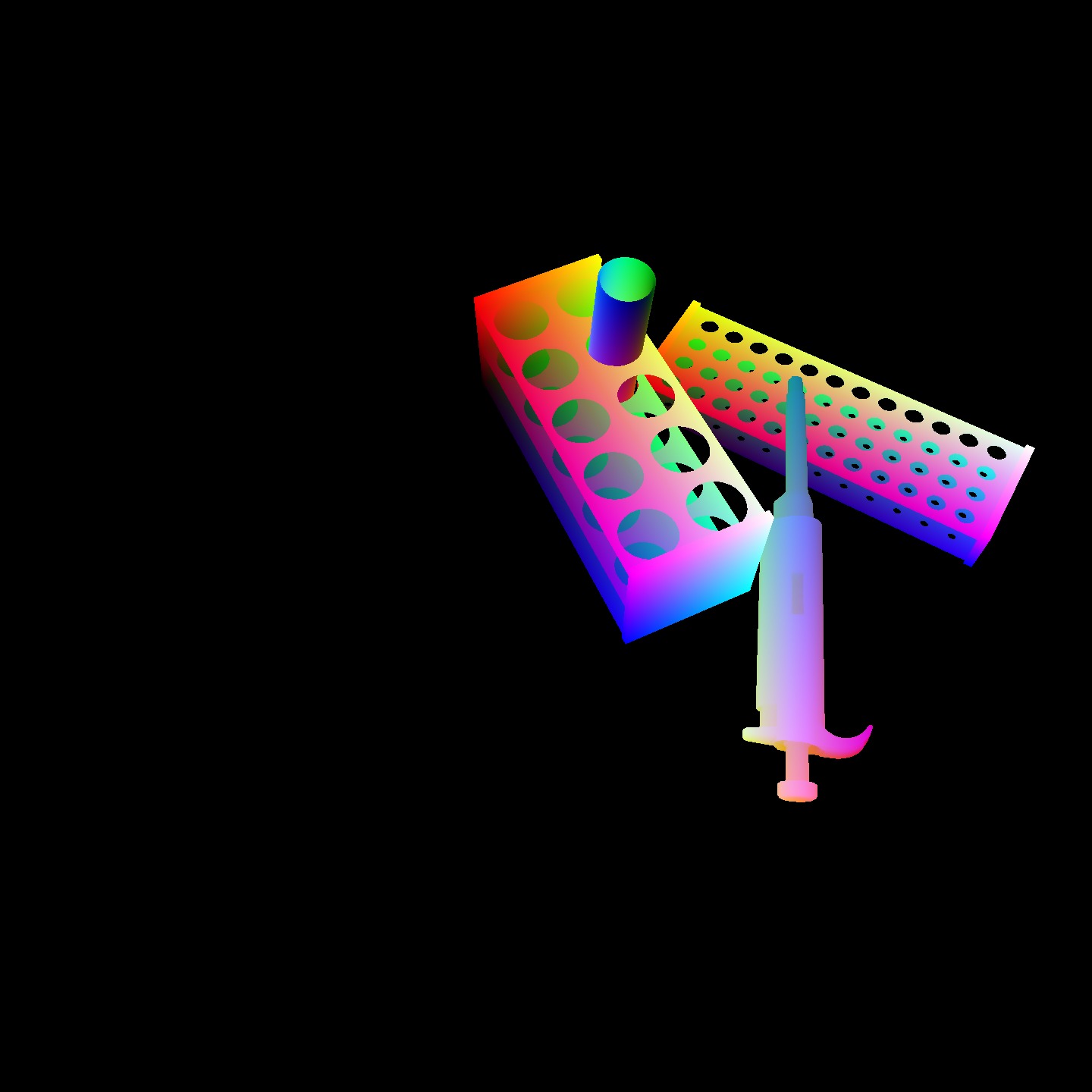}
}
\subfloat{
    \includegraphics[height=\sampleheight\linewidth, valign=t]{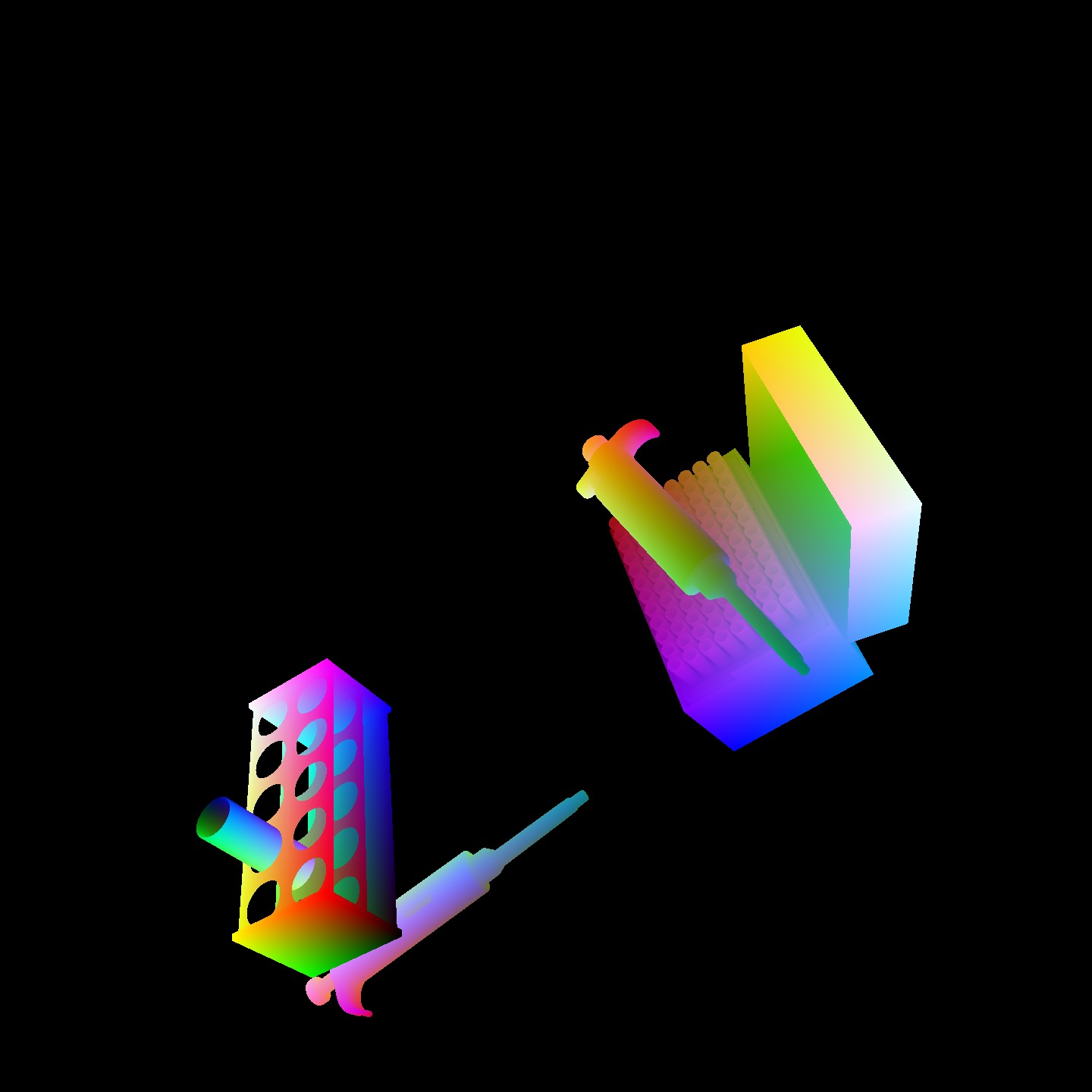}
}
\subfloat{
    \includegraphics[height=\sampleheight\linewidth, valign=t]{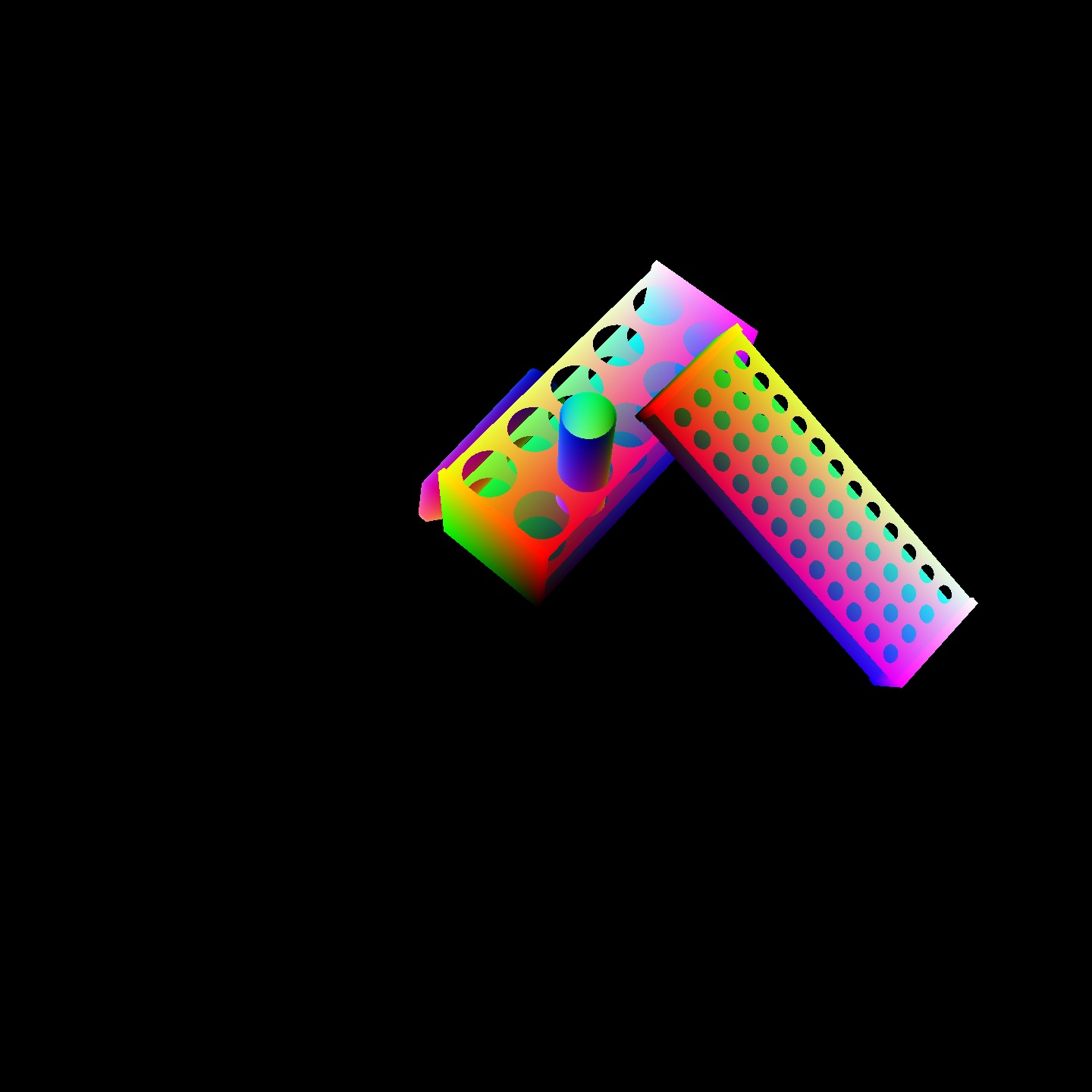}
}
\subfloat{
    \includegraphics[height=\sampleheight\linewidth, valign=t]{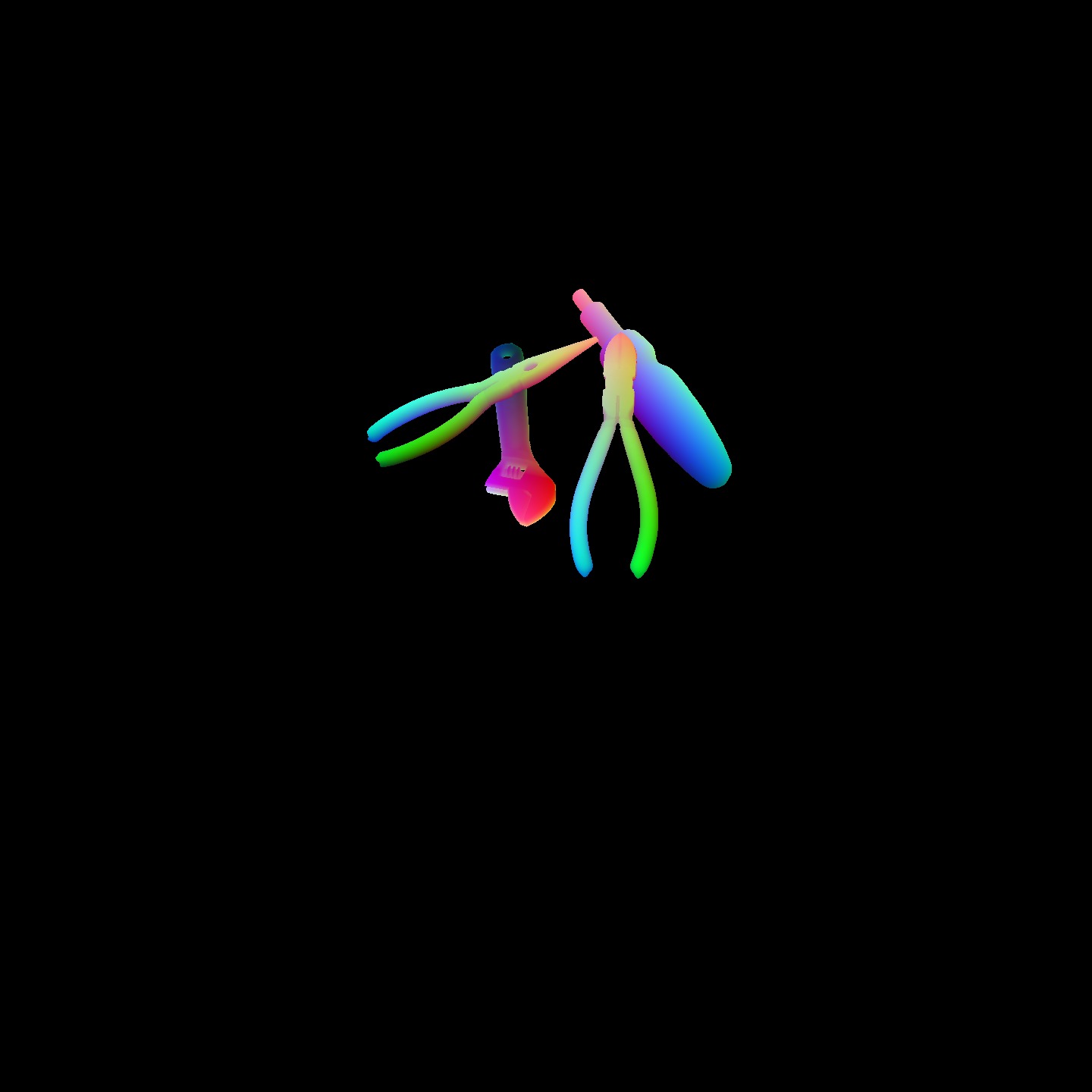}
}
\subfloat{
    \includegraphics[height=\sampleheight\linewidth, valign=t]{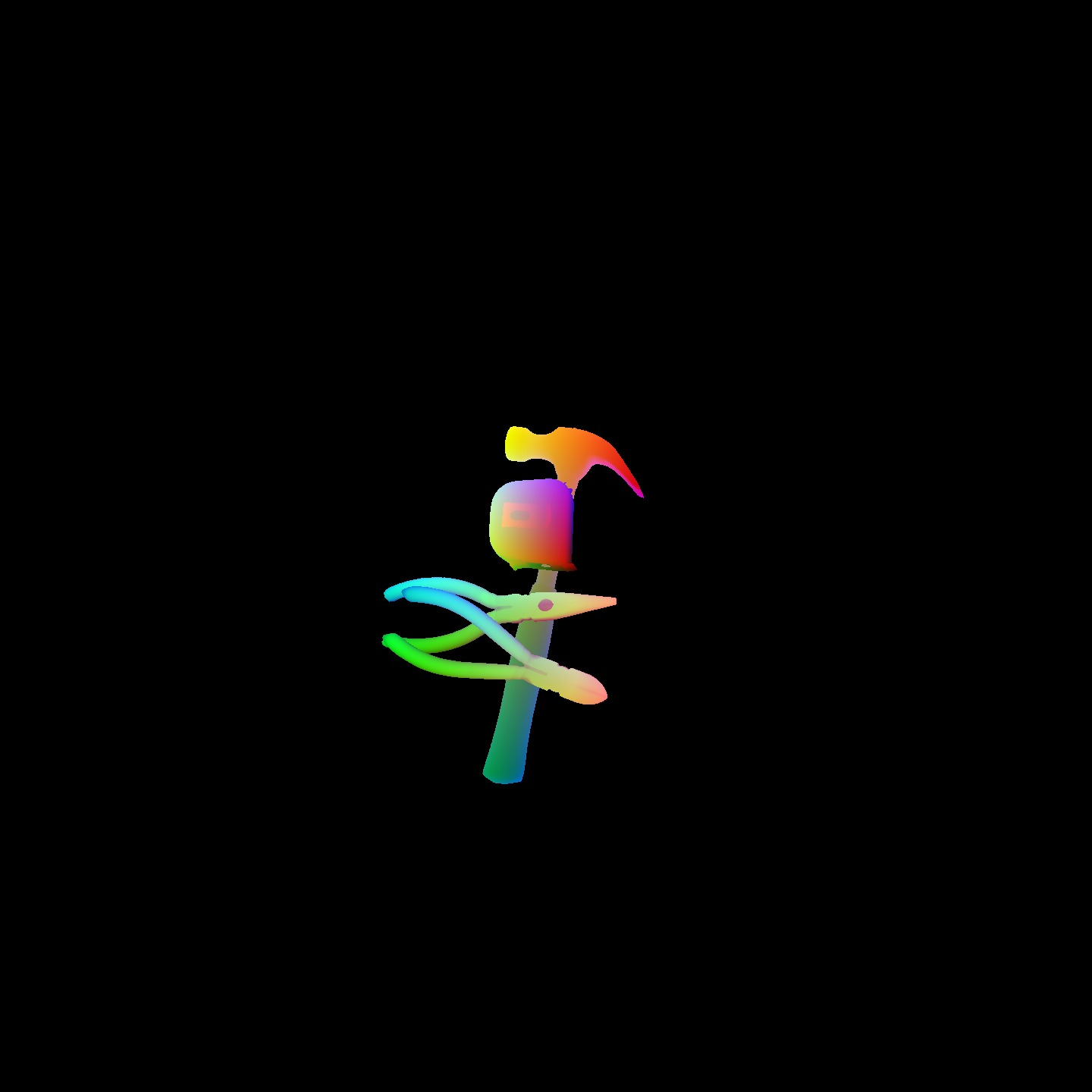}
}
\vspace{-1ex}
\caption{
\textbf{Visualization of data samples from StereOBJ-1M dataset. }
The first row is left stero images with semantic masks and bounding boxes superimposed.
In the second row, we use normalized coordinate map \cite{pix2pose,nocs} to illustrate the 6D poses of the corresponding objects, where the coordinates of the object surface points are normalized to $[0,1]^3$ and converted to RGB values in $[0,255]^3$ at projected pixels.
}
\vspace{-1ex}
\label{fig:stereobj:samples}
\end{figure*}

Viewpoint coverage of each object is illustrated in Figure \ref{fig:viewpoint:coverage}.
For objects such as microplate and sterile tip racks, there is only one possible side when putting on a desktop, so at most 50\% of viewpoint coverage.
Annotations in our dataset are 6D poses for every object in the scene, from which object instance segmentation masks, 2D and 3D bounding boxes, and normalized coordinate maps \cite{nocs} can be inferred.
We visualize some data samples from our dataset in Figure \ref{fig:stereobj:samples}.
As illustrated in the figure, the annotations of our dataset have high quality.

\subsection{Benchmark and Evaluation}
\label{sec:benchmark:evaluation}

\textbf{Train/Validation/Test Split.}
The image sequences are divided into train/validation and test sets such that scenes presented in the training set are held out in the validation and test set.
The test set contains 32 image sequences that are selected to cover most environments and ensure every object is tested in at least 4,000 images across at least 3 different scenes.
In the baseline experiments in Section \ref{sec:experiments}, we did not render additional synthetic data except basic geometric and photometric augmentation, because the capacity of StereOBJ-1M training set is sufficient to train large deep models.
However, users of the dataset can still opt to render additional data using the 3D mesh models we provide.

Among the objects, \texttt{centrifuge\_tube} is the only category with multiple instances recorded in a scene and is used in the multi-object pose detection task.
The rest of 17 objects are used in single-object pose estimation task which is the main focus of this paper.
Results for pose detection of \texttt{centrifuge\_tube} are provided in supplementary.

\textbf{Evaluation Metrics} 
We use the popular ADD \cite{linemod} and ADD-S \cite{posecnn} in our evaluation for 6D pose.
When computing ADD distance, we transform the model point set by the predicted and the ground truth poses respectively, and compute the mean 3D Euclidean distance between the two point sets.
Given an object with 3D model point set of $\mathcal{M} = \{\mathbf{x}_i \in \mathbb{R}^3\} $, the ADD distance is calculated as:
\begin{equation}\label{eq:add}
   \text{ADD} = \frac{1}{|\mathcal{M}|} \sum_{\mathbf{x}\in\mathcal{M}} || (\mathbf{R}\mathbf{x} + \mathbf{T}) - (\mathbf{R}^*\mathbf{x} + \mathbf{T}^*) ||_2 
\end{equation}
where $[\mathbf{R}^* | \mathbf{T}^*]$ and $[\mathbf{R} | \mathbf{T}]$ are the ground truth and estimated 6D poses.
For symmetric objects, ADD-S \cite{posecnn} is used instead.
When computing ADD-S distance, the 3D distances are calculated as the average of each point's closest distance to the other point set:
\begin{equation}\label{eq:add:s}
   \text{ADD-S} = \frac{1}{|\mathcal{M}|} \sum_{\mathbf{x}_1\in\mathcal{M}} \min_{\mathbf{x}_2\in\mathcal{M}} || (\mathbf{R}\mathbf{x}_1 + \mathbf{T}) - (\mathbf{R}^*\mathbf{x}_2 + \mathbf{T}^*) ||_2 
\end{equation}

\begin{table*}[ht]
\centering
\small
\centering
\begin{tabular}{l|cc|cc|cc}
\hline
input modality & \multicolumn{2}{c|}{ monocular RGB } & \multicolumn{4}{c}{ binocular stereo RGB } \\
\hline
pose optimization & \multicolumn{2}{c|}{ PnP \cite{pnp:ransac} } & \multicolumn{2}{c|}{ classic triangulation } & \multicolumn{2}{c}{ object triangulation } \\
\hline
\multirow{2}{*}{metrics} & \multirow{2}{*}{\begin{tabular}[c]{@{}c@{}} ADD(-S) \\ AUC-10cm \end{tabular}} &
\multirow{2}{*}{\begin{tabular}[c]{@{}c@{}} ADD(-S) \\ accuracy-0.1d \end{tabular}} & 
\multirow{2}{*}{\begin{tabular}[c]{@{}c@{}} ADD(-S) \\ AUC-10cm \end{tabular}} &
\multirow{2}{*}{\begin{tabular}[c]{@{}c@{}} ADD(-S) \\ accuracy-0.1d \end{tabular}} & 
\multirow{2}{*}{\begin{tabular}[c]{@{}c@{}} ADD(-S) \\ AUC-10cm \end{tabular}} &
\multirow{2}{*}{\begin{tabular}[c]{@{}c@{}} ADD(-S) \\ accuracy-0.1d \end{tabular}}
\\
& & & & & & \\
\hline
\texttt{blade\_razor} & 20.79 & 3.89 & 40.65 & 0.00 & \textbf{55.17} & \textbf{12.00} \\
\texttt{hammer} & 10.35 & 3.66 & 18.91 & 2.12 & \textbf{38.62} & \textbf{18.45} \\
\texttt{microplate} & 42.39 & 27.34 &  58.95 & 23.18 & \textbf{60.65} & \textbf{39.22} \\
\texttt{needle\_nose\_pliers} & 38.63 & 23.39 & 62.82 & 11.21 & \textbf{74.13} & \textbf{51.72} \\
\texttt{pipette\_0.5\_10} & 21.96 & 15.03 & 19.32 & 2.82 & \textbf{35.68} & \textbf{21.90} \\
\texttt{pipette\_100\_1000} & 9.51 & 0.49 & 14.53 & 0.38 & \textbf{21.53} & \textbf{1.77} \\
\texttt{pipette\_10\_100} & 25.21 & 14.05 & 25.70 & 1.43 & \textbf{47.65} & \textbf{25.38} \\
\texttt{screwdriver} & 32.73 & 23.71 & 65.21 & 21.68 & \textbf{72.57} & \textbf{50.56} \\
\texttt{side\_cutters} & 18.00 & 6.52 & 60.89 & 9.85 & \textbf{69.36} & \textbf{39.12} \\
\texttt{sterile\_tip\_rack\_10} & 71.85 & 64.28 & 38.73 & 1.81 & \textbf{73.88} & \textbf{64.06} \\
\texttt{sterile\_tip\_rack\_1000} & \textbf{76.10} & \textbf{69.12} & 38.71 & 11.02 & 75.84 & 68.53 \\
\texttt{sterile\_tip\_rack\_200} & 75.74 & 66.99 & 41.00 & 1.91 & \textbf{78.44} & \textbf{67.04} \\
\texttt{tape\_measure} & 18.55 & 1.47 & 57.43 & 0.00 & \textbf{68.86} & \textbf{15.07} \\
\texttt{tube\_rack\_1.5\_2\_ml} & 44.67 & 33.34 & 58.27 & 35.74 & \textbf{60.10} & \textbf{44.34} \\
\texttt{tube\_rack\_50\_ml} & 66.91 & 63.92 & 58.93 & 33.54 & \textbf{75.99} & \textbf{72.05} \\
\texttt{wire\_stripper} & 31.12 & 22.11 & 65.15 & 26.59 & \textbf{81.93} & \textbf{71.78} \\
\texttt{wrench} & 8.03 & 0.66 & 33.84 & 0.03 & \textbf{42.54} & \textbf{7.10} \\
\hline
average & 36.03 & 25.88 & 44.65 & 10.78 & \textbf{60.76} & \textbf{39.42} \\
\hline
\end{tabular}
\vspace{-1ex}
\caption{ The results of \textbf{KeyPose \cite{keypose}} on single-object pose estimation in terms of \textbf{ADD(-S) AUC} and \textbf{ADD(-S) accuracy} on StereOBJ-1M dataset. 
The input modality include monocular and binocular stereo RGB images.
}
\label{tab:keypose:results}
\end{table*}

\begin{table*}[ht]
\centering
\small
\centering
\begin{tabular}{l|cc|cc|cc}
\hline
input modality & \multicolumn{2}{c|}{ monocular RGB } & \multicolumn{4}{c}{ binocular stereo RGB } \\
\hline
pose optimization & \multicolumn{2}{c|}{ PnP \cite{pnp:ransac} } & \multicolumn{2}{c|}{ classic triangulation } & \multicolumn{2}{c}{ object triangulation } \\
\hline
\multirow{2}{*}{metrics} & \multirow{2}{*}{\begin{tabular}[c]{@{}c@{}} ADD(-S) \\ AUC-10cm \end{tabular}} &
\multirow{2}{*}{\begin{tabular}[c]{@{}c@{}} ADD(-S) \\ accuracy-0.1d \end{tabular}} & 
\multirow{2}{*}{\begin{tabular}[c]{@{}c@{}} ADD(-S) \\ AUC-10cm \end{tabular}} &
\multirow{2}{*}{\begin{tabular}[c]{@{}c@{}} ADD(-S) \\ accuracy-0.1d \end{tabular}} & 
\multirow{2}{*}{\begin{tabular}[c]{@{}c@{}} ADD(-S) \\ AUC-10cm \end{tabular}} &
\multirow{2}{*}{\begin{tabular}[c]{@{}c@{}} ADD(-S) \\ accuracy-0.1d \end{tabular}}
\\
& & & & & & \\
\hline
\texttt{blade\_razor} & 24.63 & 11.45 & 41.78 & 0.02 & \textbf{75.81} & \textbf{46.97} \\
\texttt{hammer} & 12.52 & 3.59 & 17.61 & 2.11 & \textbf{38.23} & \textbf{19.73} \\
\texttt{microplate} & 15.30 & 6.08 & \textbf{44.15} & 8.71 & 43.32 & \textbf{18.72} \\
\texttt{needle\_nose\_pliers} & 8.87 & 4.14 & 59.20 & 8.80 & \textbf{74.40} & \textbf{51.78} \\
\texttt{pipette\_0.5\_10} & 7.12 & 2.57 & 20.68 & 2.44 & \textbf{40.07} & \textbf{19.30} \\
\texttt{pipette\_100\_1000} & 2.55 & 0.02 & 12.37 & 0.19 & \textbf{24.83} & \textbf{0.65} \\
\texttt{pipette\_10\_100} & 17.72 & 4.83 & 20.73 & 0.57 & \textbf{50.51} & \textbf{25.73} \\
\texttt{screwdriver} & 43.50 & 29.71 & 57.50 & 16.00 & \textbf{77.01} & \textbf{55.94} \\
\texttt{side\_cutters} & 52.52 & 31.27 & 62.61 & 13.14 & \textbf{84.41} & \textbf{69.11} \\
\texttt{sterile\_tip\_rack\_10} & 63.83 & 51.93 & 19.68 & 0.84 & \textbf{66.28} & \textbf{50.17} \\
\texttt{sterile\_tip\_rack\_1000} & 66.62 & 56.20 & 20.67 & 2.90 & \textbf{72.61} & \textbf{63.44} \\
\texttt{sterile\_tip\_rack\_200} & 65.54 & 49.73 & 24.31 & 0.57 & \textbf{76.46} & \textbf{62.52} \\
\texttt{tape\_measure} & 51.30 & 7.23 & 56.27 & 0.00 & \textbf{79.73} & \textbf{29.20} \\
\texttt{tube\_rack\_1.5\_2\_ml} & 32.75 & 22.32 & 48.51 & 18.77 & \textbf{52.05} & \textbf{35.11} \\
\texttt{tube\_rack\_50\_ml} & 71.67 & 68.37 & 52.13 & 22.20 & \textbf{76.91} & \textbf{75.04} \\
\texttt{wire\_stripper} & 73.02 & 56.73 & 55.81 & 16.11 & \textbf{82.74} & \textbf{72.63} \\
\texttt{wrench} & 18.13 & 4.60 & 33.62 & 0.02 & \textbf{59.71} & \textbf{22.72} \\
\hline
average & 36.92 & 24.16 & 38.10 & 6.67 & \textbf{63.24}  & \textbf{42.28}  \\
\hline
\end{tabular}
\vspace{-1ex}
\caption{ The results of \textbf{PVNet \cite{pvnet}} on single-object pose estimation in terms of \textbf{ADD(-S) AUC} and \textbf{ADD(-S) accuracy} on StereOBJ-1M dataset. 
The input modality include monocular and binocular stereo RGB images.
}
\label{tab:pvnet:results}
\end{table*} 

We use the following two evaluation metrics.
(1) \textbf{ADD(-S) accuracy}: ADD(-S) accuracy measures the proportion of correct pose predictions. A pose prediction is considered correct if the ADD(-S) distance is less than the threshold of 10\% of the model's diameter.
(2) \textbf{ADD(-S) AUC}: the area under ADD(-S) accuracy-threshold curve where the maximum threshold is set to 10cm.

\section{Experiments}
\label{sec:experiments}

\subsection{Baseline Methods}

We implement and evaluate two methods on StereOBJ-1M dataset as baselines for future experiments.
Specifically, we implement PVNet \cite{pvnet} and KeyPose \cite{keypose}, two classic keypoint-based 6D pose estimation frameworks that have achieved state-of-the-art performance on various datasets.
We train the two models on the merged train and validation sets and report their performance on the test set.

\textbf{PVNet \cite{pvnet}} is a single-RGB keypoint-based method.
It represents keypoints using a 2D direction field and estimates the 2D locations of the keypoints by RANSAC-based voting scheme \cite{pnp:ransac}.
The 6D poses are determined by solving a Perspective-n-Point (PnP) problem \cite{pnp:ransac}.

\textbf{KeyPose \cite{keypose}} is a stereo-RGB keypoint-based method.
Different from PVNet, it localizes object keypoint by predicting heatmaps in both stereo images. 
The 6D object poses are calculated by keypoint triangulation from two-view stereo and Orthogonal Procrustes pose fitting.

\subsection{Monocular Image Experiments}

We conduct monocular image experiments where only the left images are used as input to predict the 6D pose.
The stereo method KeyPose \cite{keypose} is adapted to its monocular variant where only keypoints in the left stereo images are predicted using heatmaps, and the 6D poses are calculated by solving PnP problem \cite{pnp:ransac}.
The results are illustrated in columns 1-2 in Tables \ref{tab:keypose:results} and \ref{tab:pvnet:results}.
KeyPose and PVNet respectively achieve 36.03\% and 36.92\%  in average ADD(-S) AUC, and 25.88\% and 24.16\% in average ADD(-S) accuracy.
Among the objects, the performance of both baseline methods suffers especially on pipette categories, which highlights the challenge of pose estimation of visually similar but different object instances.

\subsection{Stereo Image Experiments}

We conduct stereo experiments where both stereo images are used as input to predict 6D pose.
The monocular method PVNet \cite{pvnet} is adapted to its stereo variant where keypoints in both stereo images are predicted individually.
For both baseline methods, suppose $[u_k^L, v_k^L]$ and $[u_k^R, v_k^R]$ are the predicted 2D locations of $k$-th keypoint in left and right cameras respectively, $\Pi_L$ and $\Pi_R$ are the camera projection of left and right cameras respectively, and $\mathbf{x}_k^* \in \mathbb{R}^3$ is the $k$-th keypoint in the canonical object pose. 
We investigate the following two methods for calculating 6D pose from keypoints predictions in both stereo images.

\textbf{Classic Triangulation.}
Given $[u_k^L, v_k^L]$ and $[u_k^R, v_k^R]$ for all $k$, a na\"ive method to compute object 6D pose is to follow classic point-level triangulation used in KeyPose \cite{keypose}, i.e.  triangulate 3D keypoints from stereo and fit them to canonical object 3D keypoints by solving an Orthogonal Procrustes problem, to obtain the estimated pose $[\mathbf{R}_c | \mathbf{T}_c]$:
\begin{multline}\label{eq:classic:triangulation}
\hspace*{-2ex}
\mathbf{x}_k = \underset{\mathbf{x} \in \mathbb{R}^3}{\arg\min}  ||\Pi_L(\mathbf{x}) - [u_k^L, v_k^L]||_2^2 + ||\Pi_R(\mathbf{x}) - [u_k^R, v_k^R]||_2^2 \\
[\mathbf{R}_c|\mathbf{T}_c] = \underset{\mathbf{R},\mathbf{T}}{\arg\min} \sum_{k=1}^K ||(\mathbf{R}\mathbf{x}_k^*+\mathbf{T}) - \mathbf{x}_k||_2^2 
\end{multline}
where $\mathbf{x}_k \in \mathbb{R}^3$ is the triangulated 3D location of the $k$-th keypoint.
The second step in \eqref{eq:classic:triangulation} can use RANSAC \cite{pnp:ransac}.

\textbf{Object Triangulation. }
We propose a novel object-level triangulation approach as a stronger baseline.
Compared to classic triangulation which optimizes the 3D location of a \emph{point}, we directly optimize the 6D pose of an object from 2D keypoint predictions in both images.
Mathematically, Object Triangulation combines the two steps in Equation \eqref{eq:classic:triangulation} into one unified optimization of the 6D pose $[\mathbf{R}_o | \mathbf{T}_o]$:
\begin{multline}\label{eq:object:triangulation}
\hspace*{-2ex}
[\mathbf{R}_o|\mathbf{T}_o] = \underset{\mathbf{R},\mathbf{T}}{\arg\min} \sum_{k=1}^K ||\Pi_L(\mathbf{R}\mathbf{x}_k^*+\mathbf{T}) - [u_k^L, v_k^L]||_2^2 + \\
||\Pi_R(\mathbf{R}\mathbf{x}_k^*+\mathbf{T}) - [u_k^R, v_k^R]||_2^2 
\end{multline}

We use the Levenberg-Marquardt algorithm \cite{lm:optimization} as the non-linear optimization method together with RANSAC \cite{pnp:ransac}.
The results of the two baseline architectures with two pose optimization methods are illustrated in columns 3-6 in Tables \ref{tab:keypose:results} and \ref{tab:pvnet:results}.
Baseline methods with Object Triangulation consistently improve over monocular variants on all object categories significantly while classic triangulation can yield worse results.
With Object Triangulation, the stereo variants of both baseline methods significantly outperform their monocular counterparts on StereOBJ-1M, by at least 24\% in ADD(-S) AUC and 13\% in ADD(-S) accuracy.

\section{Conclusions}

In this work, we propose a novel object pose data capturing and annotation pipeline and present a large-scale object pose dataset with stereo RGB as input.
We benchmark two state-of-the-art algorithms for 6D object pose estimation and propose a novel method for stereo object pose optimization that outperforms classic triangulation method.
In addition to pose estimation, our dataset enables future research directions such as object reconstruction and scene flow estimation \cite{flownet3d,meteornet} from stereo RGB.

\textbf{Acknowledgement. }
This work is funded in part by JST AIP Acceleration, Grant Number JPMJCR20U1, Japan.

{\small
\bibliographystyle{ieee_fullname}
\bibliography{bib}
}

\appendix

\section{Overview}

In this document, we provide additional details on StereOBJ-1M dataset as presented in the main paper. 
We present additional baseline results on instance-level pose detection for \texttt{centrifuge\_tube} class in Section \ref{sec:pose:det}. 
In Section \ref{sec:data:hardware}, we provide details on the hardware of data capturing. 
In Section \ref{sec:viewpoint:coverage}, we provide more details on viewpoint distribution of each object class.
Lastly, in Section \ref{sec:more:viz}, we visualize more data samples from our dataset.

\section{Multi-instance Pose Detection Results}
\label{sec:pose:det}

In the main paper, we report the results of two baselines on \textbf{single-object pose estimation} of 17 out of 18 objects on the test set where there is at most one object instance from a category in a scene.
However, for \texttt{centrifuge\_tube}, there are usually multiple instances recorded in a scene.
Therefore, \texttt{centrifuge\_tube} is used in \textbf{multi-object pose detection} task.
In this task, the framework is supposed to perform instance-level detection and pose estimation simultaneously.

To adapt to instance-level pose detection, we modify the baseline formulation by introducing additional 2D object detection before pose estimation.
Given a detected rough 2D bounding box of an object instance, we crop the image patch and send it to pose estimation baselines, i.e. PVNet \cite{pvnet} and KeyPose \cite{keypose}, to estimate the 2D keypoint locations and therefore 6D pose of that object instance.
The 2D object detector we used is Faster-RCNN \cite{faster:rcnn}.

We use Average Precision (AP) as the evaluation metrics of multi-instance pose detection.
When calculating AP in 2D object detection, a detection result is considered correct if the IoU between the detected bounding box and a ground truth bounding box is larger than a threshold.
Different from 2D object detection, we consider a pose detection result to be correct if the ADD(-S) distance between the detected 6D pose and a ground truth pose is smaller than a threshold.
We use 10\% of the object diameter as the threshold of ADD(-S).
We report the pose detection results with single-RGB image as input in Table \ref{tab:pose:det:results}.
We notice that the above baseline suffers when two or multiple instances object overlap in the image and are included in the same image patch.
In this case, the pose estimation framework cannot distinguish different instances and fails in keypoint prediction.

\section{Data Capturing Hardware}
\label{sec:data:hardware}

\begin{figure}[t] 
\newcommand\hardwareheight{0.107}
\centering
\small
\subfloat[]{
    \includegraphics[width=0.48\linewidth, valign=t]{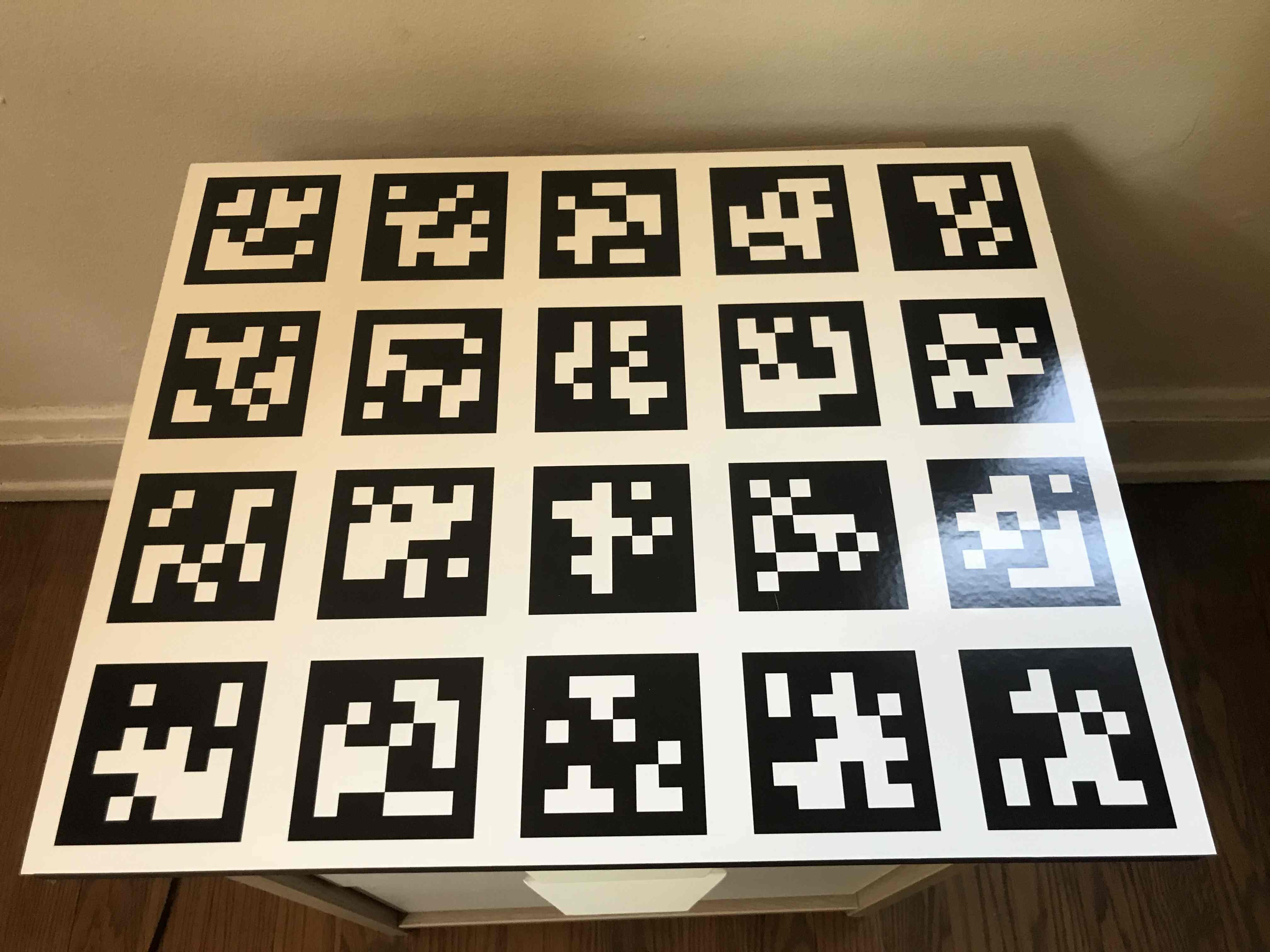}
}
\subfloat[]{
    \includegraphics[width=0.48\linewidth, valign=t]{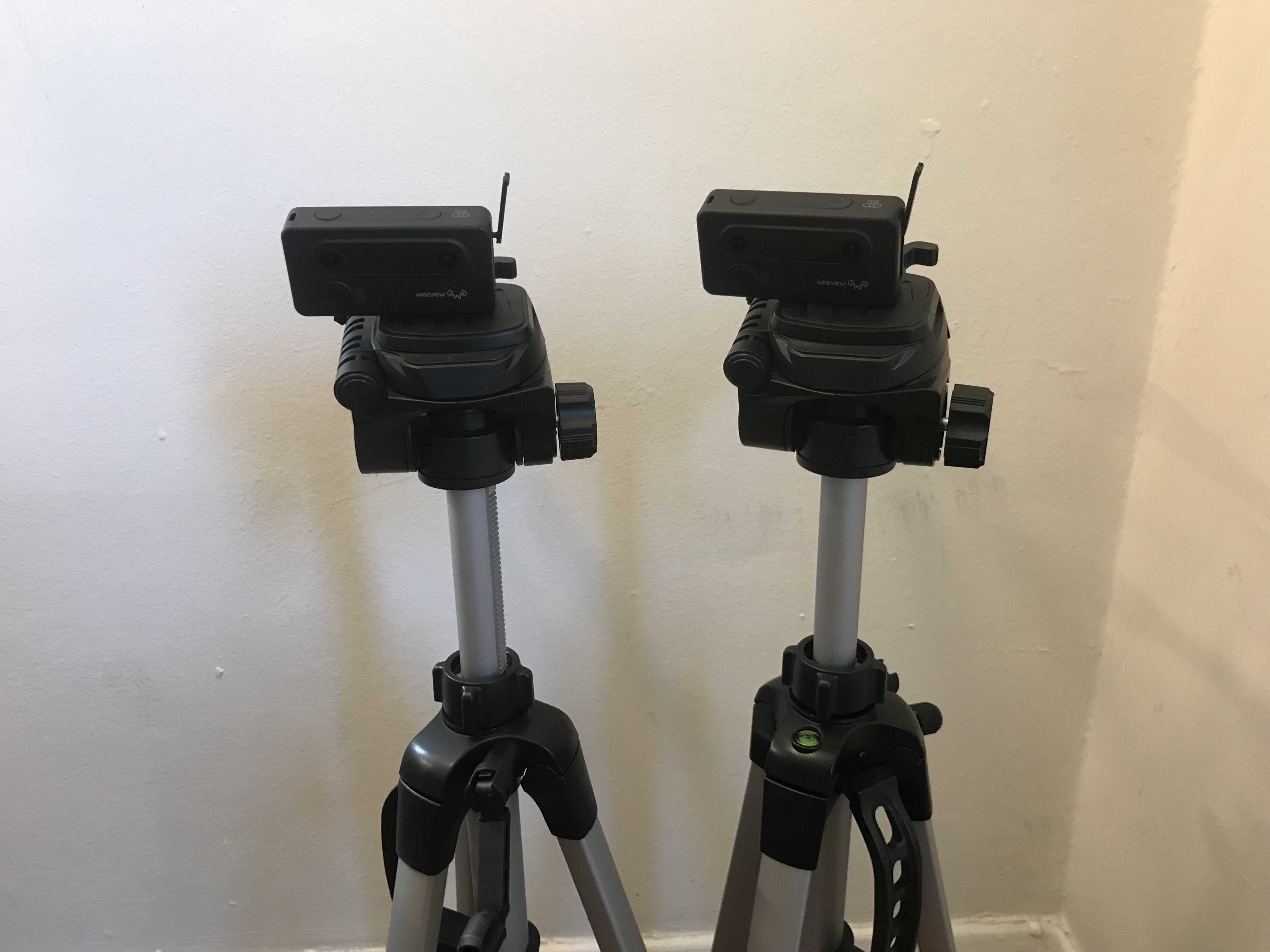}
}
\\
\vspace{-2ex}
\subfloat[]{
    \includegraphics[width=0.48\linewidth, valign=t]{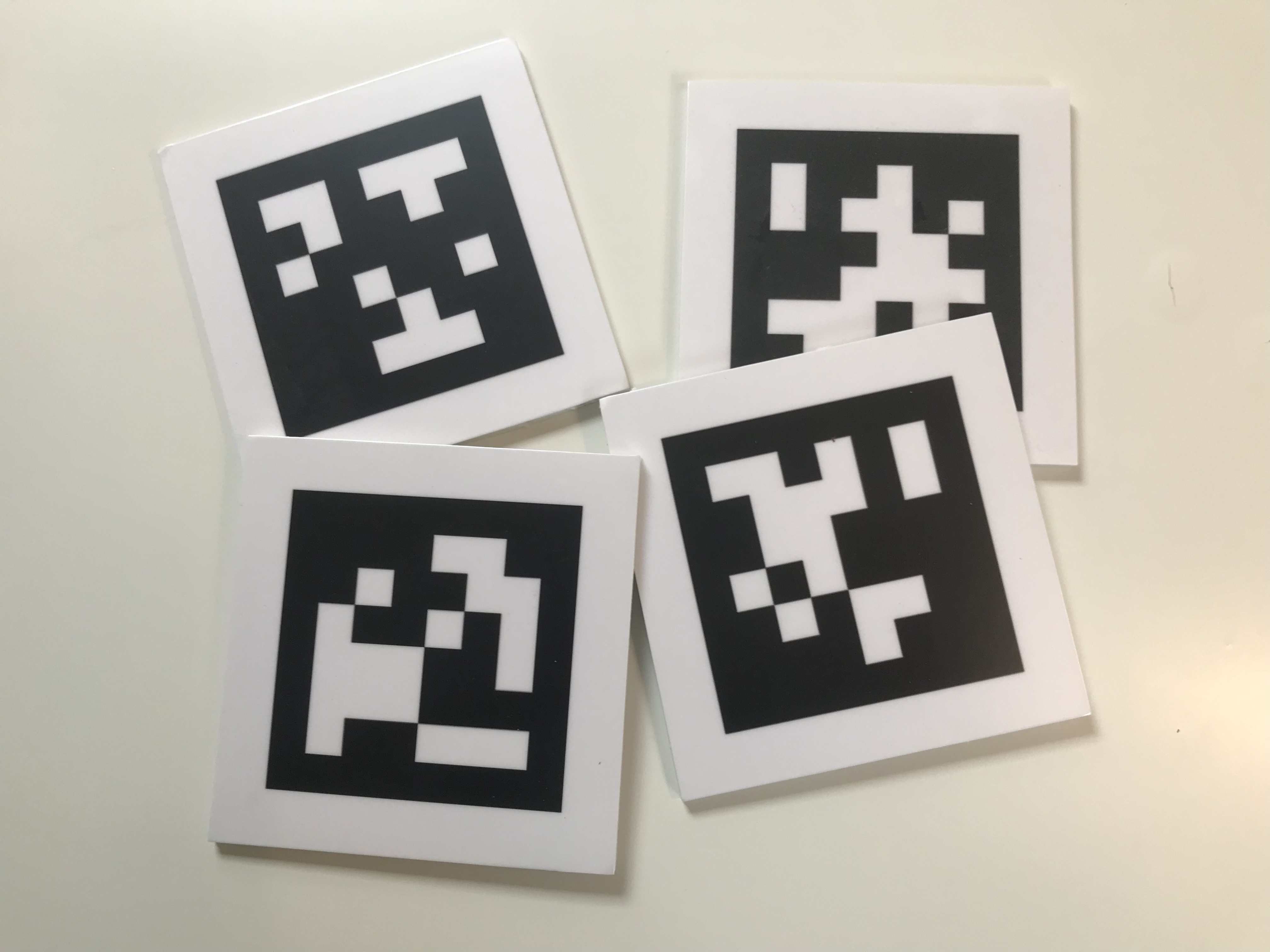}
}
\subfloat[]{
    \includegraphics[width=0.48\linewidth, valign=t]{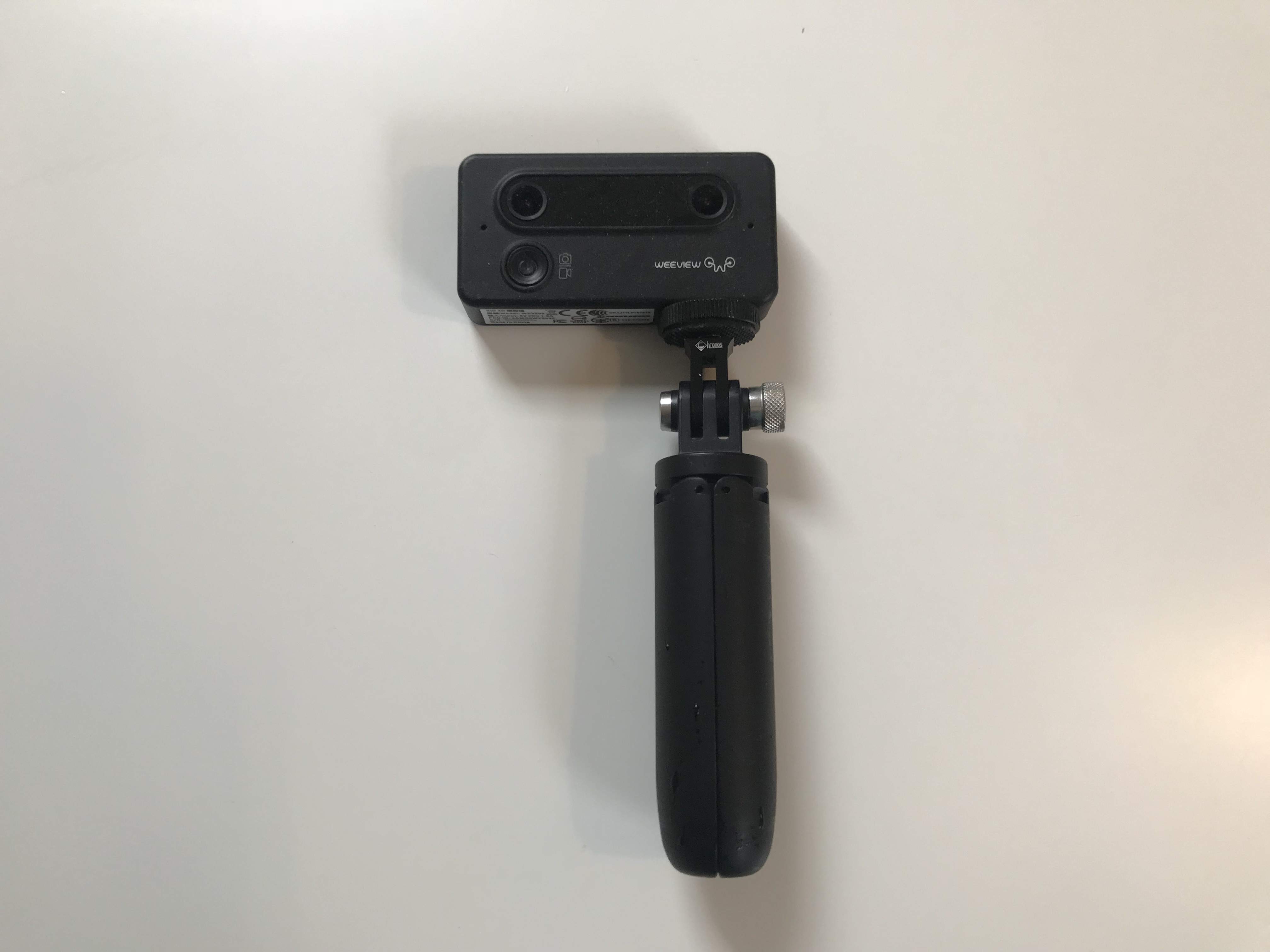}
}
\vspace{-1ex}
\caption{\textbf{Hardware for data collection.} (a) large fiducial marker board; (b) static cameras with tripods; (c) small fiducial markers; (d) moving stereo camera.
}
\label{fig:data:hardware}
\end{figure}

We present the hardware used for capturing the data in Figure \ref{fig:data:hardware}, including a large fiducial marker board, several small fiducial markers, two static cameras with two tripods, and one moving stereo camera.
We used the same Weewiew stereo camera \cite{weeview} for all three cameras, though the two stereo cameras can be monocular.
Weewiew stereo camera has a stereo baseline of approximately 4.5cm which is close to the distance between the two human eyes.
All three cameras are calibrated.

The fiducial markers are the first 20 AprilTags \cite{apriltag2}.
The large fiducial marker board is printed on a 20in $\times$ 16in plastic picture frame.
Though the large fiducial marker board needs to be accurately measured by its physical dimensions with a vernier caliper, the small fiducial markers do not.

\begin{table}[t]
\centering
\small
\caption{ AP results of \textbf{object pose detection} with single RGB image as input.}
\label{tab:pose:det:results}
\centering
\begin{tabular}{l|c|c}
\hline
method & PVNet \cite{pvnet} & KeyPose \cite{keypose} \\
\hline
\texttt{centrifuge\_tube} & 15.19 & 17.64 \\
\hline
\end{tabular}
\end{table}

\begin{figure*}[h]
\newcommand\objheight{0.169}
\centering
\small
\subfloat[]{
    \includegraphics[height=\objheight\linewidth, valign=t]{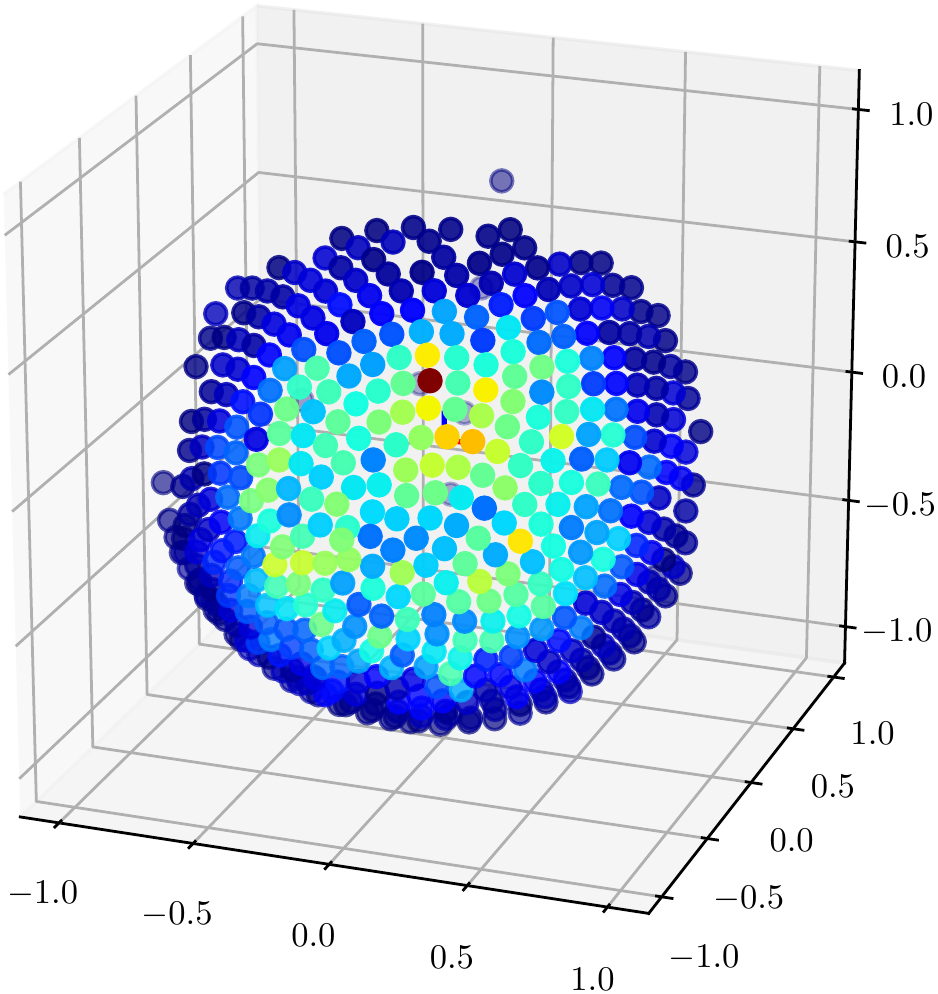}
}
\hspace{-0.01\linewidth}
\subfloat[]{
    \includegraphics[height=\objheight\linewidth, valign=t]{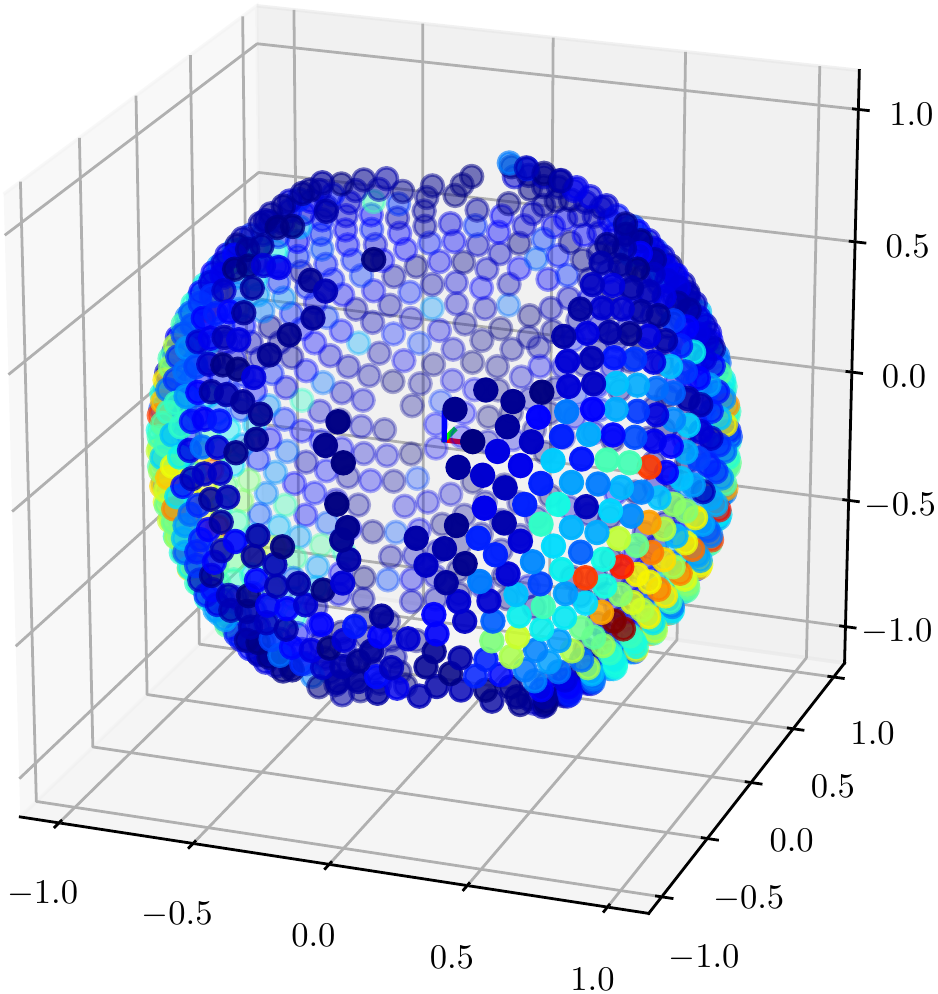}
}
\hspace{-0.01\linewidth}
\subfloat[]{
    \includegraphics[height=\objheight\linewidth, valign=t]{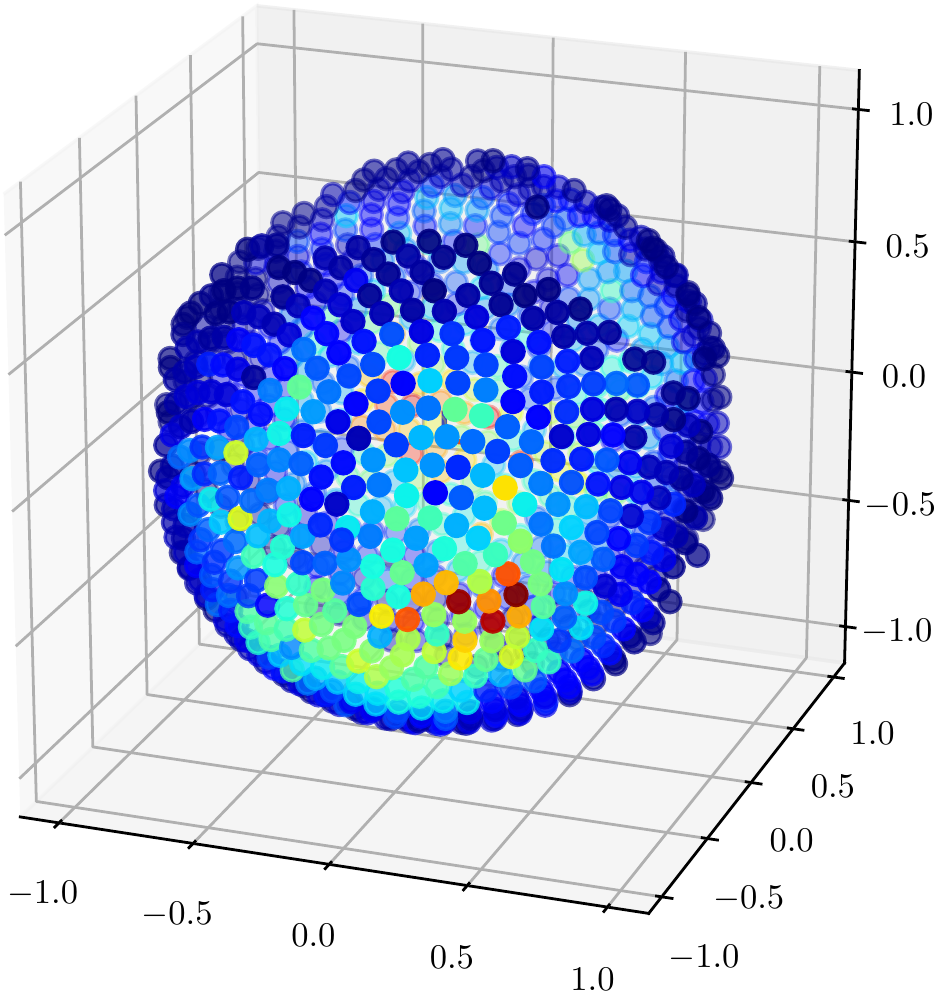}
}
\hspace{-0.01\linewidth}
\subfloat[]{
    \includegraphics[height=\objheight\linewidth, valign=t]{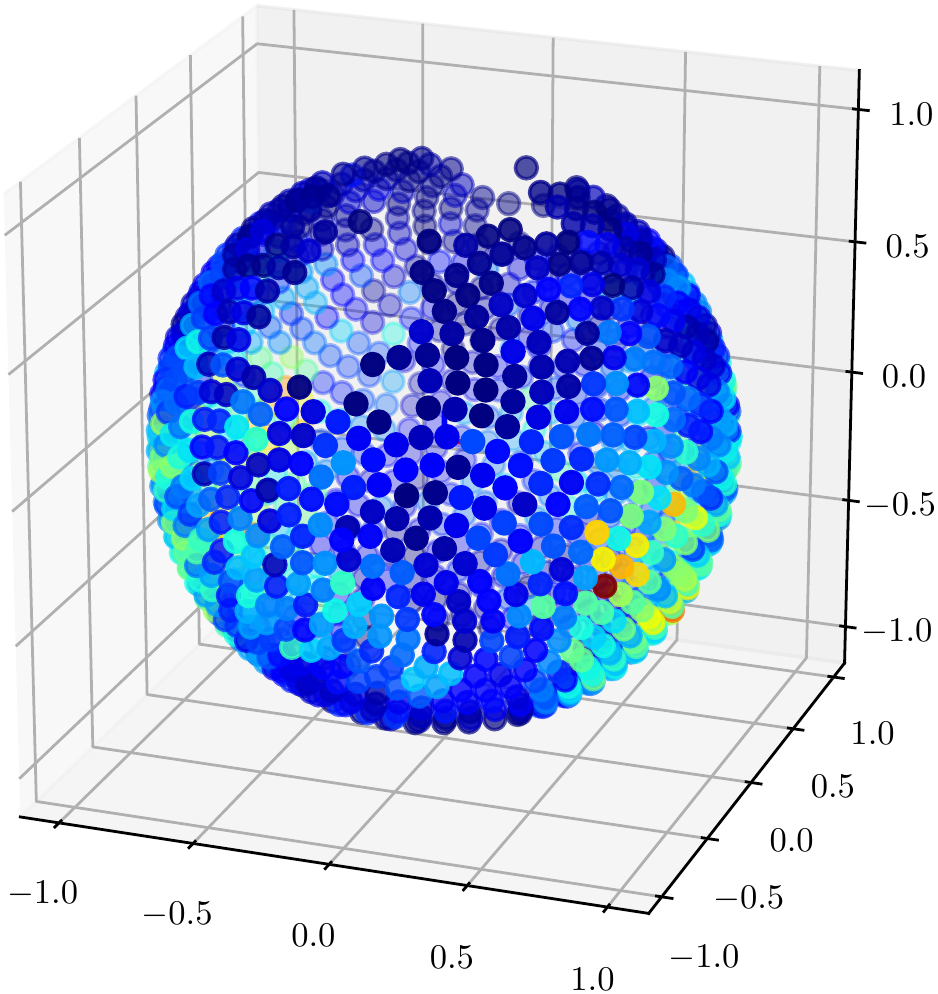}
}
\hspace{-0.01\linewidth}
\subfloat[]{
    \includegraphics[height=\objheight\linewidth, valign=t]{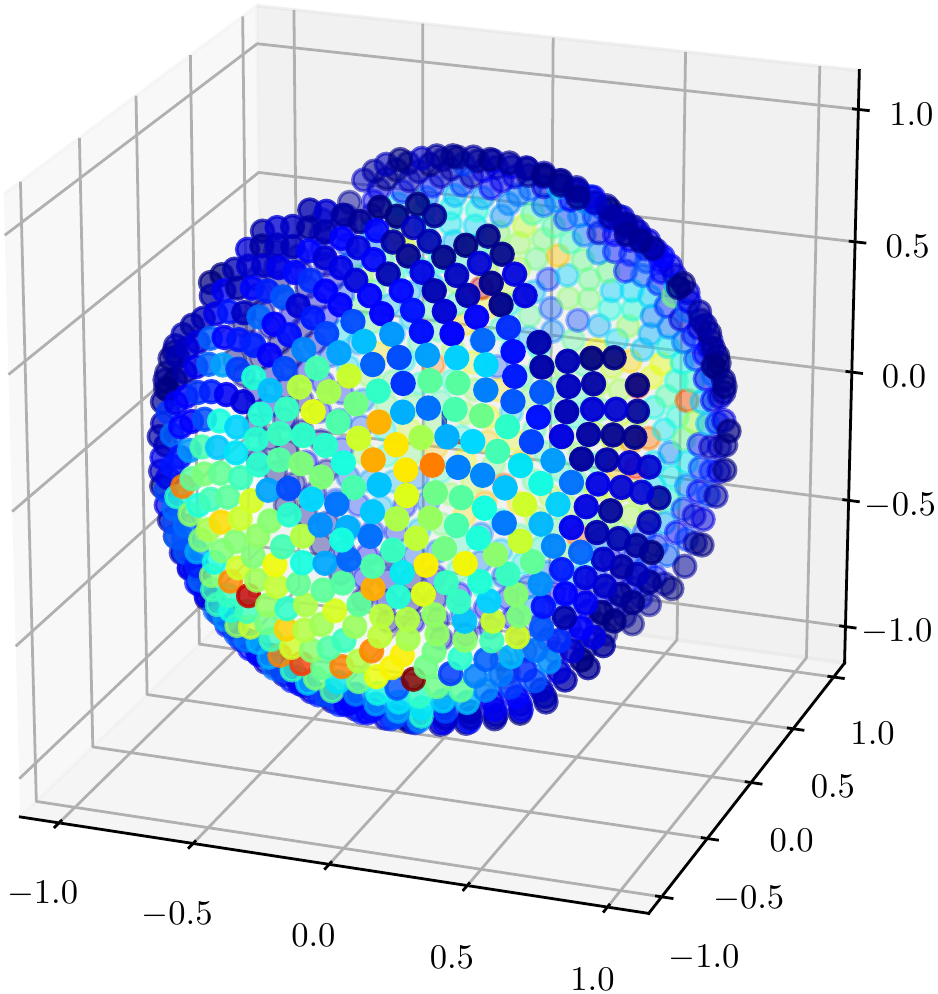}
}
\hspace{-0.01\linewidth}
\subfloat[]{
    \includegraphics[height=\objheight\linewidth, valign=t]{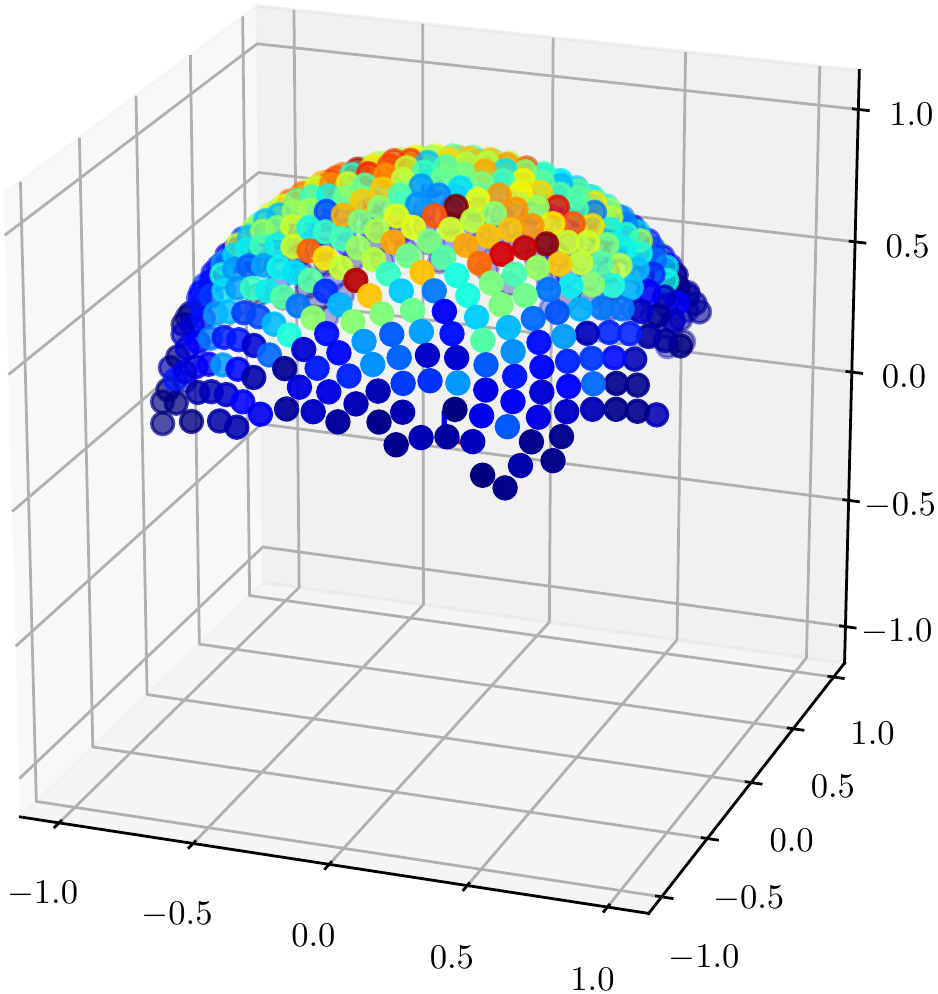}
}
\\
\subfloat[]{
    \includegraphics[height=\objheight\linewidth, valign=t]{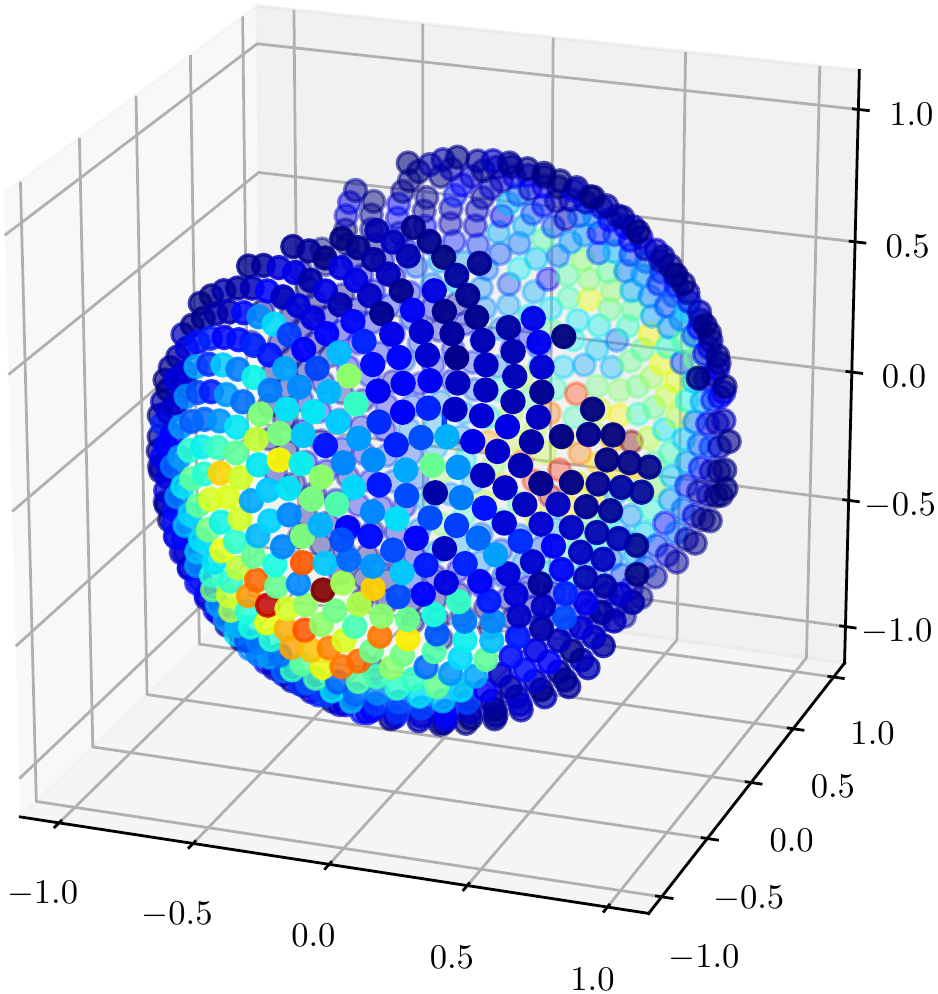}
}
\hspace{-0.01\linewidth}
\subfloat[]{
    \includegraphics[height=\objheight\linewidth, valign=t]{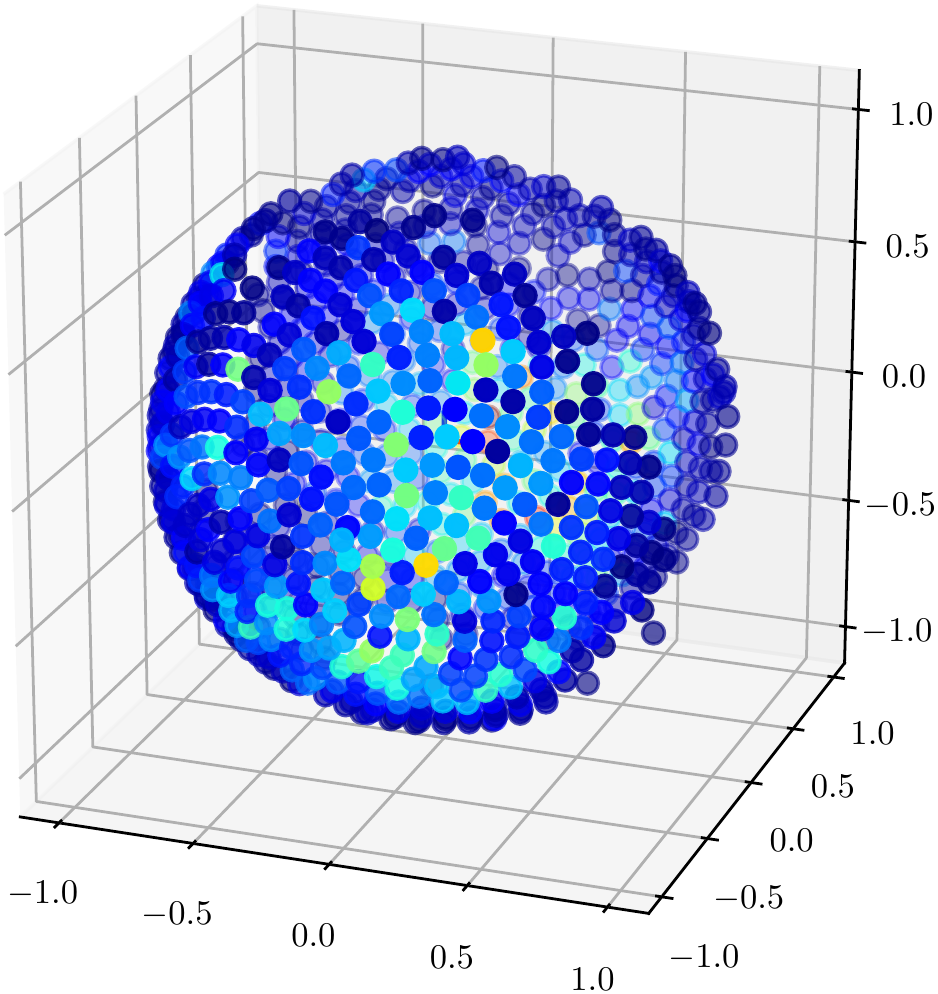}
}
\hspace{-0.01\linewidth}
\subfloat[]{
    \includegraphics[height=\objheight\linewidth, valign=t]{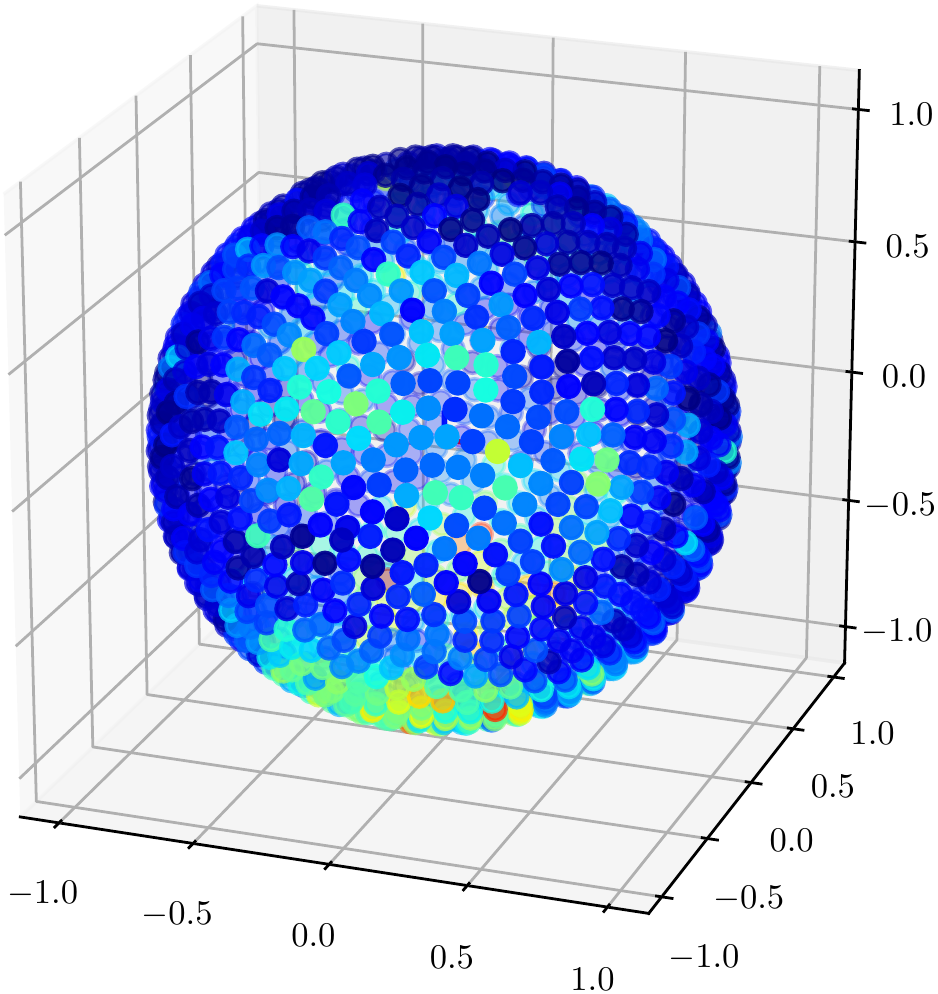}
}
\hspace{-0.01\linewidth}
\subfloat[]{
    \includegraphics[height=\objheight\linewidth, valign=t]{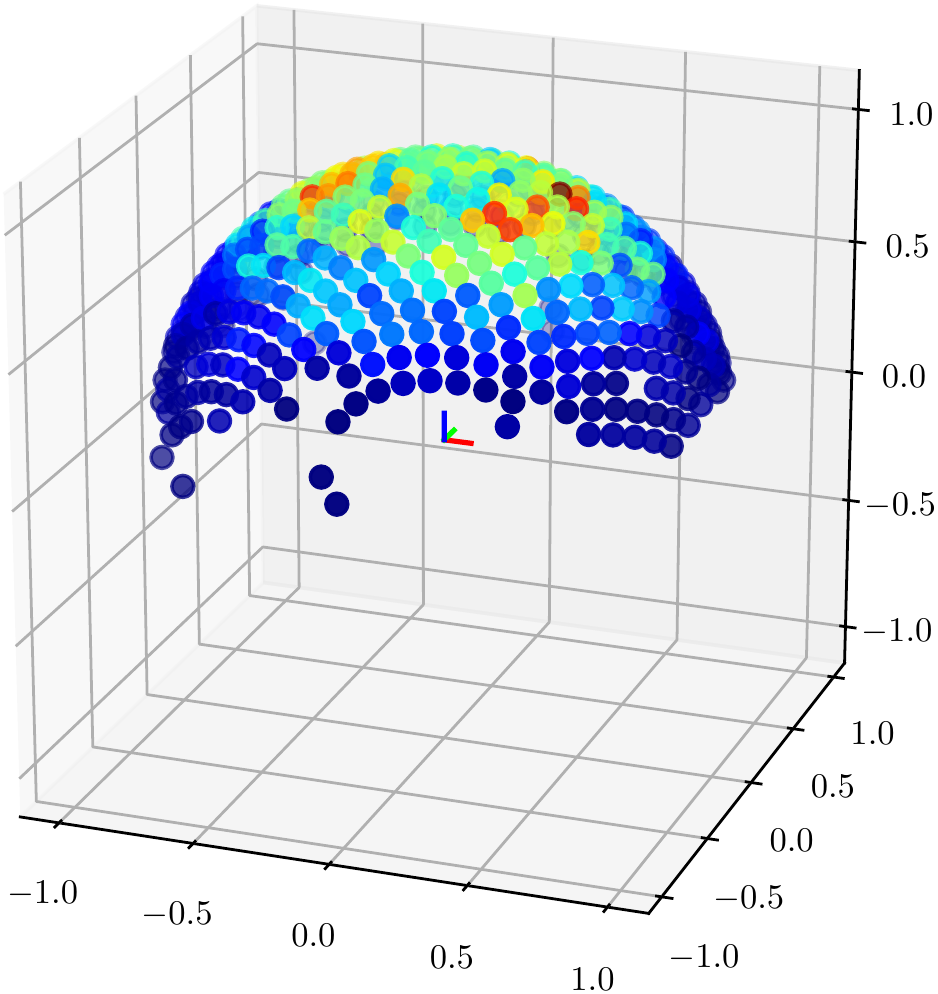}
}
\hspace{-0.01\linewidth}
\subfloat[]{
    \includegraphics[height=\objheight\linewidth, valign=t]{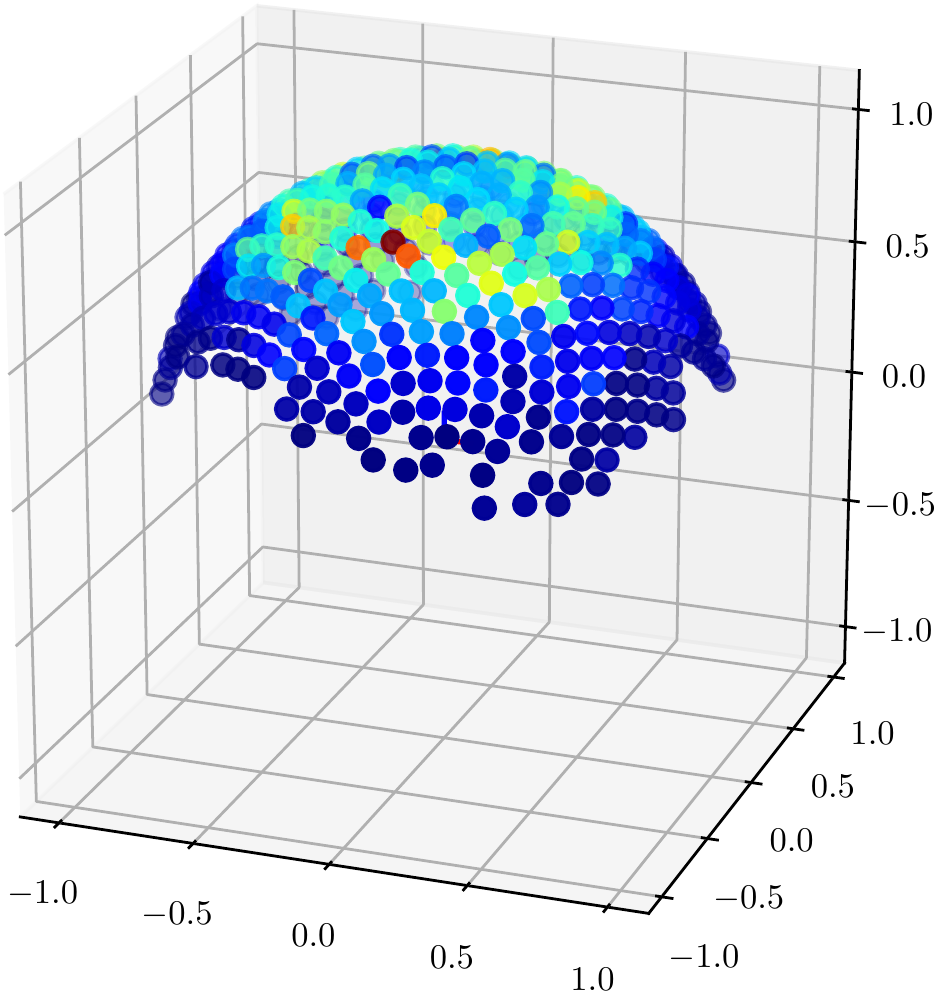}
}
\hspace{-0.01\linewidth}
\subfloat[]{
    \includegraphics[height=\objheight\linewidth, valign=t]{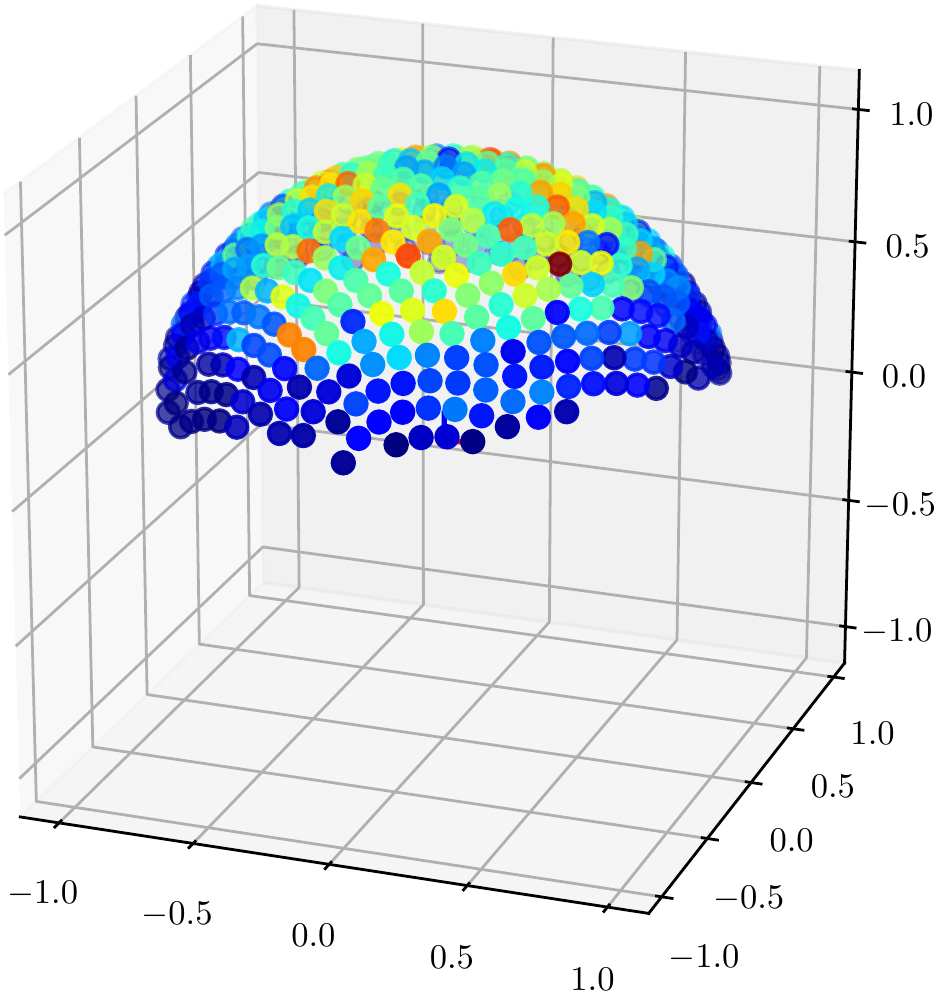}
}
\\
\subfloat[]{
    \includegraphics[height=\objheight\linewidth, valign=t]{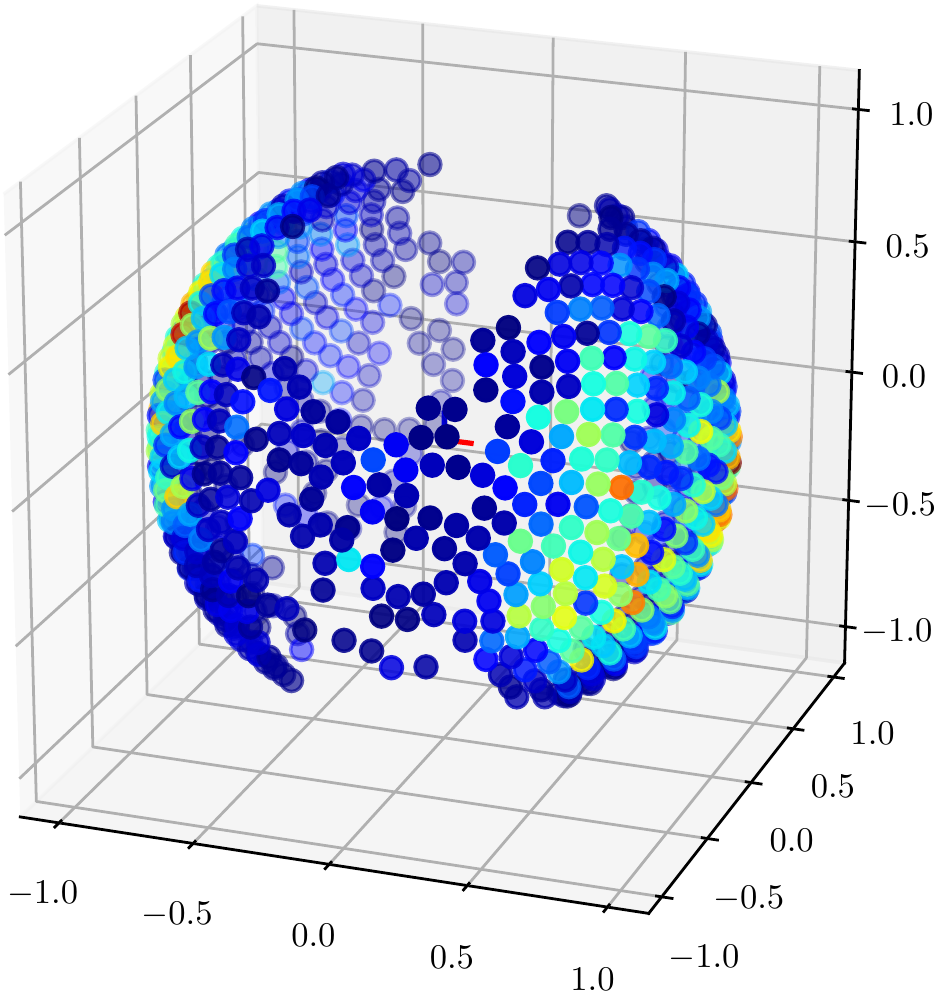}
}
\hspace{-0.01\linewidth}
\subfloat[]{
    \includegraphics[height=\objheight\linewidth, valign=t]{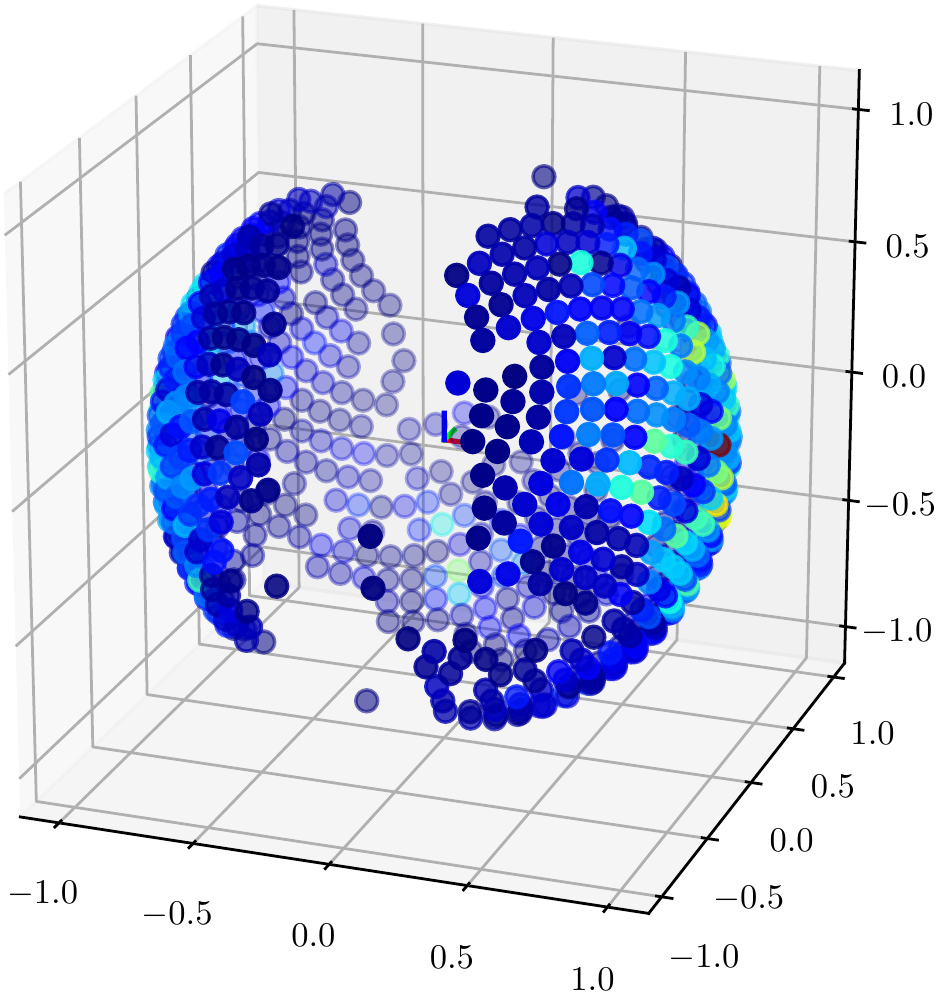}
}
\hspace{-0.01\linewidth}
\subfloat[]{
    \includegraphics[height=\objheight\linewidth, valign=t]{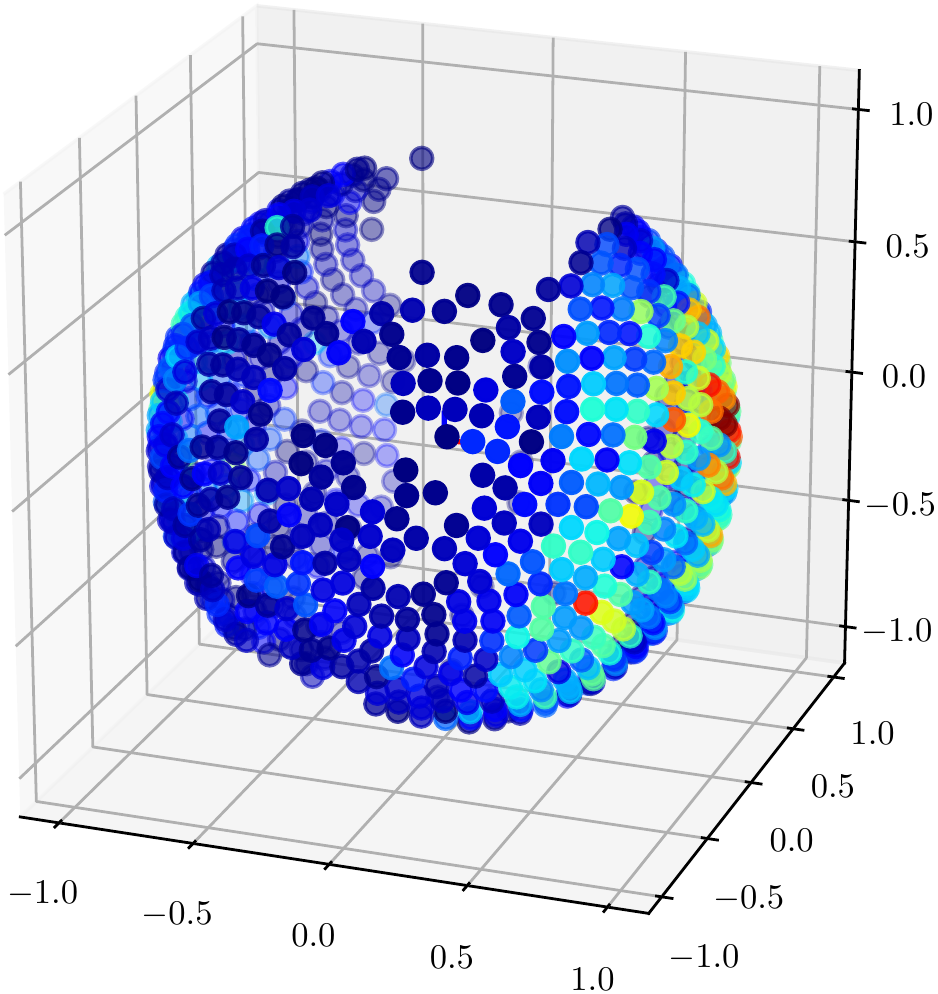}
}
\hspace{-0.01\linewidth}
\subfloat[]{
    \includegraphics[height=\objheight\linewidth, valign=t]{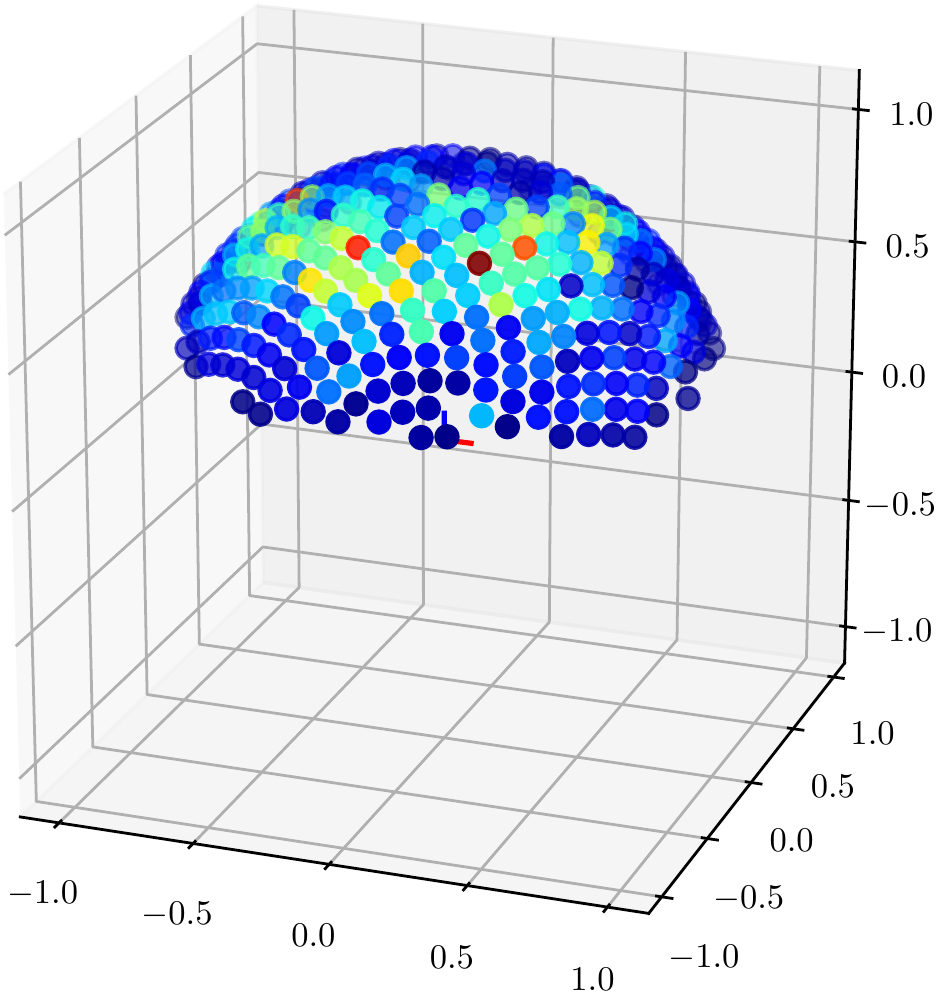}
}
\hspace{-0.01\linewidth}
\subfloat[]{
    \includegraphics[height=\objheight\linewidth, valign=t]{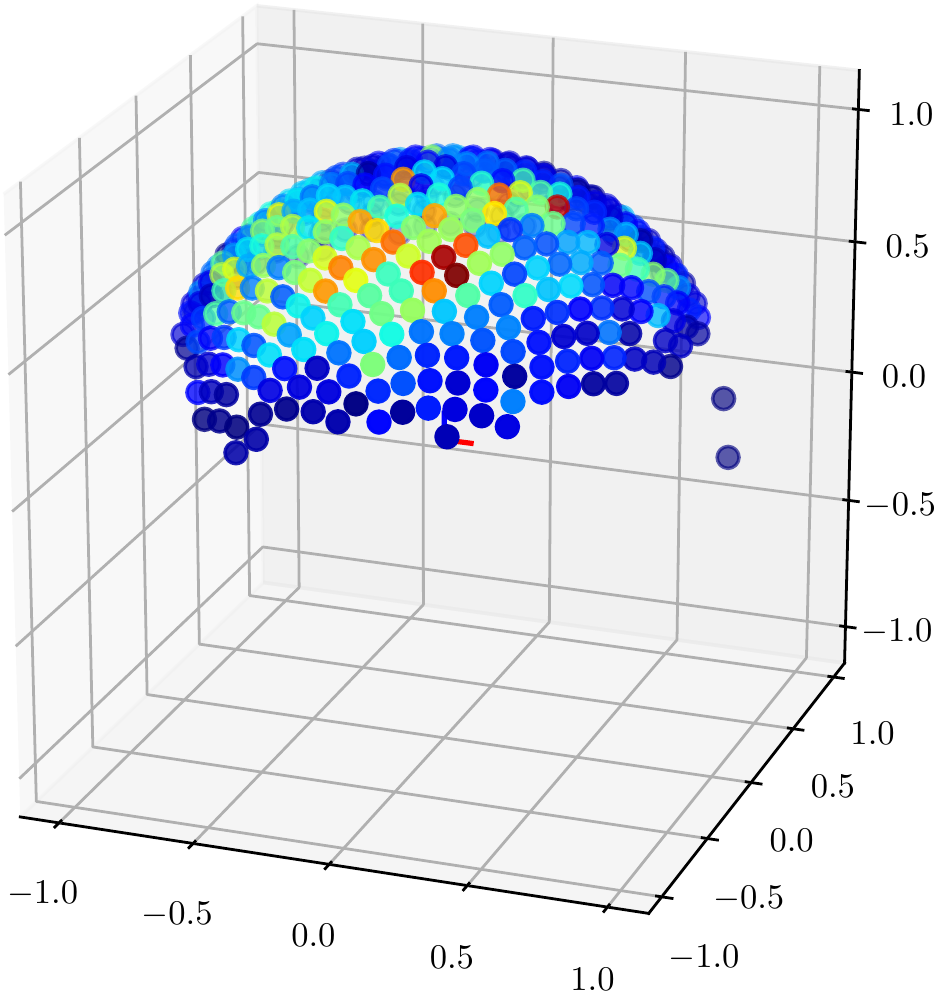}
}
\hspace{-0.01\linewidth}
\subfloat[]{
    \includegraphics[height=\objheight\linewidth, valign=t]{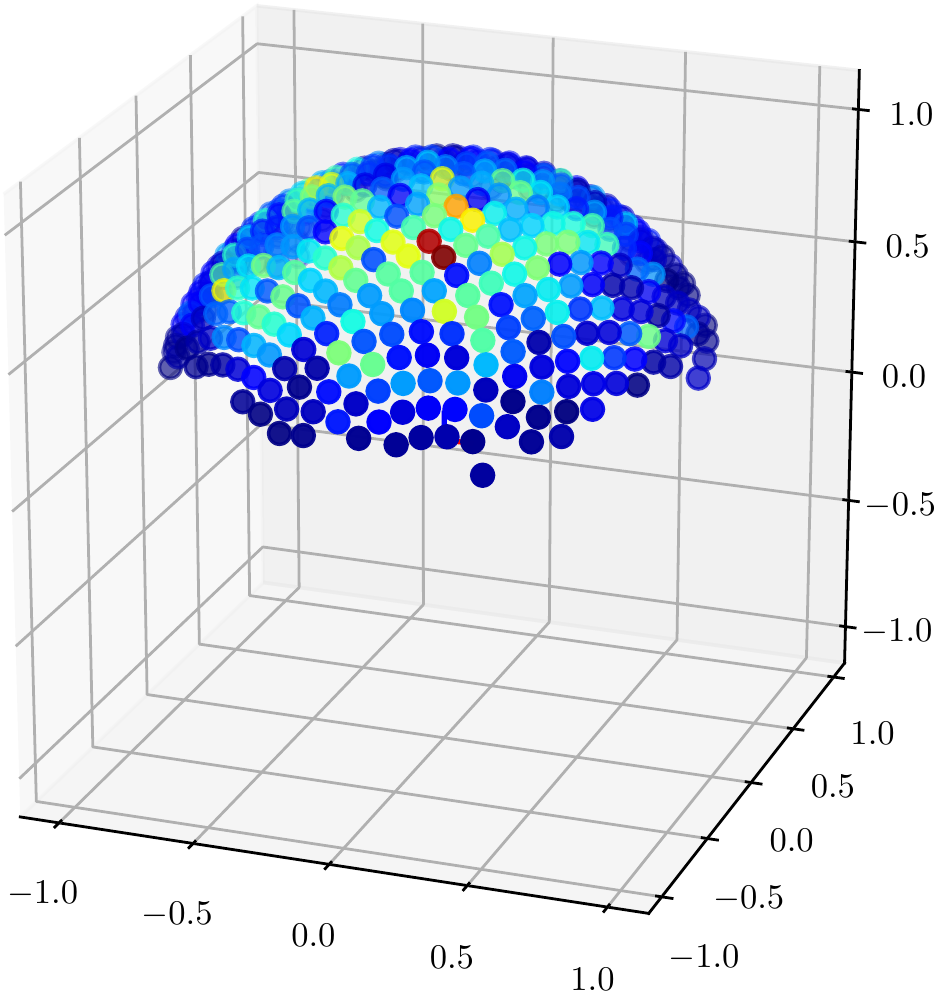}
}
\caption{\textbf{Viewpoint distribution of the 18 objects in our dataset.}
(a) \texttt{blade\_razor}; 
(b) \texttt{hammer}; 
(c) \texttt{needle\_nose\_pliers}
(d) \texttt{screwdriver}; 
(e) \texttt{side\_cutters}; 
(f) \texttt{tape\_measure};
(g) \texttt{wire\_stripper};
(h) \texttt{wrench}; 
(i) \texttt{centrifuge\_tube};
(j) \texttt{microplate};
(k) \texttt{tube\_rack\_2};
(l) \texttt{tube\_rack\_50};
(m) \texttt{pipette\_0.5\_10};
(n) \texttt{pipette\_10\_100}; 
(o) \texttt{pipette\_100\_1000}; 
(p) \texttt{sterile\_rack\_10};
(q) \texttt{sterile\_rack\_200};
(r) \texttt{sterile\_rack\_1000}.}
\label{fig:viewpoint:distr}
\end{figure*}

\section{Viewpoint Coverage Distribution}
\label{sec:viewpoint:coverage}

Viewpoint coverage percentage is illustrated in Section 4.2 and Figure 5 of the main paper.
We illustrate a more detailed viewpoint distribution for each object in Figure \ref{fig:viewpoint:distr}.
The viewpoints are drawn as 3D points on the unit sphere centered at the object center.
Their positions on the unit sphere are determined by the Azimuth and Elevation of the viewpoint.
Their density on the sphere is shown by heatmap color.
Notice that for objects such as \texttt{microplate} and \texttt{tape\_measure}, there is no viewpoint distributed on the $-z$ space, because there is only one possible side up when being put on a desktop.

\section{More Visualizations of Data Samples}
\label{sec:more:viz}

\begin{figure*}[h]
\newcommand\sampleheight{0.19}
\centering
\small
\subfloat{
    \includegraphics[height=\sampleheight\linewidth, valign=t]{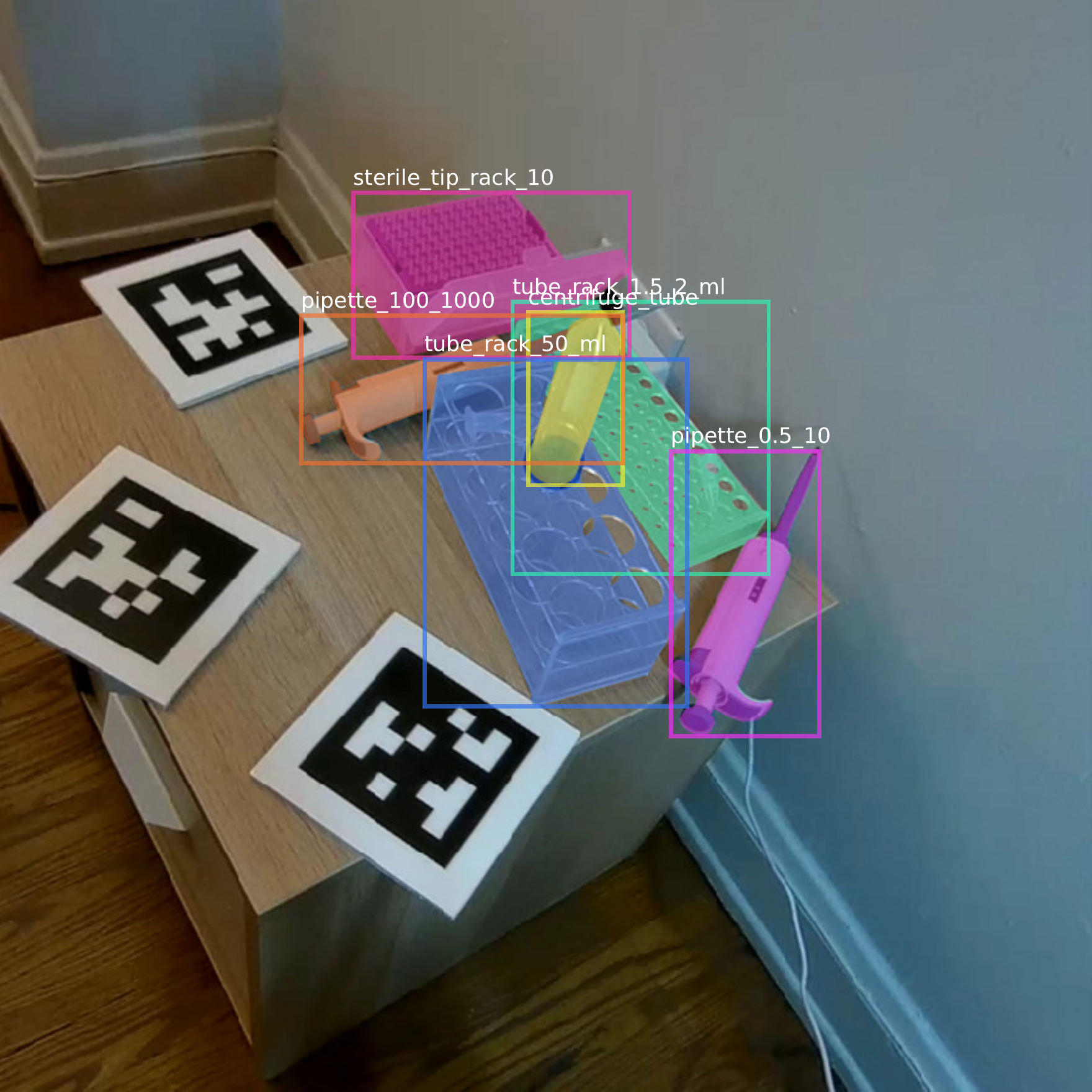}
}
\subfloat{
    \includegraphics[height=\sampleheight\linewidth, valign=t]{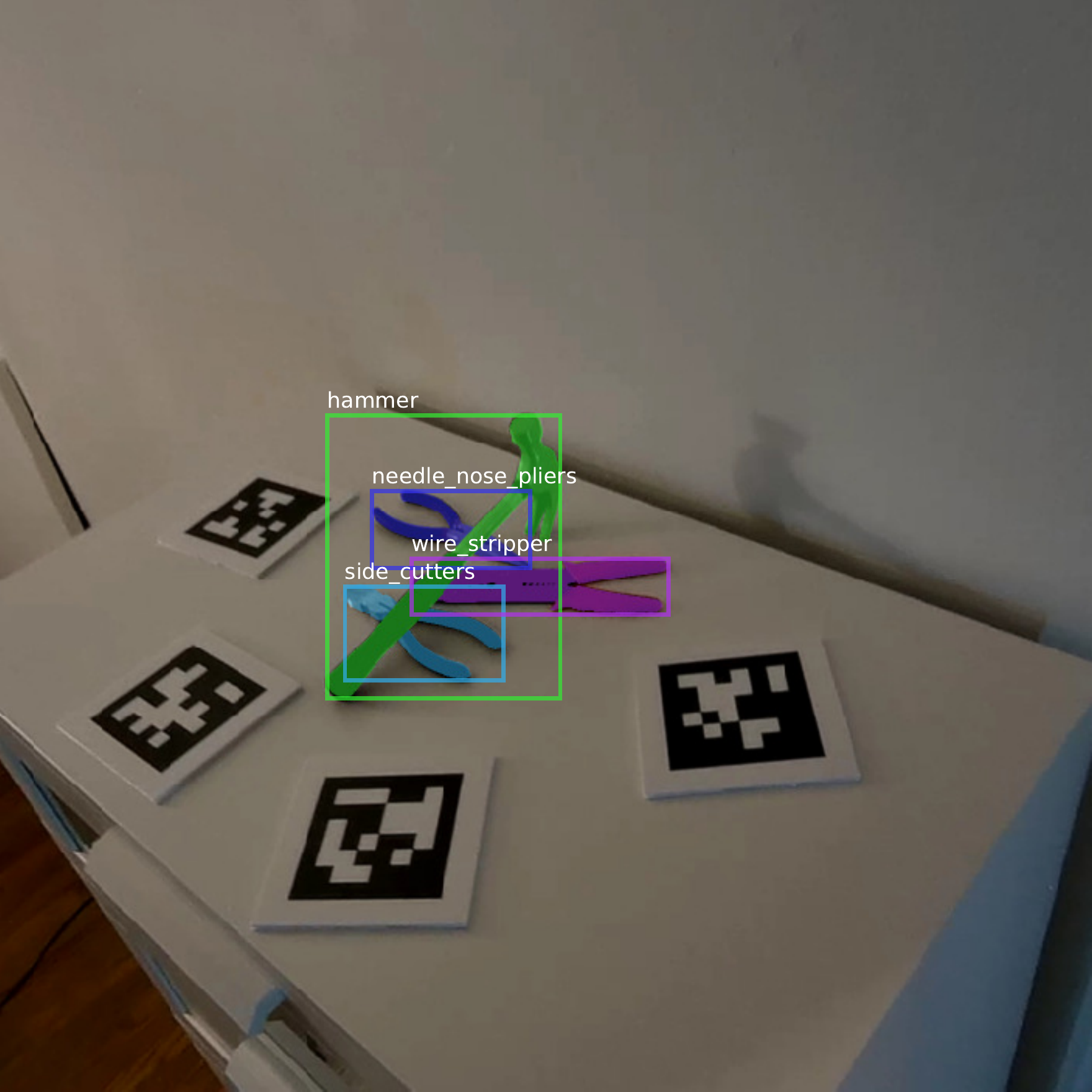}
}
\subfloat{
    \includegraphics[height=\sampleheight\linewidth, valign=t]{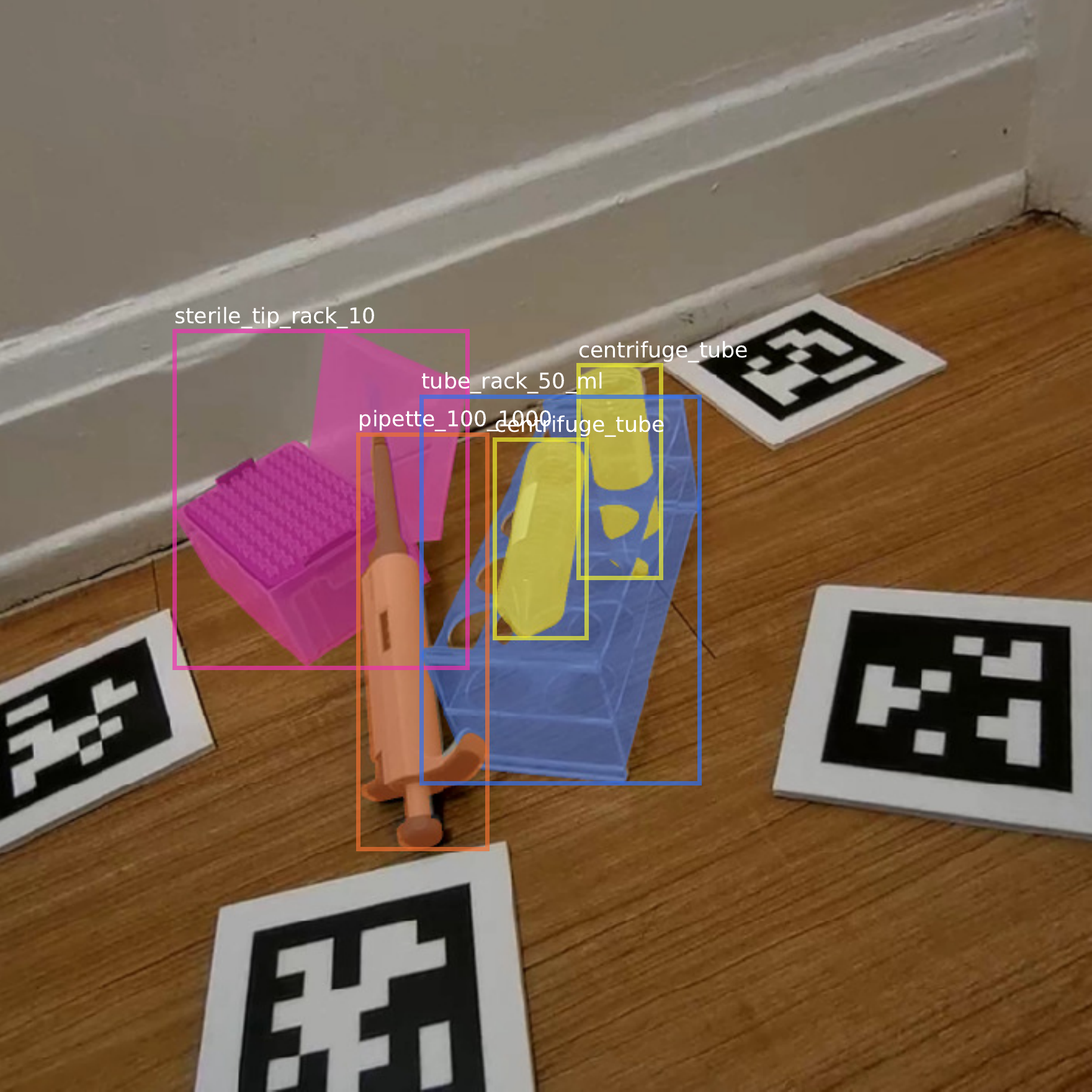}
}
\subfloat{
    \includegraphics[height=\sampleheight\linewidth, valign=t]{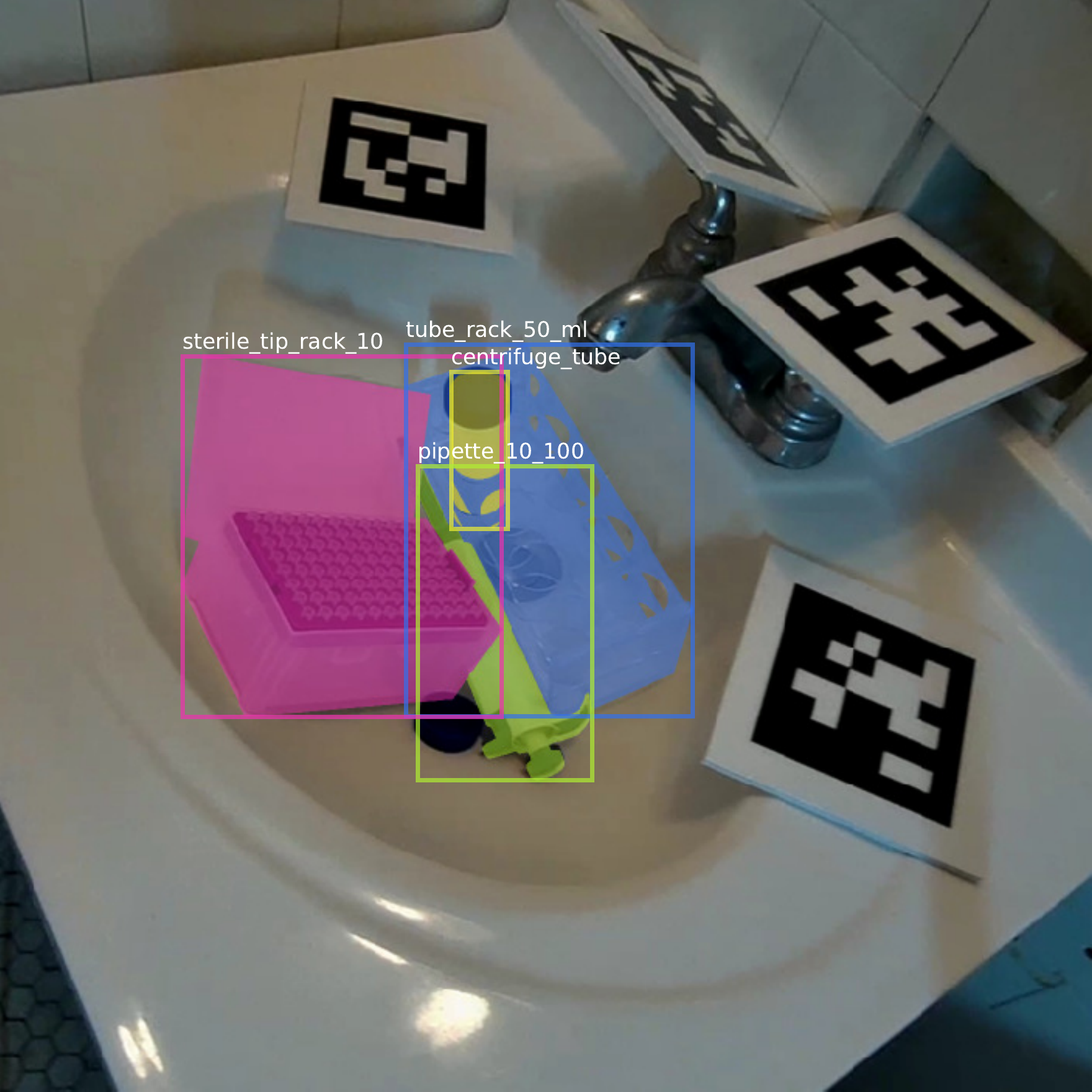}
}
\subfloat{
    \includegraphics[height=\sampleheight\linewidth, valign=t]{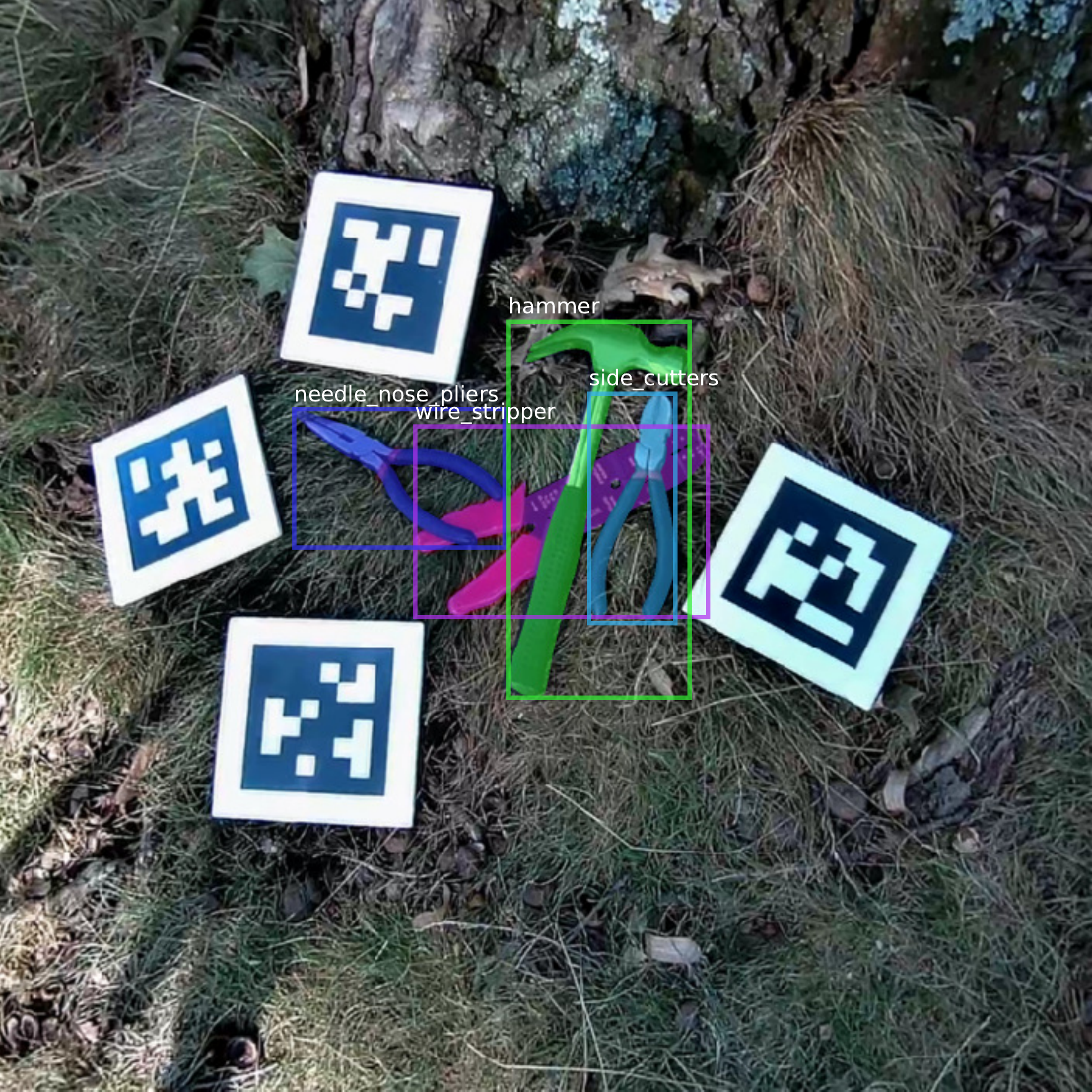}
}
\\
\vspace{-1.5ex}
\subfloat{
    \includegraphics[height=\sampleheight\linewidth, valign=t]{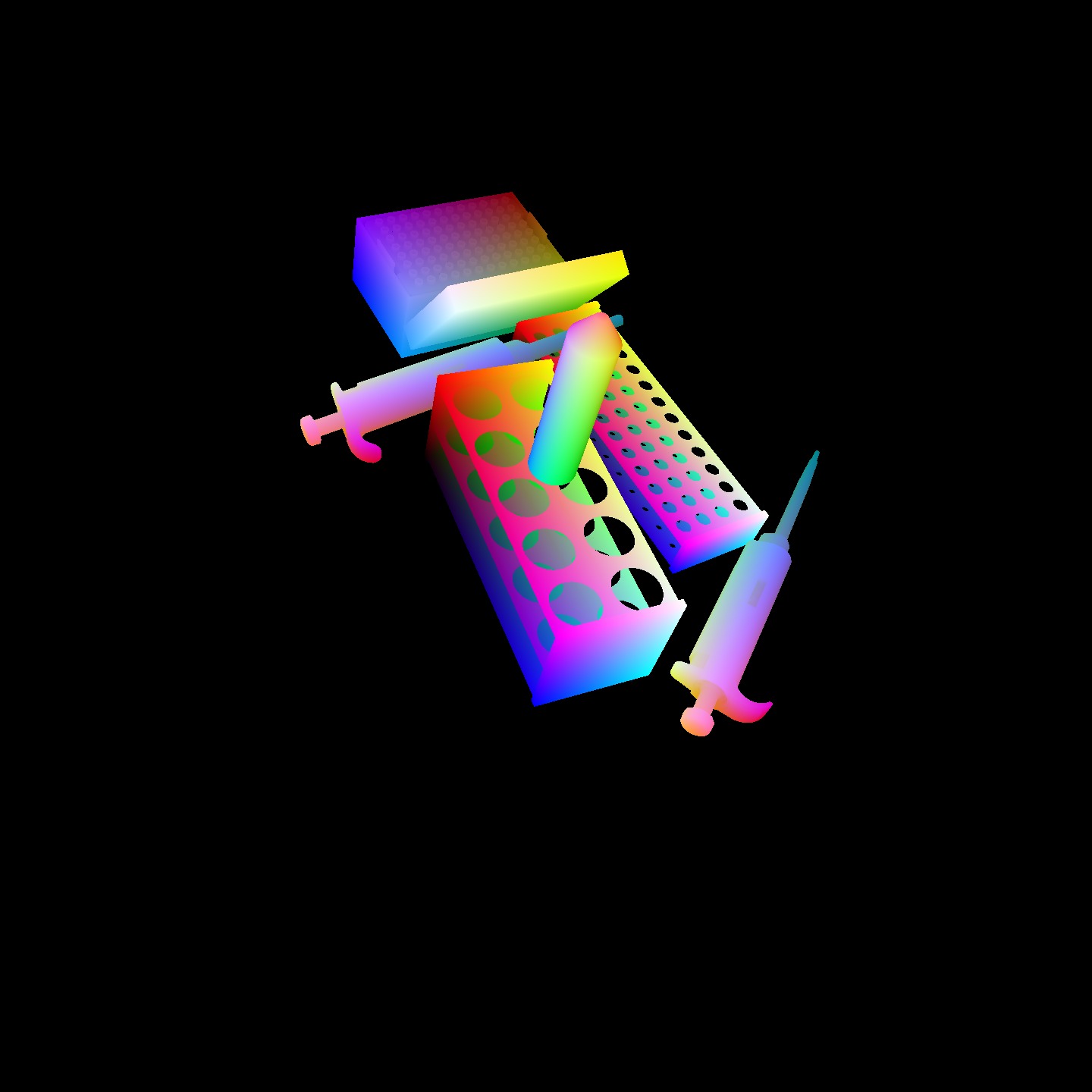}
}
\subfloat{
    \includegraphics[height=\sampleheight\linewidth, valign=t]{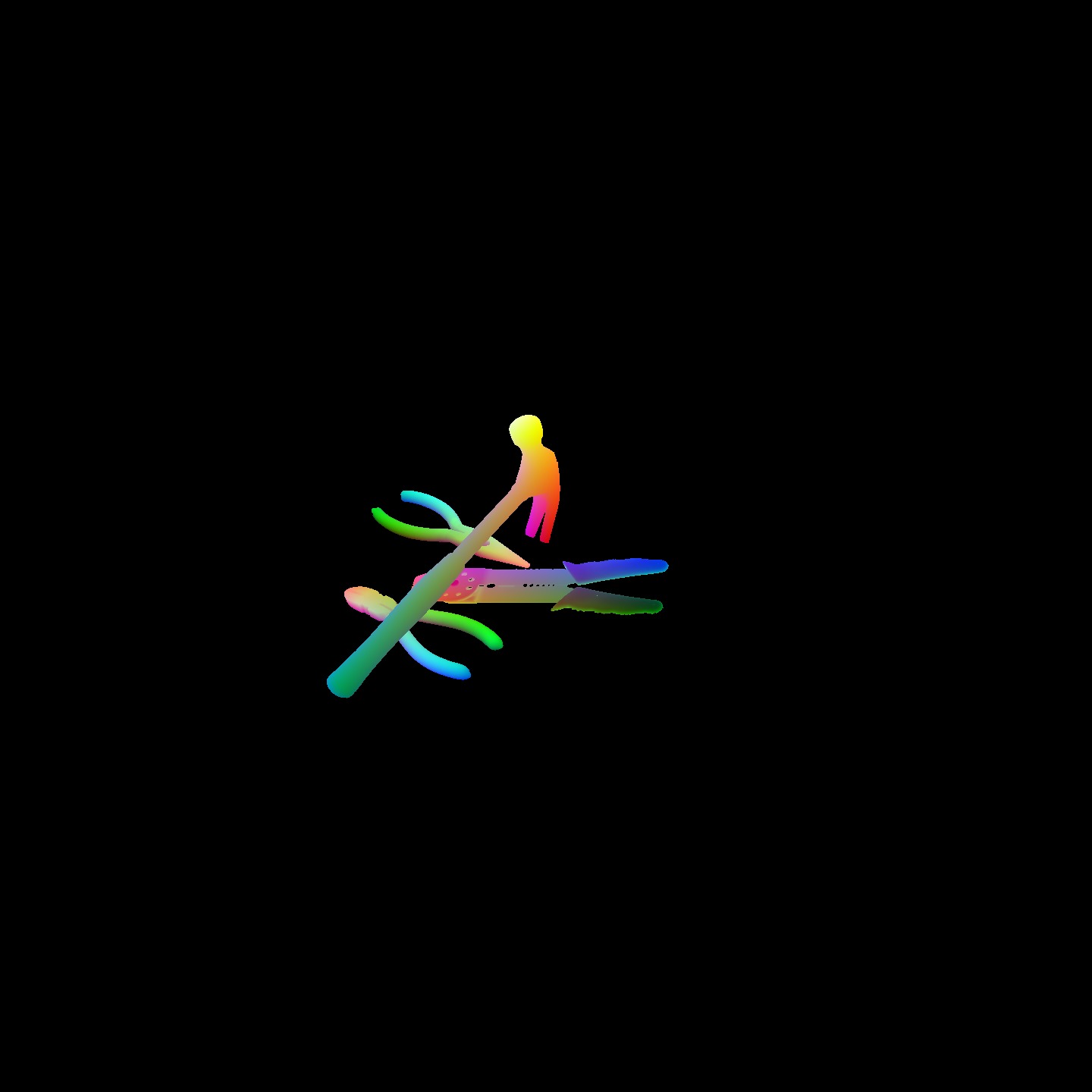}
}
\subfloat{
    \includegraphics[height=\sampleheight\linewidth, valign=t]{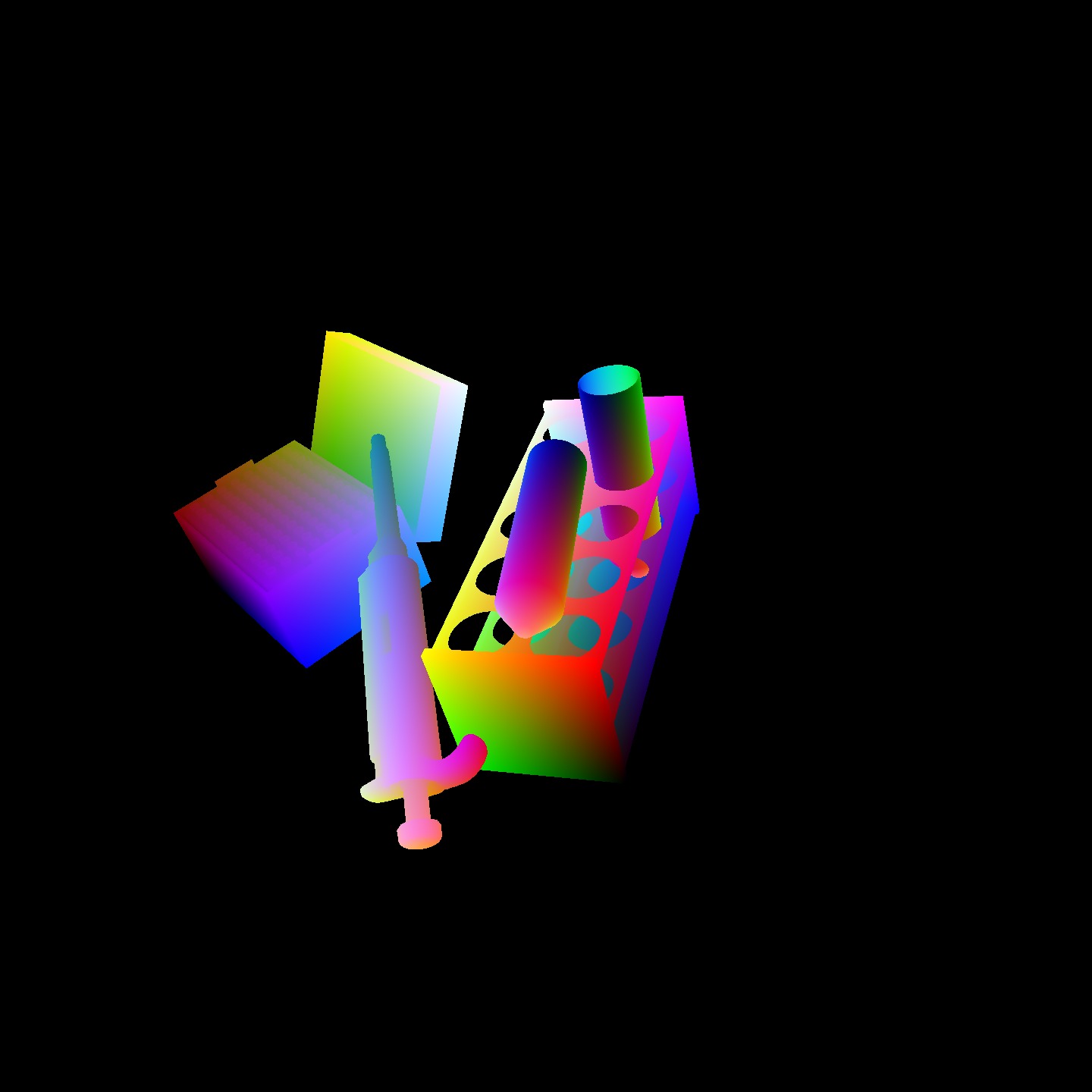}
}
\subfloat{
    \includegraphics[height=\sampleheight\linewidth, valign=t]{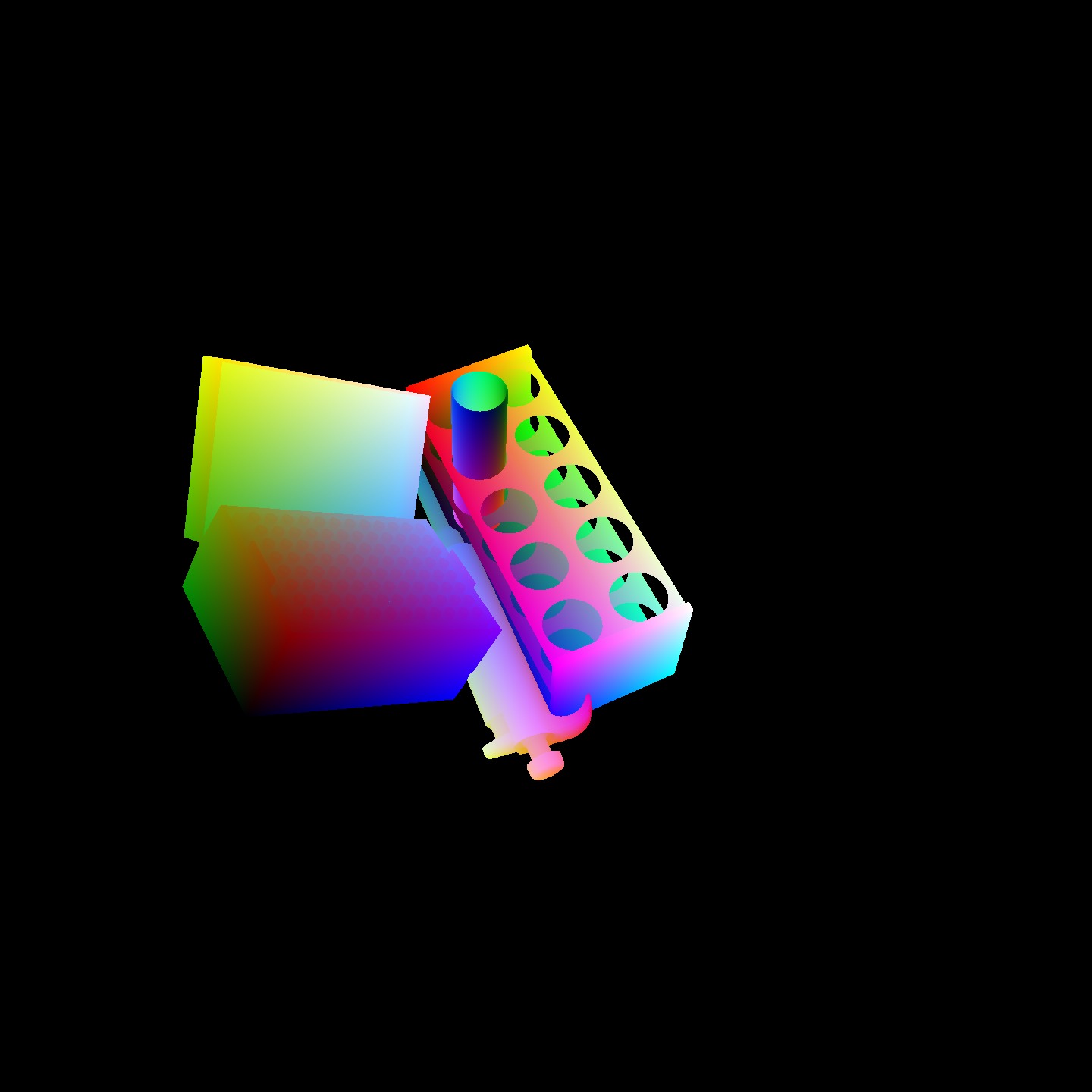}
}
\subfloat{
    \includegraphics[height=\sampleheight\linewidth, valign=t]{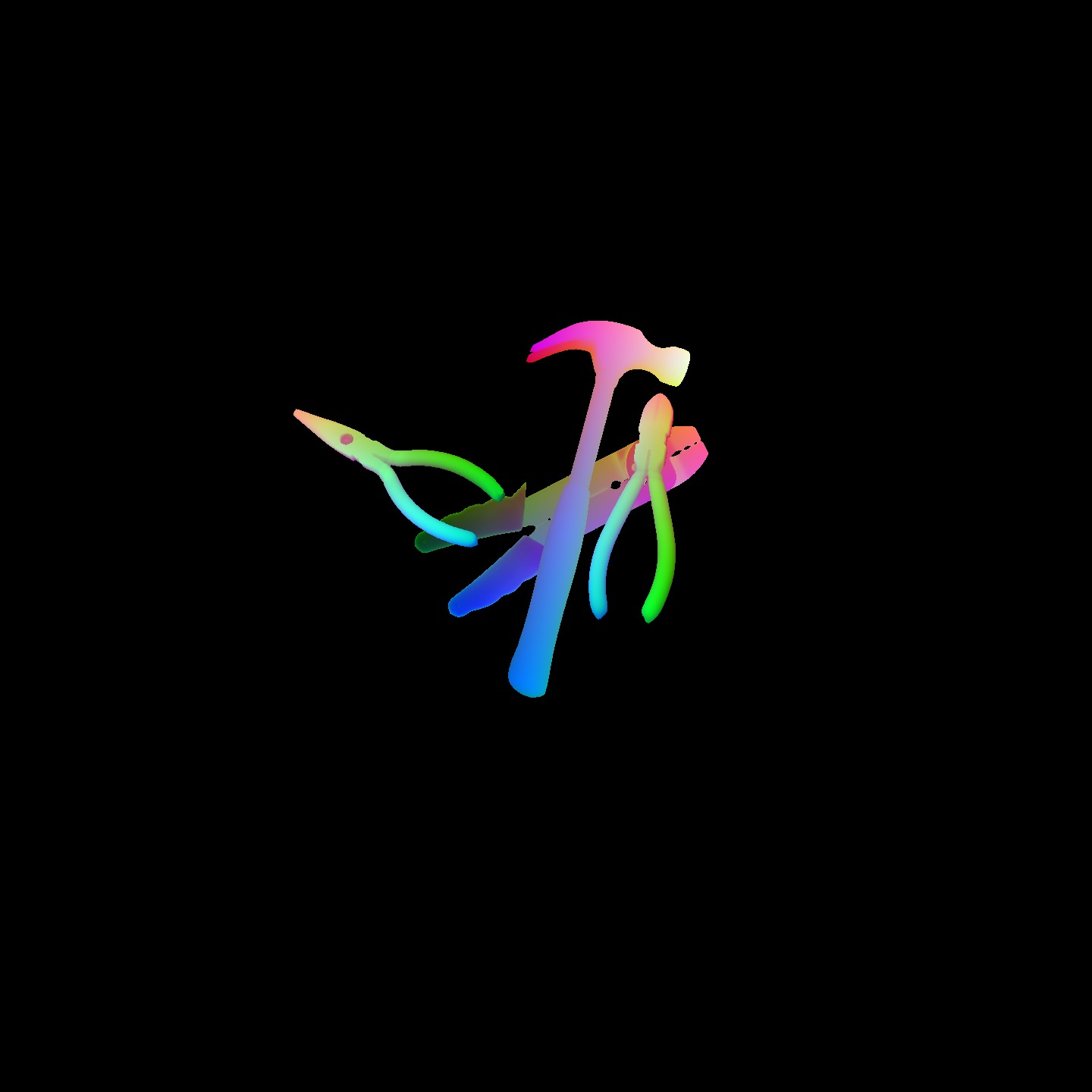}
}
\\
\vspace{-1.5ex}
\subfloat{
    \includegraphics[height=\sampleheight\linewidth, valign=t]{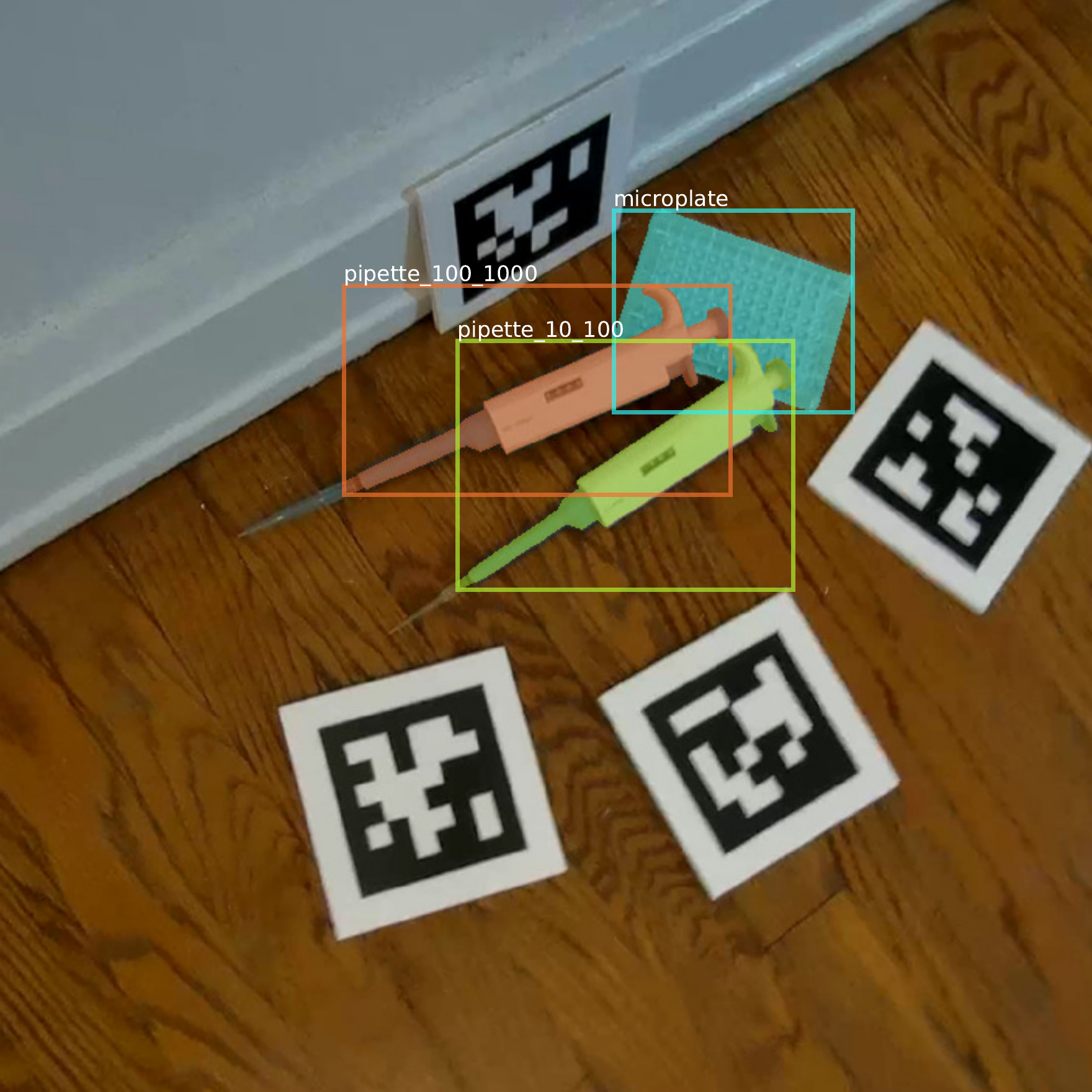}
}
\subfloat{
    \includegraphics[height=\sampleheight\linewidth, valign=t]{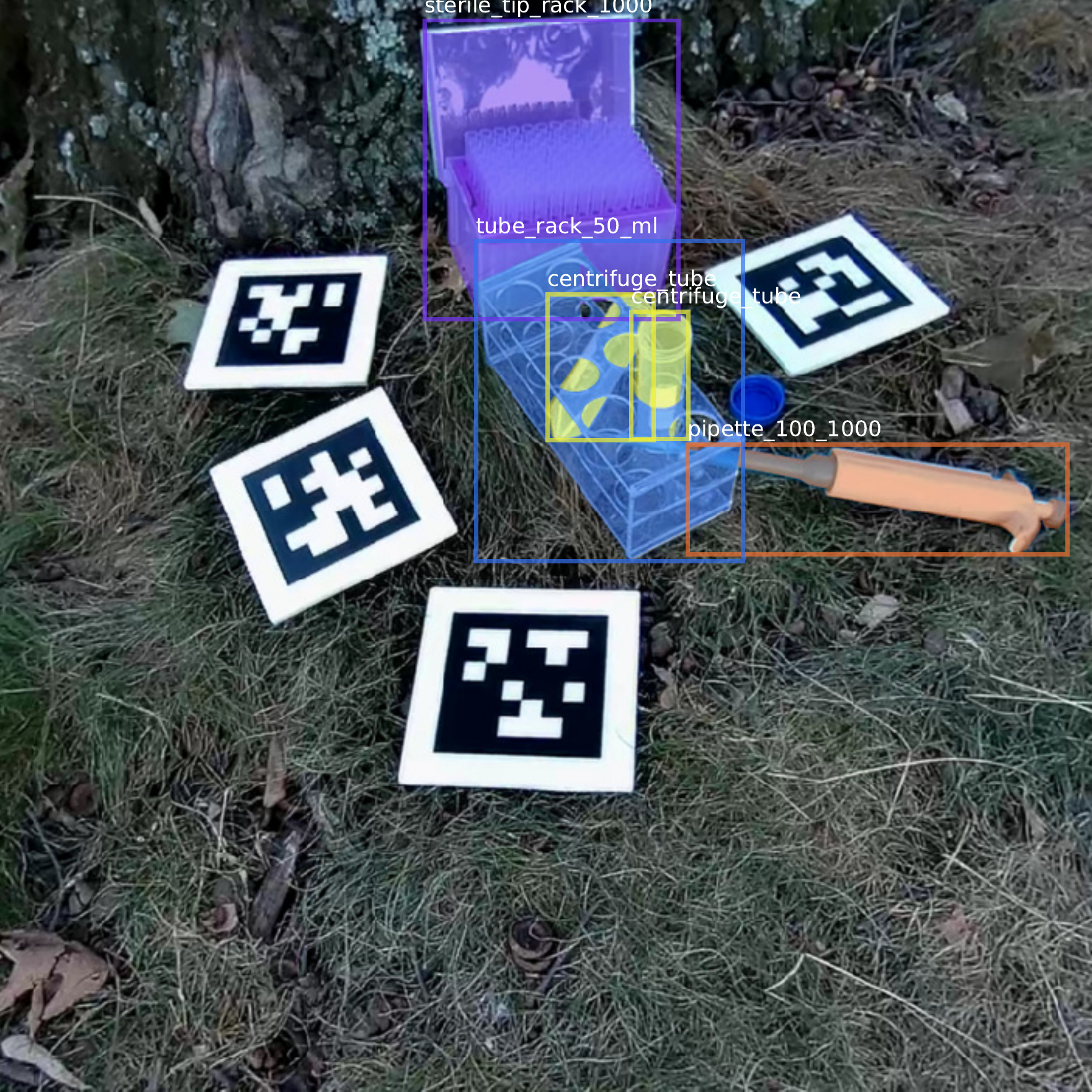}
}
\subfloat{
    \includegraphics[height=\sampleheight\linewidth, valign=t]{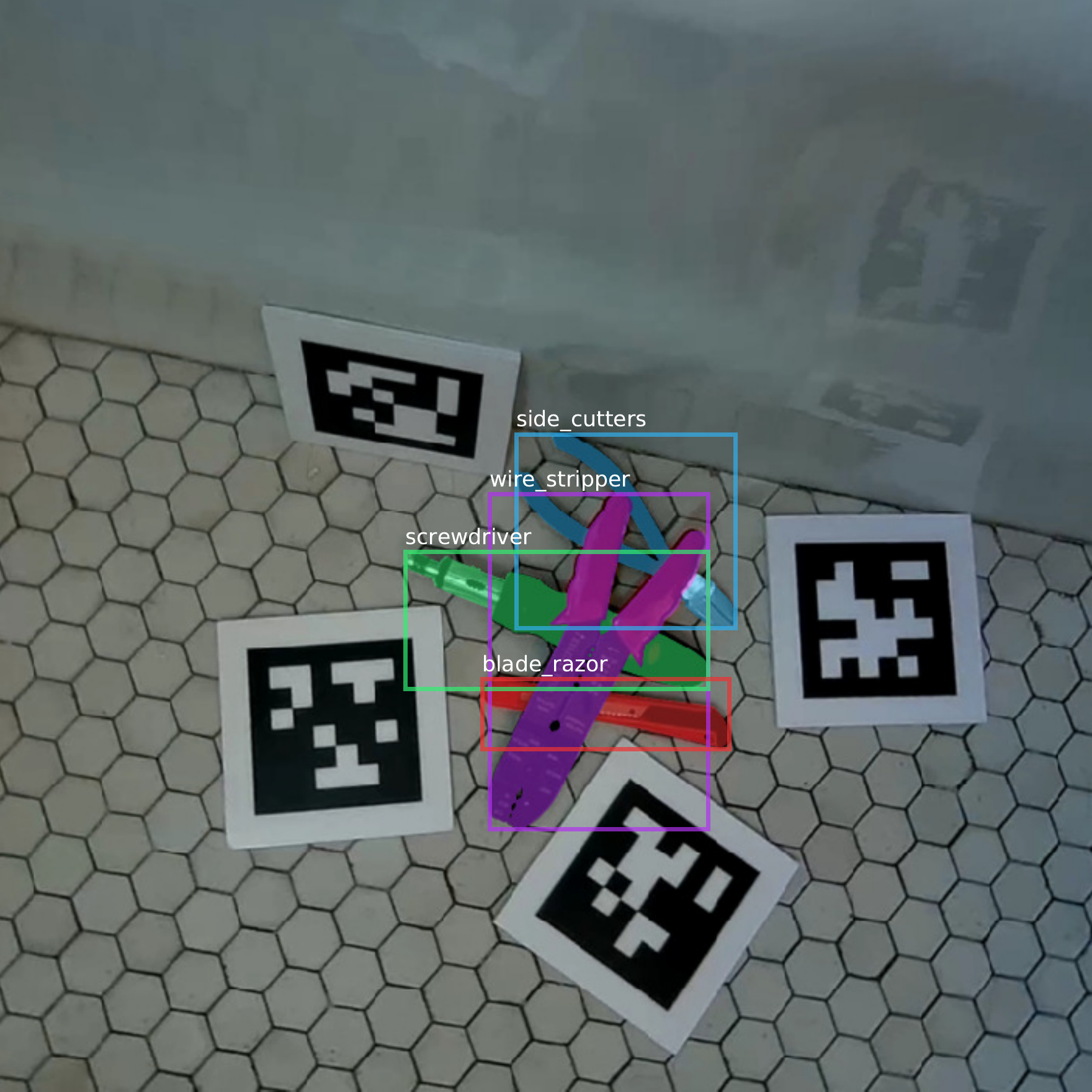}
}
\subfloat{
    \includegraphics[height=\sampleheight\linewidth, valign=t]{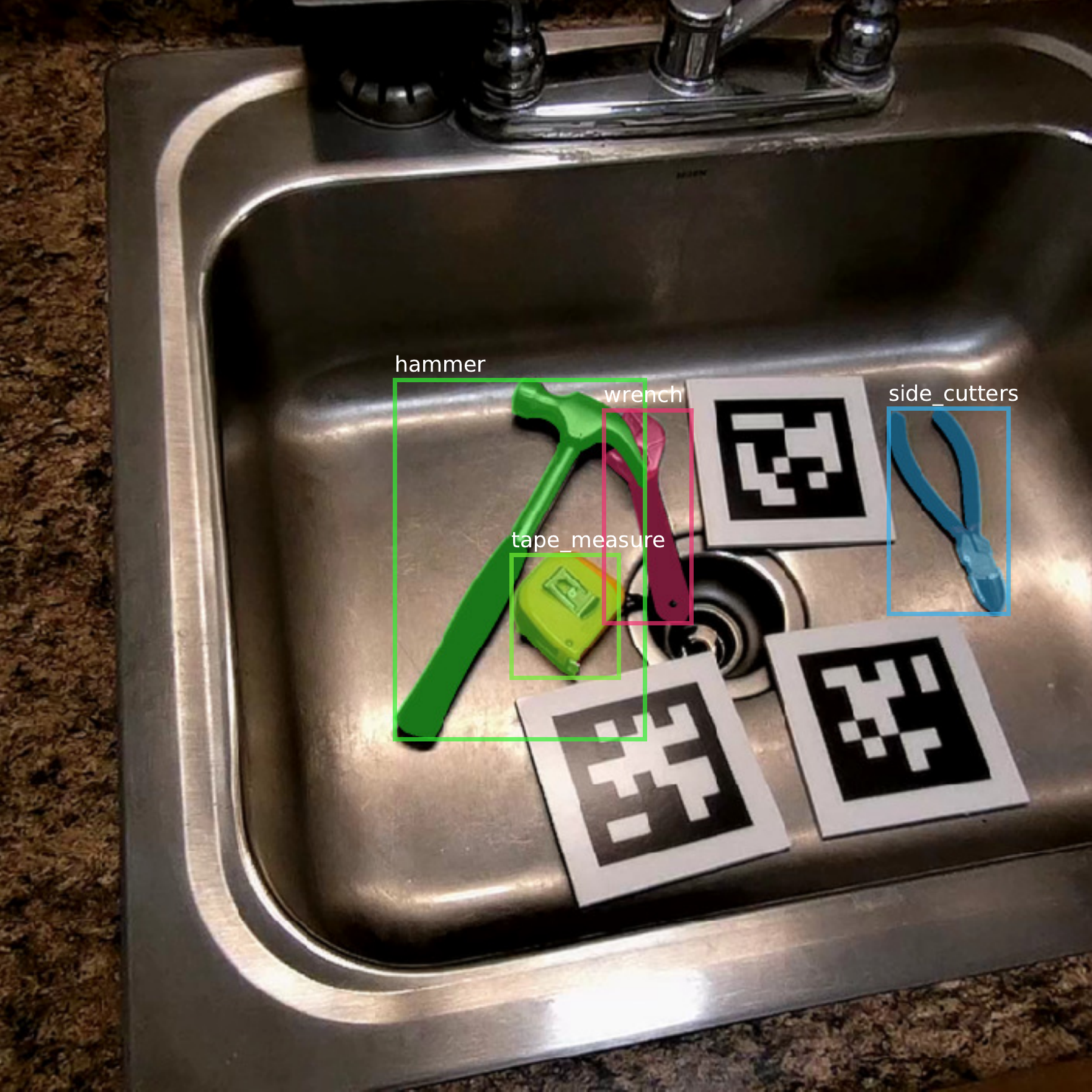}
}
\subfloat{
    \includegraphics[height=\sampleheight\linewidth, valign=t]{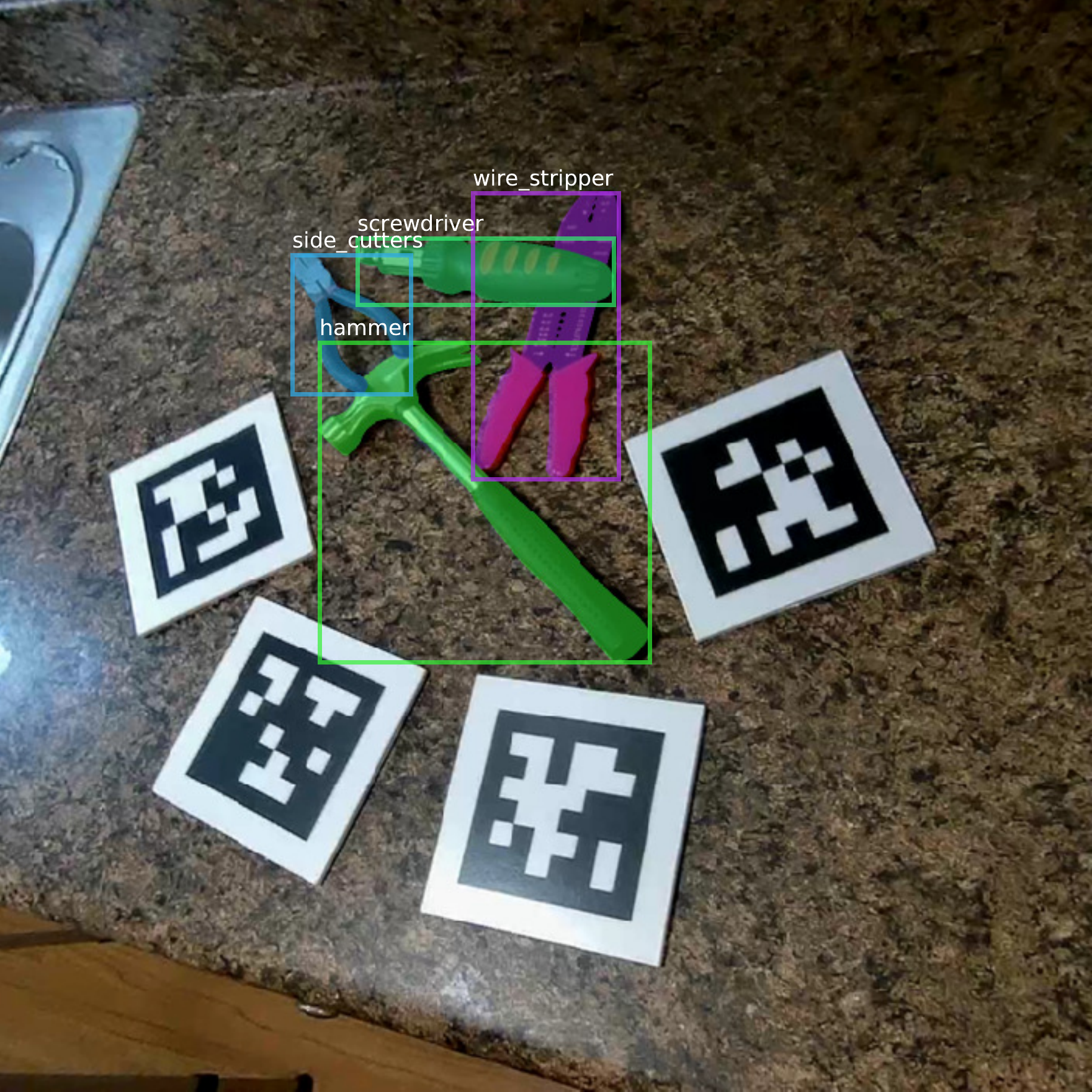}
}
\\
\vspace{-1.5ex}
\subfloat{
    \includegraphics[height=\sampleheight\linewidth, valign=t]{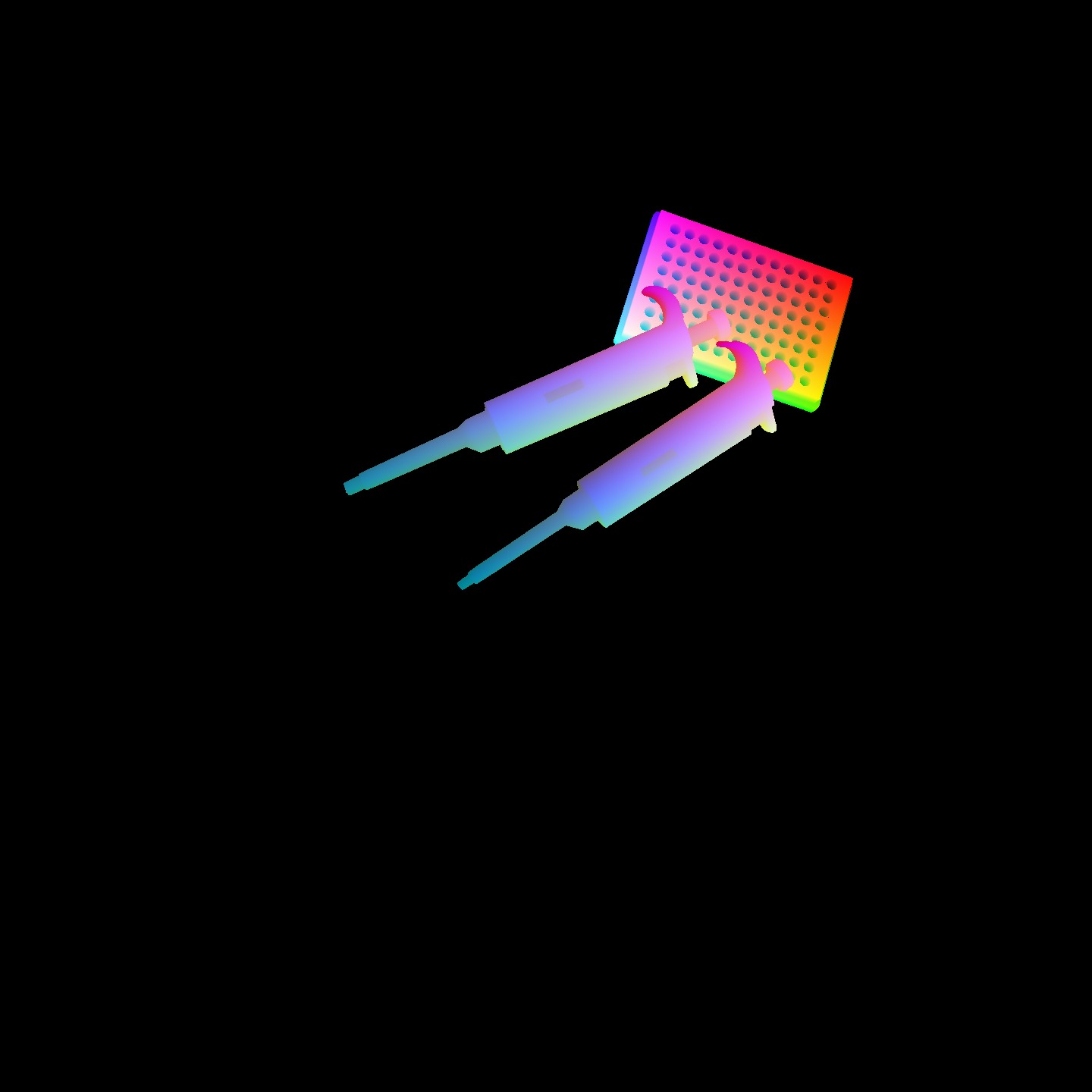}
}
\subfloat{
    \includegraphics[height=\sampleheight\linewidth, valign=t]{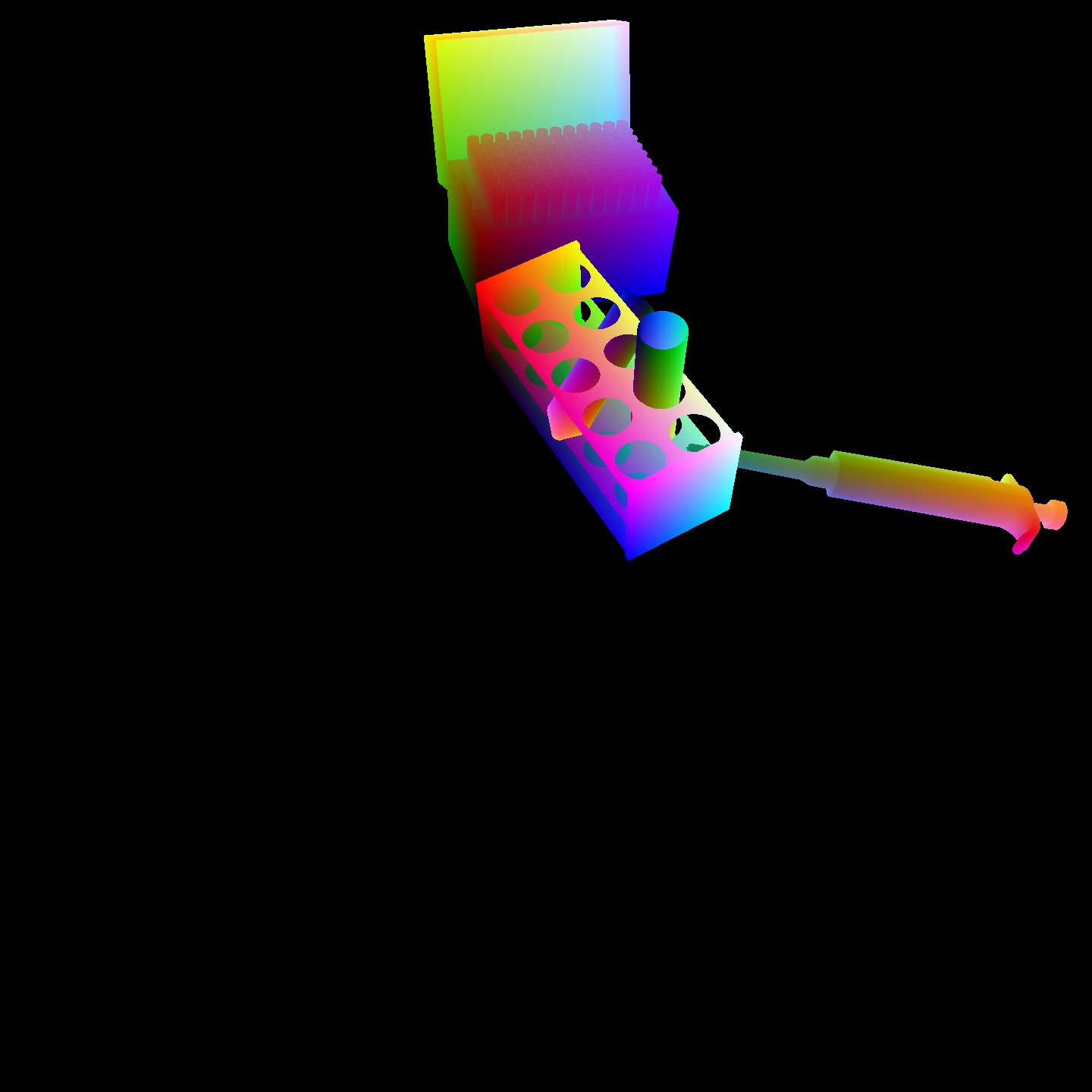}
}
\subfloat{
    \includegraphics[height=\sampleheight\linewidth, valign=t]{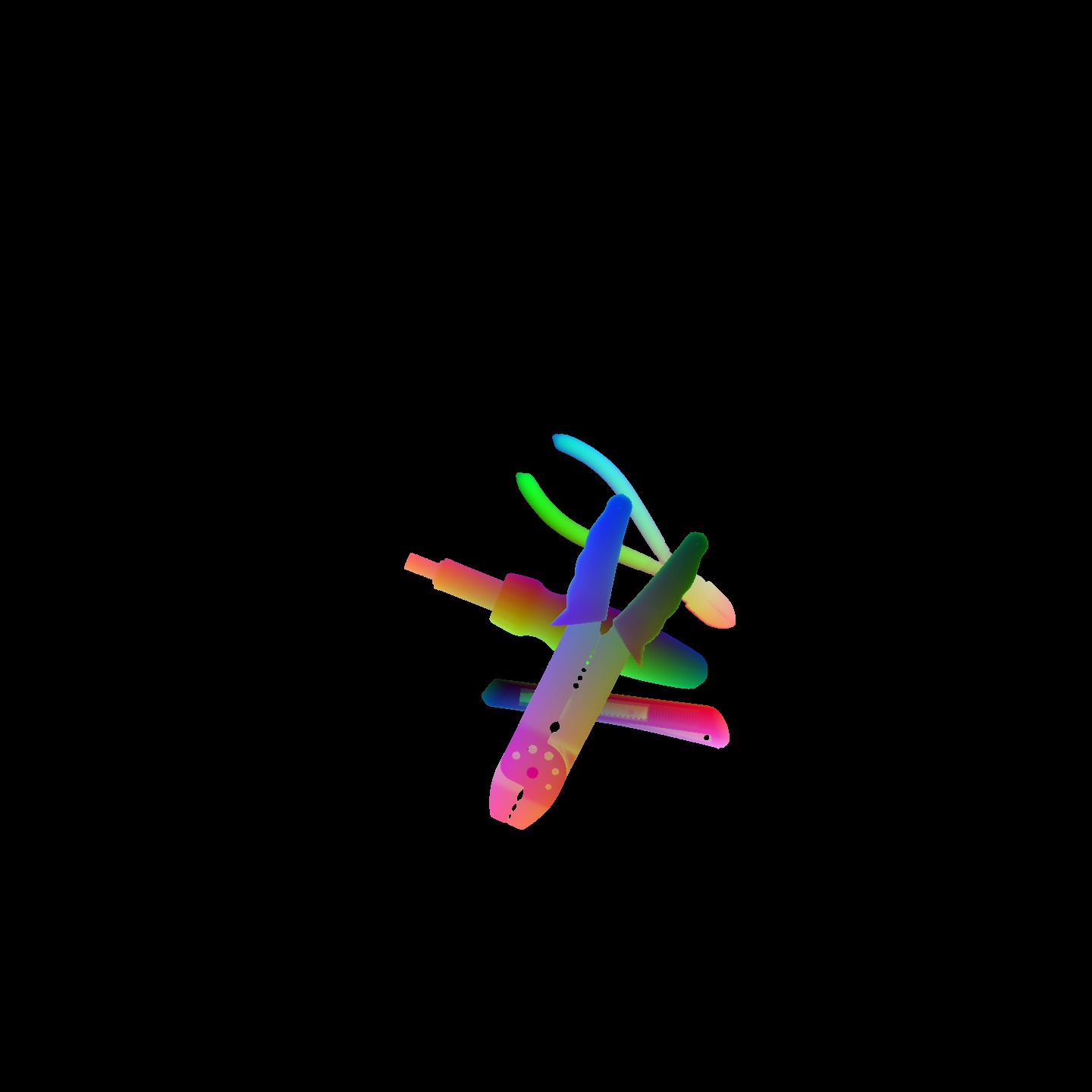}
}
\subfloat{
    \includegraphics[height=\sampleheight\linewidth, valign=t]{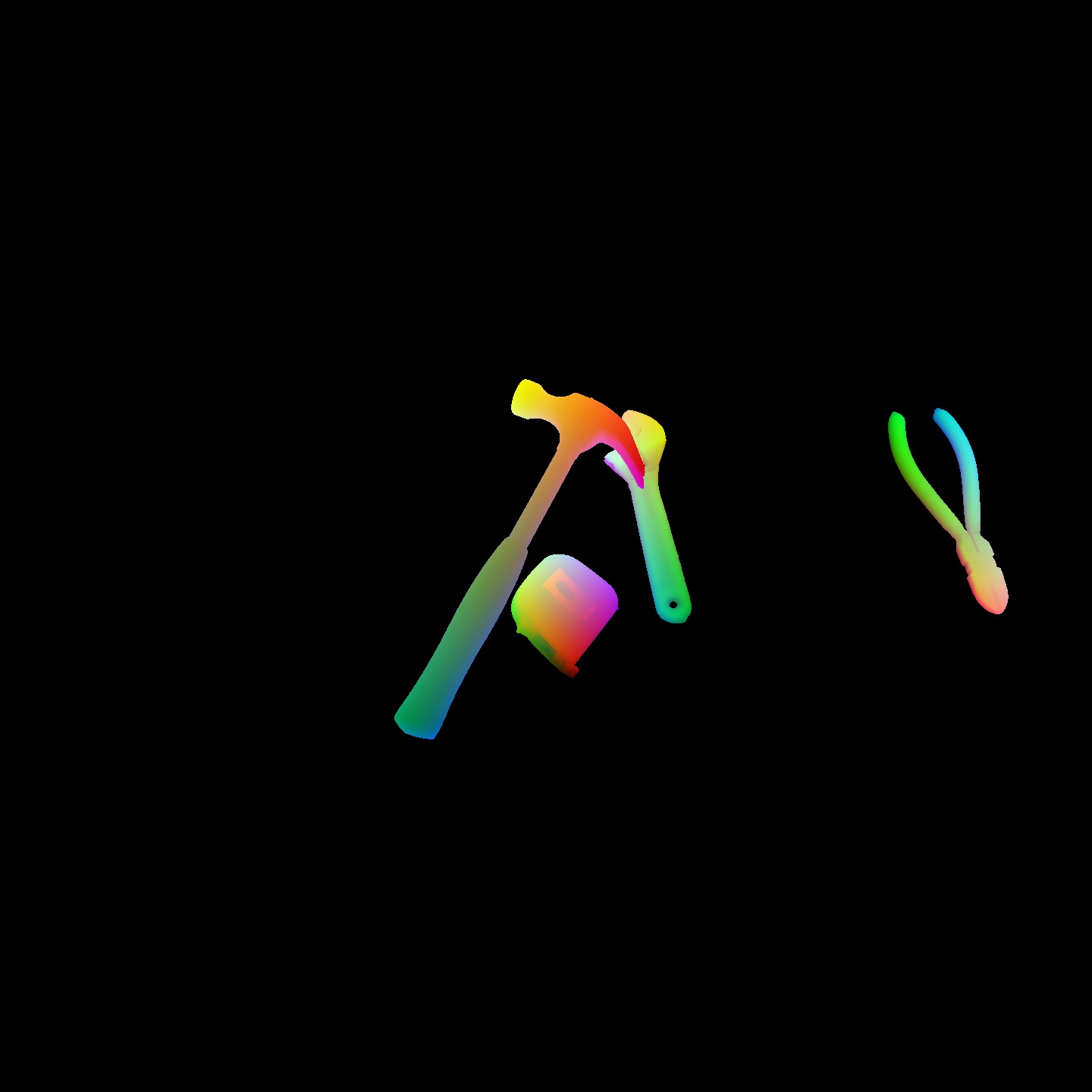}
}
\subfloat{
    \includegraphics[height=\sampleheight\linewidth, valign=t]{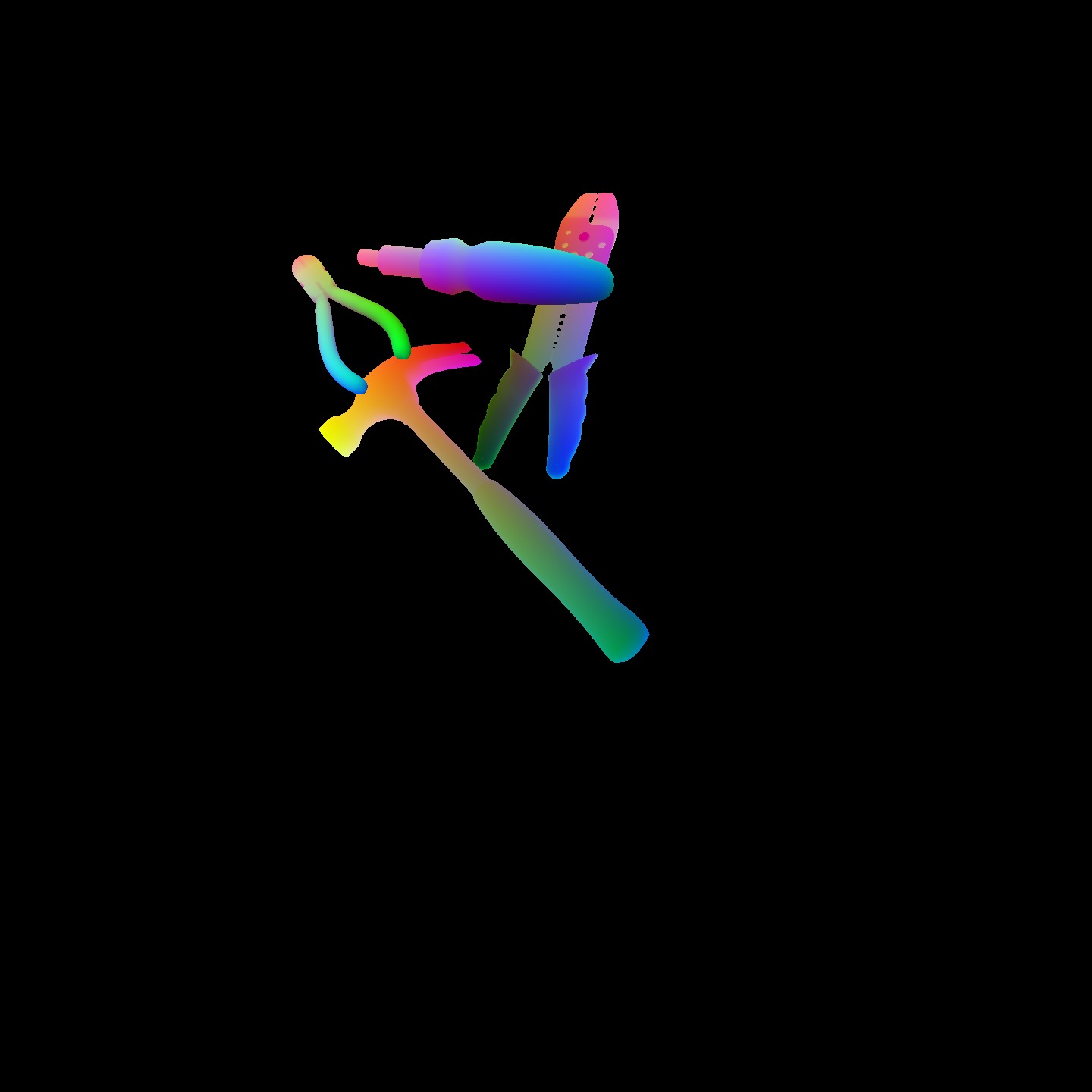}
}
\caption{
\textbf{Visualization of data samples from StereOBJ-1M dataset. }
The first row is left stero images with semantic masks and bounding boxes superimposed.
In the second row, we use normalized coordinate map \cite{pix2pose,nocs} to illustrate the 6D poses of the corresponding objects, where the coordinates of the object surface points are normalized to $[0,1]^3$ and converted to RGB values in $[0,255]^3$ at projected pixels.
}
\vspace{-1ex}
\label{fig:stereobj:samples:supp}
\end{figure*}

We provide more visualizations of data samples from our dataset.
As illustrated in Figure \ref{fig:stereobj:samples:supp}, the data annotation has high precision.

\section{Change Log}

Mar 15, 2022: Updated dataset statistics and baseline performance results after cleaning up the data.

\end{document}